\documentclass{article} 
\usepackage{iclr2023_conference,times}

\usepackage{amsmath,amsfonts,bm}

\def\eqref#1{equation~\ref{#1}}

\def\1{\bm{1}}

\DeclareMathAlphabet{\mathsfit}{\encodingdefault}{\sfdefault}{m}{sl}
\SetMathAlphabet{\mathsfit}{bold}{\encodingdefault}{\sfdefault}{bx}{n}

\usepackage{multirow}
\usepackage{subfig}
\usepackage{float}
\usepackage{multicol}
\usepackage{lipsum}
\usepackage{csquotes}
\usepackage{array}
\usepackage{enumitem}
\usepackage[scientific-notation=true]{siunitx}
\usepackage[normalem]{ulem}
\useunder{\uline}{\ul}{}
\usepackage{array}
\usepackage{amsmath,amssymb,amsfonts}
\usepackage{algorithmic}
\usepackage[thinlines]{easytable}
\usepackage{graphicx}
\usepackage{textcomp}
\usepackage{xcolor}
\usepackage[utf8]{inputenc}

\title{Multi-Knowledge Fusion Network for Time Series Representation Learning}

\author{Sagar Srinivas Sakhinana\thanks{Conceived, designed, implemented the research(programmed the software) and drafted the manuscript}, \{Shivam Gupta, Krishna Sai Sudhir Aripirala\}\thanks{Performed computational experiments, interpretation and visualization analysis of the results}, Venkataramana Runkana \\
TCS - Research\\
\texttt{\{sagar.sakhinana, g.shivam4, k.aripirala, venkat.runkana\}@tcs.com} \\
}

\iclrfinalcopy 
\begin{document}

\maketitle

\vspace{-2mm}
\begin{abstract} 
\vspace{-2mm}
Forecasting the behaviour of complex dynamical systems such as interconnected sensor networks characterized by high-dimensional multivariate time series(MTS) is of paramount importance for making informed decisions and planning for the future in a broad spectrum of applications. Graph forecasting networks(GFNs) are well-suited for forecasting MTS data that exhibit spatio-temporal dependencies. However, most prior works of GFN-based methods on MTS forecasting rely on domain-expertise to model the nonlinear dynamics of the system, but neglect the potential to leverage the inherent relational-structural dependencies among time series variables underlying MTS data. On the other hand, contemporary works attempt to infer the relational structure of the complex dependencies between the variables and simultaneously learn the nonlinear dynamics of the interconnected system but neglect the possibility of incorporating domain-specific prior knowledge to improve forecast accuracy. To this end, we propose a hybrid architecture that combines explicit prior knowledge with implicit knowledge of the relational structure within the MTS data. It jointly learns intra-series temporal dependencies and inter-series spatial dependencies by encoding time-conditioned structural spatio-temporal inductive biases to provide more accurate and reliable forecasts. It also models the time-varying uncertainty of the multi-horizon forecasts to support decision-making by providing estimates of prediction uncertainty. The proposed architecture has shown promising results on multiple benchmark datasets and outperforms state-of-the-art forecasting methods by a significant margin. We report and discuss the ablation studies to validate our forecasting architecture.
\end{abstract}

\vspace{-6mm}
\section{Introduction}
\vspace{-4mm}
Accurate multivariate time series forecasting(MTSF) is critical for a broad spectrum of domains that have significant financial or operational impacts, including retail and finance, intelligent transportation systems, logistics and supply chain management, and many others. However, MTSF can be challenging due to the complexity of the relationships between time series variables and the unique characteristics of the MTS data, such as non-linearity, heterogeneity, sparsity, and non-stationarity.
In this context, Spatial-temporal graph neural networks(STGNNs) have been widely studied for modeling the long-range intra-temporal dependencies and complex inter-dependencies among the variables in the MTS data for improved multi-horizon forecast accuracy. The explicit relationships among variables are based on prior knowledge provided by human experts in the form of a predefined or explicit graph, while implicit relationships among variables within the MTS data are obtained through neural relational inference methods(\cite{deng2021graph, kipf2018neural}). The implicit relationships are highly-complex and non-linear, can change over time, and uncover hidden relationships unknown to human experts which are not obvious. The existing “human-in-the-loop” STGNNs(\cite{yu2017spatio}, \cite{li2017diffusion}, \cite{guo2020optimized}) incorporate domain-specific knowledge of the relational-structural dependencies among the interdependent variables while simultaneously learning the dynamics from the MTS data. However, arguably, the explicit graph structures in most real-world scenarios are either unknown, inaccurate, or partially available, thus resulting in suboptimal forecasting. Even if available, the explicit graph structure represents a simplified view of dependencies and often fails to capture the non-static spatial-temporal dependencies within the MTS data. Precisely it falls short of accurately inferring the latent time-conditioned underlying relations that drive the co-movements among variables in the substantial MTS data. On the contrary, a recent class of STGNNs(\cite{shang2021discrete, deng2021graph, wu2020connecting, kipf2018neural}) jointly infer the discrete dependency graph structure describing the implicit relations between variables while simultaneously learning the dynamics in MTS data. Despite the success, these approaches neglect to exploit the predefined graph of the inter-relationships among variables obtained from the domain-expertise knowledge resulting in suboptimal performance on the graph time-series forecasting. In addition, implicit graph structure learning from MTS data suffers from inherent limitations of pairwise associations. While in contrast, the relations within the complex dynamical systems of interconnected networks could go beyond pairwise connections. Hypergraph, a generalization of a graph, offers a natural fit for modeling the higher-order structural relations underlying the interconnected networks in complex high-dimensional data. Moreover, the standard STGNNs focus on learning pointwise forecasts but do not provide uncertainty estimates of forecasts. \textcolor{black}{To overcome the challenges, we propose an explicit-implicit knowledge fusion neural network(EIKF-Net) framework with a joint learning paradigm on the explicit-implicit interaction structure for a thorough understanding of the underlying dependencies between time series variables, while simultaneously learning the complex dynamics of the MTS data for better forecast accuracy and to provide reliable uncertainty estimates of forecasts.} The proposed framework consists of two main components: spatial and temporal learning components. We adopt a space-then-time(STT, \cite{gao2022equivalence}) approach, where spatial message-passing schemes are performed prior to the temporal-encoding step. The spatial learning component is further composed of an implicit hypergraph and explicit graph learning modules. The former infers the implicit hypergraph structure, which captures the hierarchical interdependencies among variables in MTS data. Simultaneously it performs hypergraph representation learning schemes to encode the spatio-temporal dynamics underlying the hypergraph-structured MTS data into the latent hypernode-level representations. The latter performs the graph representation learning schemes to encode the pair-wise spatial relations between the multiple co-evolving variables to 
capture the spatio-temporal dynamics within the graph-structured MTS data into the latent node-level representations. We perform convex combination(i.e., ``mix up") of the latent graph and hypergraph representations through a gating mechanism. It leads to more accurate latent representations of the complex non-linear dynamics of the MTS data. The mixup representations allow the framework to capture different types of dependencies that exist at different observation scales(i.e., correlations among variables could potentially differ in the short and long-term views in the MTS data). The temporal learning component focusses on learning the time-evolving dynamics of interdependencies among the variables present in the MTS data to provide accurate multi-horizon forecasts with predictive uncertainty estimates. To summarize, our work presents an end-to-end methodological framework to infer the implicit interaction structure from MTS data. It simultaneously learns the spatio-temporal dynamics within the explicit graph and implicit hypergraph structured MTS data using graph and hypergraph neural networks, respectively, to capture the evolutionary and multi-scale interactions among the variables in the latent representations. It performs inference over these latent representations for downstream MTSF task and models the time-varying uncertainty of the forecasts in order to provide more accurate risk assessment and better decision making by estimating predictive uncertainty. The framework is designed to offer better generalization and scalability for large-scale spatio-temporal MTS data-based forecasting tasks as those found in real-world applications.

\vspace{-5mm}
\section{Problem Definition}
\label{gen_inst} 
\vspace{-4mm}
Lets us assume a historical time series data, with $n$-correlated variables, observed over $\mathrm{T}$ training steps is represented by \thickmuskip=0.15\thickmuskip\resizebox{.165\textwidth}{!}{$\mathbf{X} = \big(\mathbf{x}_{1}, \ldots, \mathbf{x}_{\mathrm{T}}  \big)$}, where the subscript refers to time step. The observations of the $n$-variables at time point t are denoted by \thickmuskip=0.15\thickmuskip\resizebox{.305\textwidth}{!}{$\mathbf{x}_{t} = \big(\mathbf{x}_t^{(1)}, \mathbf{x}_t^{(2)}, \ldots, \mathbf{x}_t^{(n)}\big) \in \mathbb{R}^{(n)}$}, where the superscript refers to variables. Under the rolling-window method for multi-step forecasting, where at the current time step $t$, we predefine a fixed-length look-back window to include the prior $\tau$-steps of historical MTS data to predict for the next $\upsilon$-steps. In the context of MTSF, the learning problem can be formalized using the rolling window method. The goal is to use a historical window of $n$-correlated variables, represented by the \resizebox{.165\textwidth}{!}{$\mathbf{X}_{(t - \tau : \hspace{1mm}t-1)} \in \mathbb{R}^{n \times \tau}$}, which have been observed over previous $\tau$-steps prior to current time step $t$, to predict about the future values of $n$ variables for the next $\upsilon$-steps denoted as \resizebox{.18\textwidth}{!}{$\mathbf{X}_{(t  : t + \upsilon - 1)} \in \mathbb{R}^{n \times \upsilon}$}. \textcolor{black}{The MTSF problem is further formulated on the graph and hypergraph structure to capture the spatial-temporal correlations among multitudinous correlated time series variables}. We represent the historical inputs as continuous-time spatial-temporal graphs, denoted as \resizebox{.25\textwidth}{!}{$\mathcal{G}_{t} = \big(\mathcal{V}, \mathcal{E}, \mathbf{X}_{(t - \tau : \hspace{1mm}t-1)}, \text{A}^{(0)}\big)$}.  \resizebox{.02\textwidth}{!}{$\mathcal{G}_{t}$} is composed of a set of nodes(\resizebox{.015\textwidth}{!}{$\mathcal{V}$}), edges(\resizebox{.015\textwidth}{!}{$\mathcal{E}$}) that describe the connections among the variables and node feature matrix \resizebox{.1\textwidth}{!}{$\mathbf{X}_{(t - \tau : \hspace{1mm}t-1)}$} that changes over time, where $t$ is the current time step. The adjacency matrix, \resizebox{.18\textwidth}{!}{$\text{A}^{(0)} \in \{0,1\}^{|\mathcal{V}| \times |\mathcal{V}|}$}, describes the explicit fixed-graph structure based on prior knowledge of time-series relationships. In addition, we treat historical MTS data as a sequence of dynamic hypergraphs, denoted as \resizebox{.275\textwidth}{!}{$\mathcal{HG}_{t} = \big(\mathcal{HV}, \mathcal{HE}, \mathbf{X}_{(t - \tau : \hspace{1mm}t-1)}, \text{I}\big)$}. The hypergraph is represented by a fixed set of hypernodes(\resizebox{.035\textwidth}{!}{$\mathcal{HV}$}) and hyperedges(\resizebox{.035\textwidth}{!}{$\mathcal{HE}$}), where time series variables denote the hypernodes and hyperedges capture the latent higher-order relationships between hypernodes. The time-varying hypernode feature matrix is given by \resizebox{.1\textwidth}{!}{$\mathbf{X}_{(t - \tau : \hspace{1mm}t-1)}$}. The implicit hypergraph structure is learned through an embedding-based similarity metric learning approach. The incidence matrix, \resizebox{.095\textwidth}{!}{$\mathbf{I} \in \mathbb{R}^{n \times m}$}, describes the hypergraph structure, where $\mathbf{I}_{p, \hspace{0.5mm}q}=1$ if the hyperedge $q$ incident with hypernode $p$ and otherwise 0. The number of hyperedges($\text{m}$) in a hypergraph determines the sparsity of the hypergraph. Given a \resizebox{.0225\textwidth}{!}{$\mathcal{G}_{t}$} and \resizebox{.0425\textwidth}{!}{$\mathcal{HG}_{t}$}, the novel framework is designed to learn a function \resizebox{.05\textwidth}{!}{$F(\theta)$} that maps historical MTS data, \resizebox{.1\textwidth}{!}{$\mathbf{X}_{(t - \tau : \hspace{1mm}t-1)}$}, to their respective future values, \resizebox{.095\textwidth}{!}{$\mathbf{X}_{(t  : t + \upsilon - 1)}$} values defined as follows,

\vspace{-8mm}
\resizebox{0.925\linewidth}{!}{
\begin{minipage}{\linewidth}
\begin{align}
\left[\mathbf{X}_{(t - \tau)}, \cdots, \mathbf{X}_{(t-1)} ; \mathcal{G}_{t}, \mathcal{HG}_{t}\right] \stackrel{F(\theta)}{\longrightarrow}\left[\mathbf{X}_{(t + 1)}, \cdots, \mathbf{X}_{(t + \upsilon-1)}\right]
\end{align}
\end{minipage}
}

\vspace{-1mm}
Simply, the MTSF task formulated on the explicit graph(\resizebox{.0225\textwidth}{!}{$\mathcal{G}_{t}$}) and implicit hypergraph(\resizebox{.04\textwidth}{!}{$\mathcal{HG}_{t}$}) is described as follows:

\vspace{-7mm}
\resizebox{0.925\linewidth}{!}{
\begin{minipage}{\linewidth}
\begin{align}
\min _{\theta} \mathcal{L}\big(\mathbf{X}_{(t  : t + \upsilon-1)}, \hat{\mathbf{X}}_{(t  : t + \upsilon-1)} ; \mathbf{X}_{(t - \tau : \hspace{1mm}t-1)}, \mathcal{G}_{t}, \mathcal{HG}_{t}\big)
\end{align}
\end{minipage}
}

\vspace{-3mm}
where $\theta$ denotes all the learnable parameters for trainable function \resizebox{.05\textwidth}{!}{$F(\theta)$}.  \resizebox{.095\textwidth}{!}{$\hat{\mathbf{X}}_{(t  : t + \upsilon-1)}$} denotes the model predictions, $\mathcal{L}$ is the loss function. The loss function to train our learning algorithm is mean absolute error(MAE) loss.

\vspace{-8mm}
\resizebox{0.925\linewidth}{!}{
\begin{minipage}{\linewidth}
\begin{align}
\mathcal{L}_{\text{MAE}}\left(\theta\right)=\frac{1}{\upsilon}\left|\mathbf{X}_{(t  : t + \upsilon-1)}-\hat{\mathbf{X}}_{(t  : t + \upsilon-1)}\right|
\end{align}
\end{minipage}
}

\vspace{-4mm}
\begin{figure}[ht!]
\center     
\includegraphics[keepaspectratio,height=5cm, width=10cm,trim=0cm 5.8cm 0cm 5.45cm,clip]{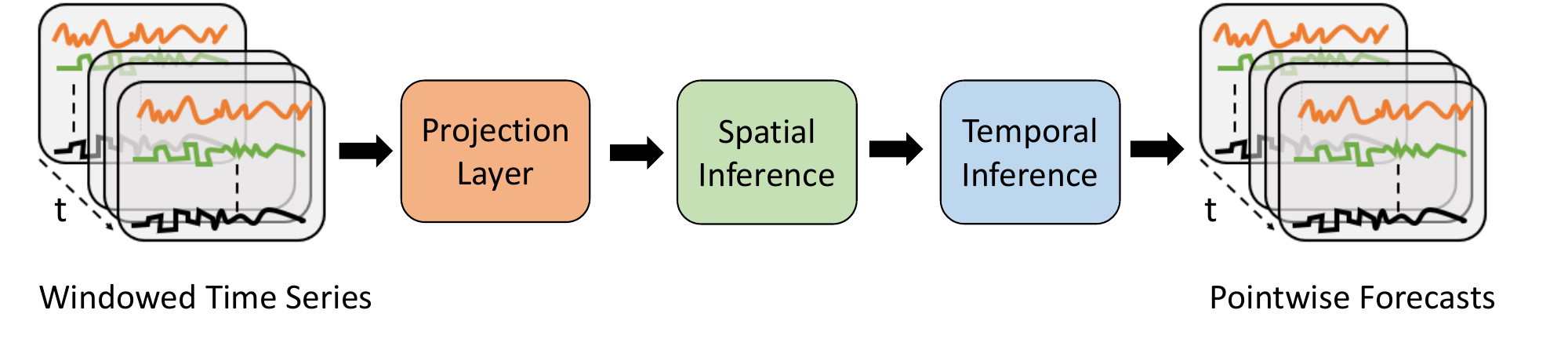} 
\vspace{-6mm}
\caption{Overview of \textbf{EIKF-Net} framework}
\label{fig:overallarch}
\end{figure}
 
\vspace{-8mm}
\section{Our Approach} 
\label{gen_inst}
\vspace{-4mm}
The overall neural forecasting architecture of our framework is illustrated in Figure \ref{fig:overallarch}. It consists of three main components: The projection layer, spatial learning, and temporal learning components. The spatial learning component includes two modules: graph and hypergraph learning modules. The hypergraph learning module infers the discrete dependency hypergraph structure to capture the interrelations between time-series variables. It also performs higher-order message-passing schemes to learn the time-conditioned optimal hypernode-level representations by modeling the hypergraph-structured MTS data. The graph learning module utilizes the predefined graph, which represents the relational structure of the variables obtained from domain-expertise knowledge to obtain the graph-structured MTS data. It performs spatial graph-filtering through neighborhood aggregation schemes to compute the optimal node-level representations that better capture the underlying dynamics within the MTS data. The temporal inference component performs a convex combination of the latent explicit-graph and  implicit-hypergraph representations and learns the time-evolving interdependencies to provide pointwise forecasts and uncertainty estimations. Overall, joint optimization of different learning components of the proposed framework effectively captures the complex relationships between time-series variables and makes accurate forecasts.

\vspace{-4mm}
\subsection{Projection Layer}
\vspace{-3mm}
The projection layer utilizes a gated linear networks(GLN, \cite{dauphin2017language}) to learn the non-linear representations of the input data, \resizebox{.16\textwidth}{!}{$\mathbf{X}_{(t - \tau : \hspace{1mm}t-1)} \in \mathbb{R}^{n \times \tau}$} through a gating mechanism to compute a transformed feature matrix, \resizebox{.16\textwidth}{!}{$\bar{\mathbf{X}}_{(t - \tau : \hspace{1mm}t-1)} \in \mathbb{R}^{n \times d}$} as follows, 

\vspace{-5mm}
\resizebox{0.935\linewidth}{!}{
\begin{minipage}{\linewidth}
\begin{align}
\centering
\bar{\mathbf{X}}_{(t - \tau : \hspace{1mm}t-1)}   = \big( \sigma(\text{W}_{0}\mathbf{X}_{(t - \tau : \hspace{1mm}t-1)}) \otimes \text{W}_{1}\mathbf{X}_{(t - \tau : \hspace{1mm}t-1)}\big)\text{W}_{2} \nonumber
\end{align}
\end{minipage}
}

\vspace{-2mm}
where \resizebox{.195\textwidth}{!}{$\text{W}_{0}, \text{W}_{1}, \text{W}_{2} \in \mathbb{R}^{\tau \times d}$} are trainable weight matrices, $\otimes$ denotes the element-wise multiplication. $\sigma$ is the non-linear activation function.

\vspace{-5mm}
\subsection{Spatial-Inference}  
\begin{figure}[ht!]
\center     
\hspace{5mm}\includegraphics[keepaspectratio,height=5.25cm, width=11cm,trim=0cm 2.5cm 0cm 2.15cm,clip]{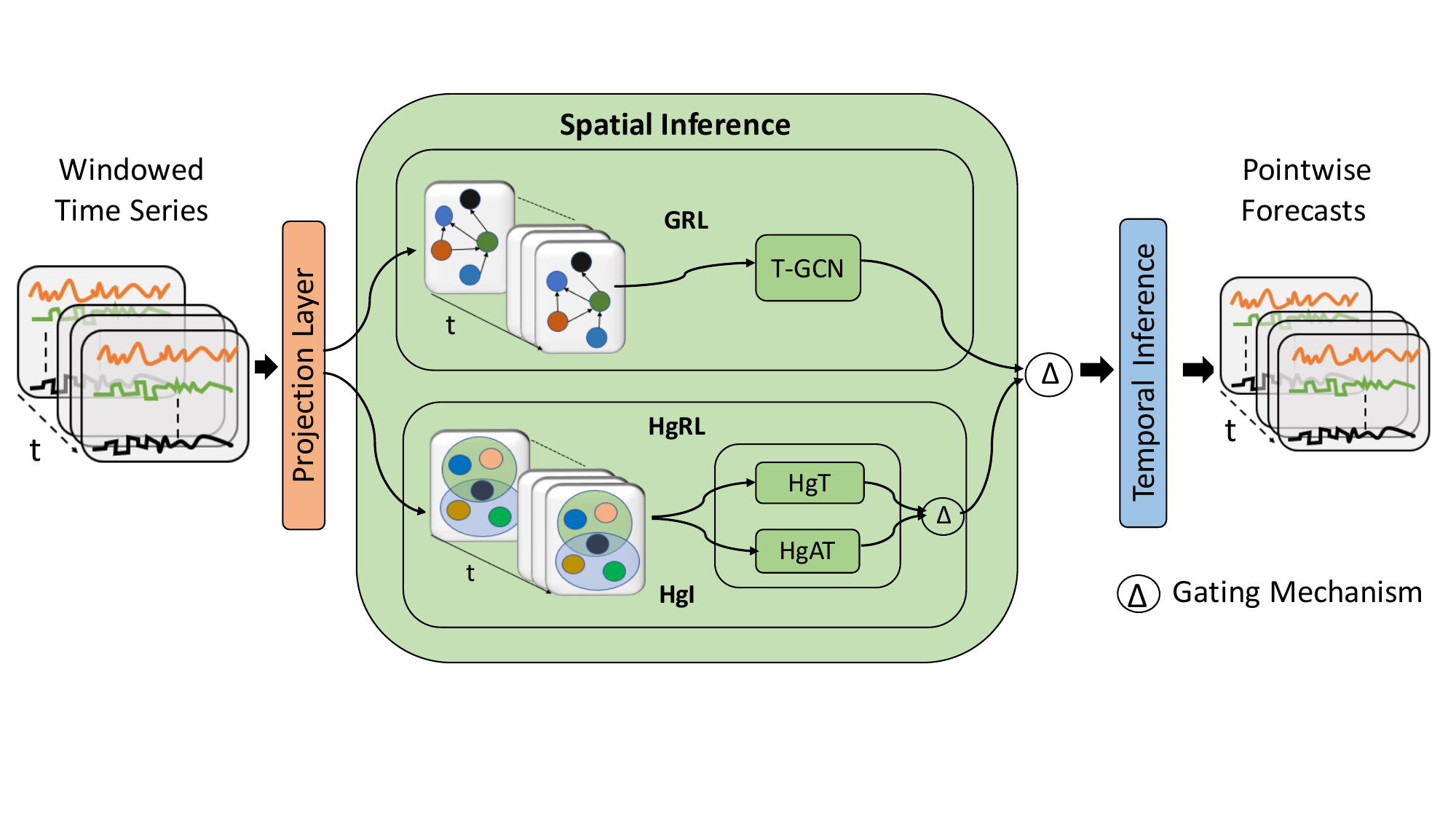} 
\vspace{-6mm}
\caption{Overview of spatial inference component.}
\label{fig:spatialarch}
\end{figure}

\vspace{-3mm}
The spatial inference component of our framework is illustrated in Figure \ref{fig:spatialarch}. The spatial-learning component encodes non-linear input data, \resizebox{.085\textwidth}{!}{$\bar{\mathbf{X}}_{(t - \tau : \hspace{1mm}t-1)}$} to obtain graph and hypergraph representations using two modules: the hypergraph learning module and the graph learning module. The hypergraph learning module performs joint hypergraph inference and representation learning, while the graph learning module performs graph representation learning. The outputs of these two modules are fused using a convex combination approach to regulate the flow of information encoded by each module. The details of each module are discussed in subsequent sections.

\vspace{-4mm}
\subsubsection{Hypergraph Inference and Representation Learning}
\vspace{-3mm}
The hypergraph learning module is composed of two units: the hypergraph inference(HgI) unit and the hypergraph representation learning(HgRL) unit. The HgI unit is a structural modeling approach that aims to infer the discrete hypergraph topology capturing the hierarchical interdependence relations among time-series variables for a hypergraph-structured representation of the MTS data. The HgI unit utilizes a similarity metric learning method to implicitly learn the task-relevant relational hypergraph structure from the hypergraph embeddings. The hypernodes and hyperedges of the hypergraph are represented by differentiable embeddings $\mathbf{z_{i}}$ and $\mathbf{z_{j}}$, respectively, where $1 \leq i \leq n$ and $1 \leq j \leq m$. The embeddings $\mathbf{z_{i}}$ and $\mathbf{z_{j}}$ capture the global-contextual behavioral patterns of hypernodes and hyperedges, respectively. These embeddings $\mathbf{z_{i}}, \mathbf{z_{j}} \in \mathbb{R}^{d}$ are continuous vector representations in the $d$-dimensional vector space, which allows the HgI unit to adapt and update as it processes new information. We compute the pairwise similarity of any pair $\mathbf{z_{i}}$ and $\mathbf{z_{j}}$ as follows,

\vspace{-5mm}
\resizebox{0.90\linewidth}{!}{
\begin{minipage}{\linewidth}
\begin{align}
\text{P}_{i,j} = \sigma \big([\text{S}_{i,j} || 1- \text{S}_{i,j}]\big); \text{S}_{i,j} = \frac{\mathbf{z^{T}_{i}} \mathbf{z_{j}} + 1}{2\left\|\mathbf{z_{i}}\right\| \cdot\left\|\mathbf{z_{j}}\right\|}  
\end{align}
\end{minipage}
}

\vspace{-2mm}
where \resizebox{.00985\textwidth}{!}{$\Vert$} denotes vector concatenation. We apply the sigmoid activation function to map the pairwise scores, \resizebox{.035\textwidth}{!}{$\text{P}_{i,j}$} into the range \resizebox{.0475\textwidth}{!}{[0, 1]}. \resizebox{.12\textwidth}{!}{$\text{P}^{(k)}_{i,j} \in \mathbb{R}^{nm \times 2}$} denote the hyperedge probability over hypernodes of the hypergraph, where \resizebox{.085\textwidth}{!}{$\text{k} \in \{0,1\}$}. \resizebox{.035\textwidth}{!}{$\text{P}^{(k)}_{i,j}$} encodes the relation between a pair of hypernodes and hyperedges \resizebox{.045\textwidth}{!}{$\left(i, j\right)$} to a scalar \resizebox{.06\textwidth}{!}{$\in[0,1]$}. \resizebox{.035\textwidth}{!}{$\text{P}^{(0)}_{i,j}$} represents the probability of a hypernode $i$ connected to the hyperedge $j$, and \resizebox{.035\textwidth}{!}{$\text{P}^{(1)}_{i,j}$} represents contrariwise probability. \textcolor{black}{We utilize the Gumbel-softmax trick(\cite{jang2016categorical, maddison2016concrete}) to sample a discrete hypergraph structure described by an incidence matrix, \resizebox{.08\textwidth}{!}{$\mathbf{I} \in \mathbb{R}^{n \times m}$}. This allows for a differentiable way to sample the latent structure from the hyperedge probability distribution \resizebox{.035\textwidth}{!}{$\text{P}_{i,j}$}}. Thus by utilizing the Gumbel-softmax trick, the hypergraph structure can be learned in an end-to-end differentiable way, allowing for the use of gradient-based optimization methods to train the model using an inductive-learning approach. It is described as,

\vspace{-5mm}
\resizebox{0.90\linewidth}{!}{
\begin{minipage}{\linewidth}
\begin{align}
\mathbf{I}_{i,j} =\exp \big(\big(g^{(k)}_{i,j} + \text{P}^{(k)}_{i,j}\big) / \gamma\big)\big/{\sum \exp \big(\big(g^{(k)}_{i,j} + \text{P}^{(k)}_{i,j}\big) / \gamma\big)}
\end{align}
\end{minipage}
}

\vspace{-2mm}
where \resizebox{.375\textwidth}{!}{$g^{(k)}_{i j} \sim \operatorname{Gumbel}(0,1) = \log (-\log (\text{U}(0,1))$} where \resizebox{.015\textwidth}{!}{$\text{U}$} is uniform distribution. $\gamma$, denotes the temperature parameter with 0.05. We regularize the learned hypergraph to be sparse by optimizing the probabilistic hypergraph distribution parameters to drop the redundant hyperedges over hypernodes. The downstream forecasting task acts as the indirect supervisory information for revealing the high-order structure,i.e., the hypergraph relation structure behind the observed data. The MTS data is represented as hypernode-attributed spatio-temporal hypergraphs. \textcolor{black}{A hypergraph representation learning unit(HgRL) is used to compute optimal hypernode-level representations by capturing the spatio-temporal dynamics within the hypergraph-structured MTS data. These representations are then used to perform inference on the downstream multi-horizon forecasting task}. In short, HgRL is a neural network architecture that utilizes both Hypergraph Attention Network(HgAT) and Hypergraph Transformer(HgT) as its backbone. The HgT uses multi-head self-attention mechanisms to learn the latent hypergraph representations, $\mathbf{h^{\prime}_{i}}^{(t)}$, at time $t$ without leveraging any prior knowledge about the structure of the hypergraph. The HgAT performs higher-order message-passing schemes on the hypergraph to aggregate information and compute the latent hypernode representations, $\mathbf{h}^{(t)}_{i}$, at time $t$. The HgAT and HgT form a powerful backbone for HgRL, allowing it to effectively learn hypergraph representations by capturing complex relationships and dependencies within the hypergraph-structured MTS data. We provide the implementation details and a more in-depth explanation in the appendix for further information. We regulate the information flow from $\mathbf{h^{\prime}_{i}}^{(t)}$  and $\mathbf{h}^{(t)}_{i}$ by applying a gating mechanism to produce a weighted combination of representations as described below,

\vspace{-7mm}
\resizebox{0.95\linewidth}{!}{
\hspace{5mm}\begin{minipage}{\linewidth}
\begin{align}
\mathbf{h^{\prime\prime}_{i}}^{(t)}  = \sigma \big( g^{\prime}(\mathbf{h^{\prime}_{i}}^{(t)})) + (1-g^{\prime})(\mathbf{h}^{(t)}_{i}) \big); g^{\prime} = \sigma \big( f^{\prime}_s(\mathbf{h^{\prime}_{i}}^{(t)}) + f^{\prime}_g(\mathbf{h}^{(t)}_{i}) \big) 
\end{align}
\end{minipage}
} 

\vspace{-1mm}
where $f^{\prime}_s$ and $f^{\prime}_g$ are linear projections. Fusing representations can be beneficial for modeling the multi-scale interactions underlying the spatio-temporal hypergraph data and also help to mitigate overfitting. The spatio-temporal data often contains correlations between variables that change over time or at different observation scales. By fusing representations, the proposed framework incorporates the most relevant information to capture the time-evolving underlying patterns in the MTS data, which leads to more accurate and robust forecasts. In brief, the hypergraph learning module optimizes the discrete hypergraph structure through the similarity metric learning technique. It formulates the posterior forecasting task as message-passing schemes with hypergraph neural networks to learn the optimal hypergraph representations, which leads to more accurate and expressive representations of the MTS data for better forecast accuracy. 

\vspace{-5mm}
\subsubsection{Graph Representation Learning(GRL)}
\vspace{-3mm}
We represent the MTS data as continuous-time spatio-temporal graphs based on predefined graphs obtained from domain-specific knowledge. We utilize the Temporal Graph Convolution Neural Network(T-GCN, \cite{zhao2019t}) to compute optimal node-level representations by modeling the graph topology dependencies and feature attributes within the graph-structured MTS data. These graph representations are further processed by the downstream temporal inference component for learning the non-linear temporal dynamics of inter-series correlations among the variables. In short, the T-GCN performs neighborhood aggregation schemes on predefined graph topology to compute the optimal node-level representations, \resizebox{.055\textwidth}{!}{$\mathbf{h^{\prime\prime\prime}_{i}}^{(t)}$}, at a specific time $t$. It effectively captures fine-grained, data-source specific patterns accurately. We discuss in the appendix a more detailed description of the technique.

\vspace{-5mm}

\vspace{-6mm}
\subsection{Temporal-Inference}
\vspace{-3mm}
The mixture-of-experts(MOE) mechanism combines the predictions of multiple subnetworks (experts) to produce a final prediction. In this specific framework, the experts are graph and hypergraph learning modules. The expert predictions are combined through a gating mechanism in an input-dependent manner by calculating a weighted sum of their predictions. The goal of training in this framework is to achieve two objectives: 1) Identifying the optimal distribution of weights for the gating function that precisely captures the underlying distribution within the MTS data, and 2) Training the experts using the distribution weights specified by the gating function. The fused representations are obtained by combining the predictions of the experts using the calculated weights as follows,

\vspace{-4mm}
\resizebox{0.935\linewidth}{!}{
\begin{minipage}{\linewidth}
\begin{align}
\mathbf{h^{\prime\prime\prime\prime}_{i}}^{(t)}  = \sigma \big( g^{\prime\prime}(\mathbf{h^{\prime\prime}_{i}}^{(t)})) + (1-g^{\prime\prime})(\mathbf{h^{\prime\prime\prime}_{i}}^{(t)}) \big); g^{\prime\prime} = \sigma \big( f^{\prime\prime}_s(\mathbf{h^{\prime\prime}_{i}}^{(t)}) + f^{\prime\prime}_g(\mathbf{h^{\prime\prime\prime}_{i}}^{(t)}) \big) 
\end{align}
\end{minipage}
} 

\vspace{-1mm}
where \resizebox{.11\textwidth}{!}{$\mathbf{h^{\prime\prime}_{i}}^{(t)}$, $\mathbf{h^{\prime\prime\prime}_{i}}^{(t)}$} are computed by the graph and hypergraph learning modules, respectively. \resizebox{.025\textwidth}{!}{$f^{\prime\prime}_s$} and \resizebox{.025\textwidth}{!}{$f^{\prime\prime}_g$} are linear projections. The temporal inference component consists of a stack of \resizebox{.05\textwidth}{!}{$1 \times 1$} convolutions. The fused representations are fed as input to the temporal inference component. This component aims to model the non-linear temporal dynamics of inter-series dependencies among variables underlying the spatial-temporal MTS data and predicts the pointwise forecasts, \resizebox{.1\textwidth}{!}{$\hat{\mathbf{X}}_{(t  : t + \upsilon-1)}$}. Our proposed framework uses the spatial-then-time modeling approach to learn the higher-order structure representation and dynamics in MTS data. This approach first encodes the spatial information of the relational structure, including both explicit graph and implicit hypergraph, which captures the complex spatial dependencies. By incorporating the temporal learning component, the framework then analyzes the evolution of these dependencies over time, which helps to improve the interpretability and generalization of the framework. This approach is beneficial for dealing with real-world applications that involve entangled complex spatial-temporal dependencies within the MTS data, which can be challenging to model using traditional methods. Additionally, by minimizing the negative Gaussian log likelihood, the uncertainty estimates of forecasts can be provided through the \textbf{w/Unc- EIKF-Net} framework(i.e., \textbf{EIKF-Net} with Local Uncertainty Estimation).  Minimizing the negative Gaussian log likelihood is equivalent to maximizing the likelihood of the model's predictions given the true values. This allows the framework to provide more accurate and reliable uncertainty estimates of forecasts. In summary, our proposed methods(\textbf{EIKF-Net}, \textbf{w/Unc- EIKF-Net}) allow for simultaneously modeling the latent interdependencies and then analyze the evolution of these dependencies over time in the sensor network based dynamical systems in an end-to-end manner. 

\vspace{-5mm} 
\section{Datasets}
\vspace{-4mm}
We conduct experiments to verify the performance of proposed models(\textbf{EIKF-Net}, \textbf{w/Unc- EIKF-Net}) on large-scale spatial-temporal datasets. The real-world traffic datasets(PeMSD3, PeMSD4, PeMSD7, PeMSD7(M), and PeMSD8) were collected by the Caltrans Performance Measurement System(PeMS, \cite{chen2001freeway}), which measures the traffic flow in real-time. We preprocess all the datasets by aggregating the 30-second intervals data to 5-minute average intervals data(\cite{choi2022graph}) to ensure consistency and fair comparison with previous works. In addition, we also utilize publicly available traffic flow prediction datasets(METR-LA, PEMS-BAY) presented by \cite{LiYS018}. We aggregate the datasets into 5-minute average intervals data which leads to 288 observations per day. This helps to demonstrate the effectiveness and benefits of the proposed methodological framework for analyzing and modeling complex spatio-temporal MTS data over existing approaches. Additional details about the benchmark datasets are described in the appendix.

\vspace{-3mm}
\begin{table}[ht!]
\setlength{\tabcolsep}{0.2em} 
\renewcommand\arraystretch{1.14} 
\centering
 \resizebox{1.035\textwidth}{!}{
\hspace{-5mm}\begin{tabular}{c|ccc|ccc|ccc|ccc|ccc}
\hline
\multirow{2}{*}{\textbf{Methods}} & \multicolumn{3}{c|}{\textbf{PeMSD3}} & \multicolumn{3}{c|}{\textbf{PeMSD4}} & \multicolumn{3}{c|}{\textbf{PeMSD7}} & \multicolumn{3}{c|}{\textbf{PeMSD8}} & \multicolumn{3}{c}{\textbf{PeMSD7(M)}} \\ \cline{2-16} 
 & \multicolumn{1}{l}{\textbf{MAE}} & \multicolumn{1}{l}{\textbf{RMSE}} & \multicolumn{1}{l|}{\textbf{MAPE}} & \multicolumn{1}{l}{\textbf{MAE}} & \multicolumn{1}{l}{\textbf{RMSE}} & \multicolumn{1}{l|}{\textbf{MAPE}} & \multicolumn{1}{l}{\textbf{MAE}} & \multicolumn{1}{l}{\textbf{RMSE}} & \multicolumn{1}{l|}{\textbf{MAPE}} & \multicolumn{1}{l}{\textbf{MAE}} & \multicolumn{1}{l}{\textbf{RMSE}} & \multicolumn{1}{l|}{\textbf{MAPE}} & \textbf{MAE} & \textbf{RMSE} & \textbf{MAPE} \\ \hline
HA & 31.58 & 52.39 & 33.78 & 38.03 & 59.24 & 27.88 & 45.12 & 65.64 & 24.51 & 34.86 & 59.24 & 27.88 & 4.59 & 8.63 & 14.35 \\
ARIMA & 35.41 & 47.59 & 33.78 & 33.73 & 48.80 & 24.18 & 38.17 & 59.27 & 19.46 & 31.09 & 44.32 & 22.73 & 7.27 & 13.20 & 15.38 \\
VAR & 23.65 & 38.26 & 24.51 & 24.54 & 38.61 & 17.24 & 50.22 & 75.63 & 32.22 & 19.19 & 29.81 & 13.10 & 4.25 & 7.61 & 10.28 \\
FC-LSTM & 21.33 & 35.11 & 23.33 & 26.77 & 40.65 & 18.23 & 29.98 & 45.94 & 13.20 & 23.09 & 35.17 & 14.99 & 4.16 & 7.51 & 10.10 \\
TCN & 19.32 & 33.55 & 19.93 & 23.22 & 37.26 & 15.59 & 32.72 & 42.23 & 14.26 & 22.72 & 35.79 & 14.03 & 4.36 & 7.20 & 9.71 \\
TCN(w/o causal) & 18.87 & 32.24 & 18.63 & 22.81 & 36.87 & 14.31 & 30.53 & 41.02 & 13.88 & 21.42 & 34.03 & 13.09 & 4.43 & 7.53 & 9.44 \\
GRU-ED & 19.12 & 32.85 & 19.31 & 23.68 & 39.27 & 16.44 & 27.66 & 43.49 & 12.20 & 22.00 & 36.22 & 13.33 & 4.78 & 9.05 & 12.66 \\
DSANet & 21.29 & 34.55 & 23.21 & 22.79 & 35.77 & 16.03 & 31.36 & 49.11 & 14.43 & 17.14 & 26.96 & 11.32 & 3.52 & 6.98 & 8.78 \\
STGCN & 17.55 & 30.42 & 17.34 & 21.16 & 34.89 & 13.83 & 25.33 & 39.34 & 11.21 & 17.50 & 27.09 & 11.29 & 3.86 & 6.79 & 10.06 \\
DCRNN & 17.99 & 30.31 & 18.34 & 21.22 & 33.44 & 14.17 & 25.22 & 38.61 & 11.82 & 16.82 & 26.36 & 10.92 & 3.83 & 7.18 & 9.81 \\
GraphWaveNet & 19.12 & 32.77 & 18.89 & 24.89 & 39.66 & 17.29 & 26.39 & 41.50 & 11.97 & 18.28 & 30.05 & 12.15 & 3.19 & 6.24 & 8.02 \\
ASTGCN(r) & 17.34 & 29.56 & 17.21 & 22.93 & 35.22 & 16.56 & 24.01 & 37.87 & 10.73 & 18.25 & 28.06 & 11.64 & 3.14 & 6.18 & 8.12 \\
MSTGCN & 19.54 & 31.93 & 23.86 & 23.96 & 37.21 & 14.33 & 29.00 & 43.73 & 14.30 & 19.00 & 29.15 & 12.38 & 3.54 & 6.14 & 9.00 \\
STG2Seq & 19.03 & 29.83 & 21.55 & 25.20 & 38.48 & 18.77 & 32.77 & 47.16 & 20.16 & 20.17 & 30.71 & 17.32 & 3.48 & 6.51 & 8.95 \\
LSGCN & 17.94 & 29.85 & 16.98 & 21.53 & 33.86 & 13.18 & 27.31 & 41.46 & 11.98 & 17.73 & 26.76 & 11.20 & 3.05 & 5.98 & 7.62 \\
STSGCN & 17.48 & 29.21 & 16.78 & 21.19 & 33.65 & 13.90 & 24.26 & 39.03 & 10.21 & 17.13 & 26.80 & 10.96 & 3.01 & 5.93 & 7.55 \\
AGCRN & 15.98 & 28.25 & 15.23 & 19.83 & 32.26 & 12.97 & 22.37 & 36.55 & 9.12 & 15.95 & 25.22 & 10.09 & 2.79 & 5.54 & 7.02 \\
STFGNN & 16.77 & 28.34 & 16.30 & 20.48 & 32.51 & 16.77 & 23.46 & 36.60 & 9.21 & 16.94 & 26.25 & 10.60 & 2.90 & 5.79 & 7.23 \\
STGODE & 16.50 & 27.84 & 16.69 & 20.84 & 32.82 & 13.77 & 22.59 & 37.54 & 10.14 & 16.81 & 25.97 & 10.62 & 2.97 & 5.66 & 7.36 \\
Z-GCNETs & 16.64 & 28.15 & 16.39 & 19.50 & 31.61 & 12.78 & 21.77 & 35.17 & 9.25 & 15.76 & 25.11 & 10.01 & 2.75 & 5.62 & 6.89 \\
STG-NCDE & 15.57 & 27.09 & 15.06 & 19.21 & 31.09 & 12.76 & 20.53 & 33.84 & 8.80 & 15.45 & 24.81 & 9.92 & 2.68 & 5.39 & 6.76 \\ \hline
\textbf{EIKF-Net} & \textbf{13.67} & \textbf{20.17} & \textbf{10.83} & \textbf{18.36} & \textbf{26.48} & \textbf{10.05} & \textbf{19.86} & \textbf{30.85} & \textbf{8.59} & \textbf{14.93} & \textbf{22.17} & \textbf{7.83} & \textbf{2.44} & \textbf{4.91} & \textbf{6.03} \\
\textbf{w/Unc- EIKF-Net} & 13.97 & 20.59 & 11.04 & 18.73 & 27.18 & 10.26 & \textbf{-} & \textbf{-} & \textbf{-} & 15.12 & 22.84 & 8.19 & 2.54 & 5.09 & 6.24 \\ \hline
\end{tabular}
}
\vspace{-3mm}
\caption{Pointwise forecast and predictive uncertainty errors on benchmark datasets at horizon$\textbf{@}$12. \enquote{\textbf{-}} indicates an OOM(Out Of Memory) error.}
\label{tab:results1}
\end{table}

\vspace{-8mm} 
\section{Experimental Results}
\vspace{-4mm} 
Table \ref{tab:results1} presents a comparison of the forecast errors of proposed models(\textbf{EIKF-Net} and \textbf{w/Unc- EIKF-Net}) with those of several baseline models on five different datasets(PeMSD3, PeMSD4, PeMSD7, PeMSD7M, and PeMSD8). The forecast errors are evaluated for a 12($\tau$)-step-prior to 12($\upsilon$)-step-ahead forecasting task which is a popular and well-established benchmark in the MTSF task. The performance of the proposed models was evaluated using multiple metrics such as mean absolute error(MAE), root mean squared error(RMSE), and mean absolute percentage error(MAPE). Using multiple evaluation metrics in multi-horizon prediction tasks provides a comprehensive evaluation of the proposed models performance with the baselines. The results for the baseline models were reported in a previous study by \cite{choi2022graph}. The proposed models(\textbf{EIKF-Net}, \textbf{w/Unc- EIKF-Net}) consistently demonstrate state-of-the-art performance compared to baseline models on various benchmark datasets based on multiple evaluation metrics. The results show that the proposed models demonstrate the best performance with lower forecast error on benchmark datasets. Specifically, they report a $12.2\%$, $14.8\%$, $8.8\%$, $10.6\%$ and $8.9\%$ significant drop in the RMSE metric compared to the next-best baseline models on PeMSD3, PeMSD4, PeMSD7, PeMSD8, and PeMSD7(M) datasets, respectively. In addition to the pointwise forecasts, the \textbf{w/Unc- EIKF-Net} model(i.e., \textbf{EIKF-Net} with local uncertainty estimation) provides time-varying uncertainty estimates. Its performance is slightly worse than the \textbf{EIKF-Net} model but still outperforms several strong baselines in the literature, as reflected in the lower prediction error. In brief, the empirical results show the efficacy of the proposed neural forecasting architecture in modeling the complex and nonlinear spatio-temporal dynamics underlying the MTS data to provide better forecasts. The appendix provides more detailed information on the experimental setup, ablation studies, and other additional experimental results on multi-horizon forecasting. Moreover, the appendix discusses the experimental results that support the \textbf{EIKF-Net} framework's ability to handle missing data and provide more insights into the \textbf{w/Unc- EIKF-Net} framework for estimating uncertainty.  Furthermore, the appendix also includes time series visualizations of model predictions with the uncertainty estimates compared to the ground truth. 

\vspace{-6mm}
\section{Conclusion}
\vspace{-4mm} 
We propose a framework that combines implicit and explicit knowledge for learning the dynamics of MTS data to provide accurate multi-horizon forecasts. The experimental results on real-world datasets demonstrate the effectiveness of the proposed framework and have shown improved forecast estimates and reliable uncertainty estimations. For future work, we would endeavor to generalize the framework to handle much larger scale graph datasets for forecasting, synthetic-private data generation, missing-data imputation etc.

\vspace{-3mm}
\bibliography{iclr2023_conference}

\begin{thebibliography}{68}
\providecommand{\natexlab}[1]{#1}
\providecommand{\url}[1]{\texttt{#1}}
\expandafter\ifx\csname urlstyle\endcsname\relax
  \providecommand{\doi}[1]{doi: #1}\else
  \providecommand{\doi}{doi: \begingroup \urlstyle{rm}\Url}\fi

\bibitem[Ba et~al.(2016)Ba, Kiros, and Hinton]{ba2016layer}
Jimmy~Lei Ba, Jamie~Ryan Kiros, and Geoffrey~E Hinton.
\newblock Layer normalization.
\newblock \emph{arXiv preprint arXiv:1607.06450}, 2016.

\bibitem[Bai et~al.(2019)Bai, Yao, Kanhere, Wang, and Sheng]{bai2019STG2Seq}
Lei Bai, Lina Yao, Salil~S. Kanhere, Xianzhi Wang, and Quan~Z. Sheng.
\newblock Stg2seq: Spatial-temporal graph to sequence model for multi-step
  passenger demand forecasting.
\newblock In \emph{IJCAI}, 7 2019.
\newblock \doi{10.24963/ijcai.2019/274}.
\newblock URL \url{https://doi.org/10.24963/ijcai.2019/274}.

\bibitem[Bai et~al.(2020{\natexlab{a}})Bai, Yao, Li, Wang, and
  Wang]{Bai2020nips}
Lei Bai, Lina Yao, Can Li, Xianzhi Wang, and Can Wang.
\newblock Adaptive graph convolutional recurrent network for traffic
  forecasting.
\newblock In \emph{NeurIPS}, 2020{\natexlab{a}}.

\bibitem[Bai et~al.(2020{\natexlab{b}})Bai, Yao, Li, Wang, and
  Wang]{NEURIPS2020_ce1aad92}
Lei Bai, Lina Yao, Can Li, Xianzhi Wang, and Can Wang.
\newblock Adaptive graph convolutional recurrent network for traffic
  forecasting.
\newblock In \emph{NeurIPS}, volume~33, pp.\  17804--17815, 2020{\natexlab{b}}.

\bibitem[Bai et~al.(2018)Bai, Kolter, and Koltun]{BaiTCN2018}
Shaojie Bai, J.~Zico Kolter, and Vladlen Koltun.
\newblock An empirical evaluation of generic convolutional and recurrent
  networks for sequence modeling.
\newblock \emph{arXiv:1803.01271}, 2018.

\bibitem[Brody et~al.(2021)Brody, Alon, and Yahav]{brody2021attentive}
Shaked Brody, Uri Alon, and Eran Yahav.
\newblock How attentive are graph attention networks?
\newblock \emph{arXiv preprint arXiv:2105.14491}, 2021.

\bibitem[Chen et~al.(2001)Chen, Petty, Skabardonis, Varaiya, and
  Jia]{chen2001freeway}
Chao Chen, Karl Petty, Alexander Skabardonis, Pravin Varaiya, and Zhanfeng Jia.
\newblock Freeway performance measurement system: mining loop detector data.
\newblock \emph{Transportation Research Record}, 1748\penalty0 (1):\penalty0
  96--102, 2001.

\bibitem[Chen et~al.(2019)Chen, Wu, and Zaki]{chen2019deep}
Yu~Chen, Lingfei Wu, and Mohammed~J Zaki.
\newblock Deep iterative and adaptive learning for graph neural networks.
\newblock \emph{arXiv preprint arXiv:1912.07832}, 2019.

\bibitem[Chen et~al.(2021)Chen, Segovia-Dominguez, and Gel]{chen2021ZGCNET}
Yuzhou Chen, Ignacio Segovia-Dominguez, and Yulia~R Gel.
\newblock Z-gcnets: Time zigzags at graph convolutional networks for time
  series forecasting.
\newblock In \emph{ICML}, 2021.

\bibitem[Chen et~al.(2022)Chen, Segovia-Dominguez, Coskunuzer, and
  Gel]{tampsgcnets2022}
Yuzhou Chen, Ignacio Segovia-Dominguez, Baris Coskunuzer, and Yulia Gel.
\newblock {TAMP}-s2{GCN}ets: Coupling time-aware multipersistence knowledge
  representation with spatio-supra graph convolutional networks for time-series
  forecasting.
\newblock In \emph{International Conference on Learning Representations}, 2022.

\bibitem[Cho et~al.(2014{\natexlab{a}})Cho, {van Merrienboer}, Gulcehre,
  Bougares, Schwenk, and Bengio]{cho2014grued}
Kyunghyun Cho, B~{van Merrienboer}, Caglar Gulcehre, F~Bougares, H~Schwenk, and
  Yoshua Bengio.
\newblock Learning phrase representations using rnn encoder-decoder for
  statistical machine translation.
\newblock In \emph{EMNLP}, 2014{\natexlab{a}}.

\bibitem[Cho et~al.(2014{\natexlab{b}})Cho, Van~Merri{\"e}nboer, Gulcehre,
  Bahdanau, Bougares, Schwenk, and Bengio]{cho2014learning}
Kyunghyun Cho, Bart Van~Merri{\"e}nboer, Caglar Gulcehre, Dzmitry Bahdanau,
  Fethi Bougares, Holger Schwenk, and Yoshua Bengio.
\newblock Learning phrase representations using rnn encoder-decoder for
  statistical machine translation.
\newblock \emph{arXiv preprint arXiv:1406.1078}, 2014{\natexlab{b}}.

\bibitem[Choi et~al.(2022)Choi, Choi, Hwang, and Park]{choi2022graph}
Jeongwhan Choi, Hwangyong Choi, Jeehyun Hwang, and Noseong Park.
\newblock Graph neural controlled differential equations for traffic
  forecasting.
\newblock In \emph{Proceedings of the AAAI Conference on Artificial
  Intelligence}, volume~36, pp.\  6367--6374, 2022.

\bibitem[Cini et~al.(2021)Cini, Marisca, and Alippi]{cini2021multivariate}
Andrea Cini, Ivan Marisca, and Cesare Alippi.
\newblock Multivariate time series imputation by graph neural networks.
\newblock \emph{arXiv e-prints}, pp.\  arXiv--2108, 2021.

\bibitem[Dauphin et~al.(2017)Dauphin, Fan, Auli, and
  Grangier]{dauphin2017language}
Yann~N Dauphin, Angela Fan, Michael Auli, and David Grangier.
\newblock Language modeling with gated convolutional networks.
\newblock In \emph{International conference on machine learning}, pp.\
  933--941. PMLR, 2017.

\bibitem[Deng \& Hooi(2021)Deng and Hooi]{deng2021graph}
Ailin Deng and Bryan Hooi.
\newblock Graph neural network-based anomaly detection in multivariate time
  series.
\newblock In \emph{Proceedings of the AAAI Conference on Artificial
  Intelligence}, volume~35, pp.\  4027--4035, 2021.

\bibitem[Fang et~al.(2021)Fang, Long, Song, and Xie]{fang2021STODE}
Zheng Fang, Qingqing Long, Guojie Song, and Kunqing Xie.
\newblock Spatial-temporal graph ode networks for traffic flow forecasting.
\newblock In \emph{KDD}, 2021.

\bibitem[Fey(2019)]{fey2019just}
Matthias Fey.
\newblock Just jump: Dynamic neighborhood aggregation in graph neural networks.
\newblock \emph{arXiv preprint arXiv:1904.04849}, 2019.

\bibitem[Gao \& Ji(2019)Gao and Ji]{gao2019graph}
Hongyang Gao and Shuiwang Ji.
\newblock Graph u-nets.
\newblock In \emph{international conference on machine learning}, pp.\
  2083--2092. PMLR, 2019.

\bibitem[Gao \& Ribeiro(2022)Gao and Ribeiro]{gao2022equivalence}
Jianfei Gao and Bruno Ribeiro.
\newblock On the equivalence between temporal and static equivariant graph
  representations.
\newblock In \emph{International Conference on Machine Learning}, pp.\
  7052--7076. PMLR, 2022.

\bibitem[Gilmer et~al.(2017)Gilmer, Schoenholz, Riley, Vinyals, and
  Dahl]{gilmer2017neural}
Justin Gilmer, Samuel~S Schoenholz, Patrick~F Riley, Oriol Vinyals, and
  George~E Dahl.
\newblock Neural message passing for quantum chemistry.
\newblock In \emph{International conference on machine learning}, pp.\
  1263--1272. PMLR, 2017.

\bibitem[Guo et~al.(2020)Guo, Hu, Qian, Liu, Zhang, Sun, Gao, and
  Yin]{guo2020optimized}
Kan Guo, Yongli Hu, Zhen Qian, Hao Liu, Ke~Zhang, Yanfeng Sun, Junbin Gao, and
  Baocai Yin.
\newblock Optimized graph convolution recurrent neural network for traffic
  prediction.
\newblock \emph{IEEE Transactions on Intelligent Transportation Systems},
  22\penalty0 (2):\penalty0 1138--1149, 2020.

\bibitem[Guo et~al.(2019)Guo, Lin, Feng, Song, and Wan]{guo2019astgcn}
Shengnan Guo, Youfang Lin, Ning Feng, Chao Song, and Huaiyu Wan.
\newblock Attention based spatial-temporal graph convolutional networks for
  traffic flow forecasting.
\newblock In \emph{AAAI}, Jul. 2019.
\newblock \doi{10.1609/aaai.v33i01.3301922}.

\bibitem[Hamilton(2020)]{hamilton2020time}
James~Douglas Hamilton.
\newblock \emph{Time series analysis}.
\newblock Princeton university press, 2020.

\bibitem[He et~al.(2016)He, Zhang, Ren, and Sun]{he2016deep}
Kaiming He, Xiangyu Zhang, Shaoqing Ren, and Jian Sun.
\newblock Deep residual learning for image recognition.
\newblock In \emph{CVPR}, 2016.

\bibitem[Huang et~al.(2020)Huang, Huang, Liu, Dai, and Kong]{huang2020lsgcn}
Rongzhou Huang, Chuyin Huang, Yubao Liu, Genan Dai, and Weiyang Kong.
\newblock Lsgcn: Long short-term traffic prediction with graph convolutional
  networks.
\newblock In \emph{IJCAI}, pp.\  2355--2361, 2020.

\bibitem[Huang et~al.(2019)Huang, Wang, Wu, and Tang]{Huang2019DSANet}
Siteng Huang, Donglin Wang, Xuehan Wu, and Ao~Tang.
\newblock Dsanet: Dual self-attention network for multivariate time series
  forecasting.
\newblock In \emph{CIKM}, November 2019.

\bibitem[Hyndman et~al.(2008)Hyndman, Koehler, Ord, and
  Snyder]{hyndman2008forecasting}
Rob Hyndman, Anne~B Koehler, J~Keith Ord, and Ralph~D Snyder.
\newblock \emph{Forecasting with exponential smoothing: the state space
  approach}.
\newblock Springer Science \& Business Media, 2008.

\bibitem[Ishiguro et~al.(2019)Ishiguro, Maeda, and Koyama]{ishiguro2019graph}
Katsuhiko Ishiguro, Shin-ichi Maeda, and Masanori Koyama.
\newblock Graph warp module: an auxiliary module for boosting the power of
  graph neural networks.
\newblock \emph{arXiv preprint arXiv:1902.01020}, 2019.

\bibitem[Jang et~al.(2016)Jang, Gu, and Poole]{jang2016categorical}
Eric Jang, Shixiang Gu, and Ben Poole.
\newblock Categorical reparameterization with gumbel-softmax.
\newblock \emph{arXiv preprint arXiv:1611.01144}, 2016.

\bibitem[Jiang et~al.(2021)Jiang, Yin, Wang, Wang, Deng, Liu, Cai, Deng, Song,
  and Shibasaki]{jiang2021dl}
Renhe Jiang, Du~Yin, Zhaonan Wang, Yizhuo Wang, Jiewen Deng, Hangchen Liu,
  Zekun Cai, Jinliang Deng, Xuan Song, and Ryosuke Shibasaki.
\newblock Dl-traff: Survey and benchmark of deep learning models for urban
  traffic prediction.
\newblock In \emph{Proceedings of the 30th ACM international conference on
  information \& knowledge management}, pp.\  4515--4525, 2021.

\bibitem[Kipf et~al.(2018)Kipf, Fetaya, Wang, Welling, and
  Zemel]{kipf2018neural}
Thomas Kipf, Ethan Fetaya, Kuan-Chieh Wang, Max Welling, and Richard Zemel.
\newblock Neural relational inference for interacting systems.
\newblock In \emph{International Conference on Machine Learning}, pp.\
  2688--2697. PMLR, 2018.

\bibitem[Kipf \& Welling(2016)Kipf and Welling]{kipf2016semi}
Thomas~N Kipf and Max Welling.
\newblock Semi-supervised classification with graph convolutional networks.
\newblock \emph{arXiv preprint arXiv:1609.02907}, 2016.

\bibitem[Lai et~al.(2018)Lai, Chang, Yang, and Liu]{lai2018modeling}
Guokun Lai, Wei-Cheng Chang, Yiming Yang, and Hanxiao Liu.
\newblock Modeling long-and short-term temporal patterns with deep neural
  networks.
\newblock In \emph{The 41st international ACM SIGIR conference on research \&
  development in information retrieval}, pp.\  95--104, 2018.

\bibitem[Lee et~al.(2019)Lee, Lee, and Kang]{lee2019self}
Junhyun Lee, Inyeop Lee, and Jaewoo Kang.
\newblock Self-attention graph pooling.
\newblock In \emph{International conference on machine learning}, pp.\
  3734--3743. PMLR, 2019.

\bibitem[Li \& Zhu(2021)Li and Zhu]{li2021stfgnn}
Mengzhang Li and Zhanxing Zhu.
\newblock Spatial-temporal fusion graph neural networks for traffic flow
  forecasting.
\newblock In \emph{AAAI}, May 2021.

\bibitem[Li et~al.(2019)Li, Jin, Xuan, Zhou, Chen, Wang, and
  Yan]{li2019enhancing}
Shiyang Li, Xiaoyong Jin, Yao Xuan, Xiyou Zhou, Wenhu Chen, Yu-Xiang Wang, and
  Xifeng Yan.
\newblock Enhancing the locality and breaking the memory bottleneck of
  transformer on time series forecasting.
\newblock \emph{Advances in neural information processing systems}, 32, 2019.

\bibitem[Li et~al.(2017)Li, Yu, Shahabi, and Liu]{li2017diffusion}
Yaguang Li, Rose Yu, Cyrus Shahabi, and Yan Liu.
\newblock Diffusion convolutional recurrent neural network: Data-driven traffic
  forecasting.
\newblock \emph{arXiv preprint arXiv:1707.01926}, 2017.

\bibitem[Li et~al.(2018{\natexlab{a}})Li, Yu, Shahabi, and Liu]{LiYS018}
Yaguang Li, Rose Yu, Cyrus Shahabi, and Yan Liu.
\newblock Diffusion convolutional recurrent neural network: Data-driven traffic
  forecasting.
\newblock In \emph{{ICLR} (Poster)}, 2018{\natexlab{a}}.

\bibitem[Li et~al.(2018{\natexlab{b}})Li, Yu, Shahabi, and
  Liu]{li2018dcrnn_traffic}
Yaguang Li, Rose Yu, Cyrus Shahabi, and Yan Liu.
\newblock Diffusion convolutional recurrent neural network: Data-driven traffic
  forecasting.
\newblock In \emph{ICLR}, 2018{\natexlab{b}}.

\bibitem[Lim et~al.(2021)Lim, Ar{\i}k, Loeff, and Pfister]{lim2021temporal}
Bryan Lim, Sercan~{\"O} Ar{\i}k, Nicolas Loeff, and Tomas Pfister.
\newblock Temporal fusion transformers for interpretable multi-horizon time
  series forecasting.
\newblock \emph{International Journal of Forecasting}, 37\penalty0
  (4):\penalty0 1748--1764, 2021.

\bibitem[Liu et~al.(2021)Liu, Yu, Liao, Li, Lin, Liu, and
  Dustdar]{liu2021pyraformer}
Shizhan Liu, Hang Yu, Cong Liao, Jianguo Li, Weiyao Lin, Alex~X Liu, and
  Schahram Dustdar.
\newblock Pyraformer: Low-complexity pyramidal attention for long-range time
  series modeling and forecasting.
\newblock In \emph{International Conference on Learning Representations}, 2021.

\bibitem[Maddison et~al.(2016)Maddison, Mnih, and Teh]{maddison2016concrete}
Chris~J Maddison, Andriy Mnih, and Yee~Whye Teh.
\newblock The concrete distribution: A continuous relaxation of discrete random
  variables.
\newblock \emph{arXiv preprint arXiv:1611.00712}, 2016.

\bibitem[Makridakis \& Hibon(1997)Makridakis and Hibon]{makridakis1997arma}
Spyros Makridakis and Michele Hibon.
\newblock Arma models and the box--jenkins methodology.
\newblock \emph{Journal of forecasting}, 16\penalty0 (3):\penalty0 147--163,
  1997.

\bibitem[Nix \& Weigend(1994)Nix and Weigend]{nix1994estimating}
David~A Nix and Andreas~S Weigend.
\newblock Estimating the mean and variance of the target probability
  distribution.
\newblock In \emph{Proceedings of 1994 ieee international conference on neural
  networks (ICNN'94)}, volume~1, pp.\  55--60. IEEE, 1994.

\bibitem[Pham et~al.(2017)Pham, Tran, Dam, and Venkatesh]{pham2017graph}
Trang Pham, Truyen Tran, Hoa Dam, and Svetha Venkatesh.
\newblock Graph classification via deep learning with virtual nodes.
\newblock \emph{arXiv preprint arXiv:1708.04357}, 2017.

\bibitem[Pilco \& Rivera(2019)Pilco and Rivera]{pilco2019graph}
Darwin~Saire Pilco and Ad{\'\i}n~Ram{\'\i}rez Rivera.
\newblock Graph learning network: A structure learning algorithm.
\newblock \emph{arXiv preprint arXiv:1905.12665}, 2019.

\bibitem[Ramp{\'a}{\v{s}}ek \& Wolf(2021)Ramp{\'a}{\v{s}}ek and
  Wolf]{rampavsek2021hierarchical}
Ladislav Ramp{\'a}{\v{s}}ek and Guy Wolf.
\newblock Hierarchical graph neural nets can capture long-range interactions.
\newblock In \emph{2021 IEEE 31st International Workshop on Machine Learning
  for Signal Processing (MLSP)}, pp.\  1--6. IEEE, 2021.

\bibitem[Roth \& Liebig(2022)Roth and Liebig]{roth2022forecasting}
Andreas Roth and Thomas Liebig.
\newblock Forecasting unobserved node states with spatio-temporal graph neural
  networks.
\newblock \emph{arXiv preprint arXiv:2211.11596}, 2022.

\bibitem[Shang et~al.(2021)Shang, Chen, and Bi]{shang2021discrete}
Chao Shang, Jie Chen, and Jinbo Bi.
\newblock Discrete graph structure learning for forecasting multiple time
  series.
\newblock \emph{arXiv preprint arXiv:2101.06861}, 2021.

\bibitem[Shih et~al.(2019)Shih, Sun, and Lee]{shih2019temporal}
Shun-Yao Shih, Fan-Keng Sun, and Hung-yi Lee.
\newblock Temporal pattern attention for multivariate time series forecasting.
\newblock \emph{Machine Learning}, 108\penalty0 (8):\penalty0 1421--1441, 2019.

\bibitem[Song et~al.(2020)Song, Lin, Guo, and Wan]{song2020stsgcn}
Chao Song, Youfang Lin, Shengnan Guo, and Huaiyu Wan.
\newblock Spatial-temporal synchronous graph convolutional networks: A new
  framework for spatial-temporal network data forecasting.
\newblock In \emph{AAAI}, Apr. 2020.
\newblock \doi{10.1609/aaai.v34i01.5438}.

\bibitem[Sutskever et~al.(2014)Sutskever, Vinyals, and
  Le]{sutskever2014sequence}
Ilya Sutskever, Oriol Vinyals, and Quoc~V Le.
\newblock Sequence to sequence learning with neural networks.
\newblock In \emph{NeurIPS}, pp.\  3104--3112, 2014.

\bibitem[Vaswani et~al.(2017)Vaswani, Shazeer, Parmar, Uszkoreit, Jones, Gomez,
  Kaiser, and Polosukhin]{vaswani2017attention}
Ashish Vaswani, Noam Shazeer, Niki Parmar, Jakob Uszkoreit, Llion Jones,
  Aidan~N Gomez, {\L}ukasz Kaiser, and Illia Polosukhin.
\newblock Attention is all you need.
\newblock \emph{Advances in neural information processing systems}, 30, 2017.

\bibitem[Veli{\v{c}}kovi{\'c} et~al.(2017)Veli{\v{c}}kovi{\'c}, Cucurull,
  Casanova, Romero, Lio, and Bengio]{velivckovic2017graph}
Petar Veli{\v{c}}kovi{\'c}, Guillem Cucurull, Arantxa Casanova, Adriana Romero,
  Pietro Lio, and Yoshua Bengio.
\newblock Graph attention networks.
\newblock \emph{arXiv preprint arXiv:1710.10903}, 2017.

\bibitem[Watson(1994)]{watson1994vector}
Mark~W Watson.
\newblock Vector autoregressions and cointegration.
\newblock \emph{Handbook of econometrics}, 4:\penalty0 2843--2915, 1994.

\bibitem[Wu et~al.(2021)Wu, Xu, Wang, and Long]{wu2021autoformer}
Haixu Wu, Jiehui Xu, Jianmin Wang, and Mingsheng Long.
\newblock Autoformer: Decomposition transformers with auto-correlation for
  long-term series forecasting.
\newblock \emph{Advances in Neural Information Processing Systems},
  34:\penalty0 22419--22430, 2021.

\bibitem[Wu et~al.(2022)Wu, Cui, Pei, Zhao, and Guo]{wu2022graph}
Lingfei Wu, Peng Cui, Jian Pei, Liang Zhao, and Xiaojie Guo.
\newblock Graph neural networks: foundation, frontiers and applications.
\newblock In \emph{Proceedings of the 28th ACM SIGKDD Conference on Knowledge
  Discovery and Data Mining}, pp.\  4840--4841, 2022.

\bibitem[Wu et~al.(2019{\natexlab{a}})Wu, Pan, Long, Jiang, and
  Zhang]{wu2019graphwavenet}
Zonghan Wu, Shirui Pan, Guodong Long, Jing Jiang, and Chengqi Zhang.
\newblock Graph wavenet for deep spatial-temporal graph modeling.
\newblock In \emph{IJCAI}, pp.\  1907--1913, 7 2019{\natexlab{a}}.

\bibitem[Wu et~al.(2019{\natexlab{b}})Wu, Pan, Long, Jiang, and
  Zhang]{zonghanwu2019}
Zonghan Wu, Shirui Pan, Guodong Long, Jing Jiang, and Chengqi Zhang.
\newblock Graph wavenet for deep spatial-temporal graph modeling.
\newblock In \emph{{IJCAI}}, pp.\  1907--1913, 2019{\natexlab{b}}.

\bibitem[Wu et~al.(2020)Wu, Pan, Long, Jiang, Chang, and
  Zhang]{wu2020connecting}
Zonghan Wu, Shirui Pan, Guodong Long, Jing Jiang, Xiaojun Chang, and Chengqi
  Zhang.
\newblock Connecting the dots: Multivariate time series forecasting with graph
  neural networks.
\newblock In \emph{Proceedings of the 26th ACM SIGKDD international conference
  on knowledge discovery \& data mining}, pp.\  753--763, 2020.

\bibitem[Xu et~al.(2018)Xu, Li, Tian, Sonobe, Kawarabayashi, and
  Jegelka]{xu2018representation}
Keyulu Xu, Chengtao Li, Yonglong Tian, Tomohiro Sonobe, Ken-ichi Kawarabayashi,
  and Stefanie Jegelka.
\newblock Representation learning on graphs with jumping knowledge networks.
\newblock In \emph{International conference on machine learning}, pp.\
  5453--5462. PMLR, 2018.

\bibitem[Yu et~al.(2017)Yu, Yin, and Zhu]{yu2017spatio}
Bing Yu, Haoteng Yin, and Zhanxing Zhu.
\newblock Spatio-temporal graph convolutional networks: A deep learning
  framework for traffic forecasting.
\newblock \emph{arXiv preprint arXiv:1709.04875}, 2017.

\bibitem[Yu et~al.(2018{\natexlab{a}})Yu, Yin, and Zhu]{YuYZ18}
Bing Yu, Haoteng Yin, and Zhanxing Zhu.
\newblock Spatio-temporal graph convolutional networks: {A} deep learning
  framework for traffic forecasting.
\newblock In \emph{{IJCAI}}, pp.\  3634--3640, 2018{\natexlab{a}}.

\bibitem[Yu et~al.(2018{\natexlab{b}})Yu, Yin, and Zhu]{bing2018stgcn}
Bing Yu, Haoteng Yin, and Zhanxing Zhu.
\newblock Spatio-temporal graph convolutional networks: A deep learning
  framework for traffic forecasting.
\newblock In \emph{IJCAI}, 7 2018{\natexlab{b}}.
\newblock \doi{10.24963/ijcai.2018/505}.
\newblock URL \url{https://doi.org/10.24963/ijcai.2018/505}.

\bibitem[Zhao et~al.(2019)Zhao, Song, Zhang, Liu, Wang, Lin, Deng, and
  Li]{zhao2019t}
Ling Zhao, Yujiao Song, Chao Zhang, Yu~Liu, Pu~Wang, Tao Lin, Min Deng, and
  Haifeng Li.
\newblock T-gcn: A temporal graph convolutional network for traffic prediction.
\newblock \emph{IEEE Transactions on Intelligent Transportation Systems},
  21\penalty0 (9):\penalty0 3848--3858, 2019.

\bibitem[Zhou et~al.(2021)Zhou, Zhang, Peng, Zhang, Li, Xiong, and
  Zhang]{zhou2021informer}
Haoyi Zhou, Shanghang Zhang, Jieqi Peng, Shuai Zhang, Jianxin Li, Hui Xiong,
  and Wancai Zhang.
\newblock Informer: Beyond efficient transformer for long sequence time-series
  forecasting.
\newblock In \emph{Proceedings of the AAAI Conference on Artificial
  Intelligence}, volume~35, pp.\  11106--11115, 2021.

\bibitem[Zhou et~al.(2022)Zhou, Ma, Wen, Wang, Sun, and Jin]{zhou2022fedformer}
Tian Zhou, Ziqing Ma, Qingsong Wen, Xue Wang, Liang Sun, and Rong Jin.
\newblock Fedformer: Frequency enhanced decomposed transformer for long-term
  series forecasting.
\newblock \emph{arXiv preprint arXiv:2201.12740}, 2022.

\end{thebibliography}
\bibliographystyle{iclr2023_conference}

\vspace{-1mm}
\appendix
\section{Appendix} 

\vspace{-2mm}
\subsection{Additional Results}
\vspace{-2mm}
Table \ref{tab:results2} compares the performance of different models(\textbf{EIKF-Net}, \textbf{w/Unc- EIKF-Net}, and several baseline models) on two benchmark datasets (METR-LA and PEMS-BAY) for multi-horizon forecasting tasks. The performance of each model is evaluated using MAE, RMSE, and MAPE metrics. The forecast errors are reported for 3-step-ahead, 6-step-ahead, and 12-step-ahead forecast horizons. The lower the forecast errors, the better the performance of the models. The results for the baseline methods are taken from a previous study by \cite{jiang2021dl}. On both METR-LA and PEMS-BAY datasets, the experimental results show that the proposed models (\textbf{EIKF-Net}, \textbf{w/Unc- EIKF-Net}) outperform the baseline models on different evaluation metrics on a range of forecast horizons. On the PEMS-BAY dataset, the proposed method(EIKF-Net) has forecast errors that are $40.6\%$, $33.5\%$, and $28.4\%$ lower than the next-to-best baseline model's forecast errors for the 3-step-ahead, 6-step-ahead, and 12-step-ahead forecast horizons, respectively, as measured by the MAPE metric. Similarly, on the METR-LA dataset, the proposed model(\textbf{EIKF-Net}) reports an improvement of $27.6\%$, $5.5\%$, and $-2.6\%$ over the next-best baseline model's for the 3-step-ahead, 6-step-ahead, and 12-step-ahead forecast horizons, respectively, as measured by the MAPE metric.

\vspace{-3mm}
\begin{table}[ht!]
\setlength{\tabcolsep}{0.35em} 
\renewcommand\arraystretch{1.1} 
\centering
 \resizebox{0.95\textwidth}{!}{
\begin{tabular}{c|c|ccc|ccc|ccc}
\hline
\multirow{2}{*}{\textbf{Datasets}}  & \multirow{2}{*}{\textbf{Methods}} & \multicolumn{3}{c|}{\textbf{Horizon$\textbf{@}$3}}       & \multicolumn{3}{c|}{\textbf{Horizon$\textbf{@}$6}}       & \multicolumn{3}{c}{\textbf{Horizon$\textbf{@}$12}}       \\ \cline{3-11} 
                                    &                                   & \textbf{RMSE} & \textbf{MAE}  & \textbf{MAPE} & \textbf{RMSE} & \textbf{MAE}  & \textbf{MAPE} & \textbf{RMSE} & \textbf{MAE}  & \textbf{MAPE} \\ \hline
\multirow{13}{*}{\textbf{METR-LA}}  & HA                       & 10.00         & 4.79          & 11.70         & 11.45         & 5.47          & 13.50         & 13.89         & 6.99          & 17.54         \\
                                    & VAR                      & 7.80          & 4.42          & 13.00         & 9.13          & 5.41          & 12.70         & 10.11         & 6.52          & 15.80         \\
                                    & SVR                      & 8.45          & 3.39          & 9.30          & 10.87         & 5.05          & 12.10         & 13.76         & 6.72          & 16.70         \\
                                    & FC-LSTM                  & 6.30          & 3.44          & 9.60          & 7.23          & 3.77          & 10.09         & 8.69          & 4.37          & 14.00         \\
                                    & DCRNN                    & 5.38          & 2.77          & 7.30          & 6.45          & 3.15          & 8.80          & 7.60          & 3.60          & 10.50         \\
                                    & STGCN                    & 5.74          & 2.88          & 7.62          & 7.24          & 3.47          & 9.57          & 9.40          & 4.59          & 12.70         \\
                                    & Graph WaveNet            & 5.15          & 2.69          & 6.90          & 6.22          & 3.07          & 8.37          & 7.37          & 3.53          & 10.01         \\
                                    & ASTGCN                   & 9.27          & 4.86          & 9.21          & 10.61         & 5.43          & 10.13         & 12.52         & 6.51          & 11.64         \\
                                    & STSGCN                   & 7.62          & 3.31          & 8.06          & 9.77          & 4.13          & 10.29         & 11.66         & 5.06          & 12.91         \\
                                    & MTGNN                    & 5.18          & 2.69          & 6.88          & 6.17          & 3.05          & 8.19          & 7.23          & 3.49          & 9.87          \\
                                    & GMAN                     & 5.55          & 2.80          & 7.41          & 6.49          & 3.12          & 8.73          & 7.35          & 3.44          & 10.07         \\
                                    & DGCRN                    & 5.01          & 2.62          & 6.63          & \textbf{6.05} & \textbf{2.99} & 8.02          & \textbf{7.19} & \textbf{3.44} & \textbf{9.73} \\ \cline{2-11} 
                                    & \textbf{EIKF-Net}               & \textbf{3.96} & \textbf{1.99} & \textbf{4.80} & 6.30          & 3.21          & \textbf{7.58} & 7.31 & 3.56          & 9.98          \\ 
                                     & \textbf{w/Unc- EIKF-Net}               & 4.13 & 2.03 & 4.97 &    6.41       &     3.34      & 7.63 &          7.38 &     3.61      &   10.15        \\ \hline \hline
\multirow{13}{*}{\textbf{PEMS-BAY}} & HA                       & 4.30          & 1.89          & 4.16          & 5.82          & 2.50          & 5.62          & 7.54          & 3.31          & 7.65          \\
                                    & VAR                      & 3.16          & 1.74          & 3.60          & 4.25          & 2.32          & 5.00          & 5.44          & 2.93          & 6.50          \\
                                    & SVR                      & 3.59          & 1.85          & 3.80          & 5.18          & 2.48          & 5.50          & 7.08          & 3.28          & 8.00          \\
                                    & FC-LSTM                  & 4.19          & 2.05          & 4.80          & 4.55          & 2.20          & 5.20          & 4.96          & 2.37          & 5.70          \\
                                    & DCRNN                    & 2.95          & 1.38          & 2.90          & 3.97          & 1.74          & 3.90          & 4.74          & 2.07          & 4.90          \\
                                    & STGCN                    & 2.96          & 1.36          & 2.90          & 4.27          & 1.81          & 4.17          & 5.69          & 2.49          & 5.79          \\
                                    & Graph WaveNet            & 2.74          & 1.30          & 2.73          & 3.70          & 1.63          & 3.67          & 4.52          & 1.95          & 4.63          \\
                                    & ASTGCN                   & 3.13          & 1.52          & 3.22          & 4.27          & 2.01          & 4.48          & 5.42          & 2.61          & 6.00          \\
                                    & STSGCN                   & 3.01          & 1.44          & 3.04          & 4.18          & 1.83          & 4.17          & 5.21          & 2.26          & 5.40          \\
                                    & MTGNN                    & 2.79          & 1.32          & 2.77          & 3.74          & 1.65          & 3.69          & 4.49          & 1.94          & 4.53          \\
                                    & GMAN                     & 2.91          & 1.34          & 2.86          & 3.76          & 1.63          & 3.68          & 4.32          & 1.86          & 4.37          \\
                                    & DGCRN                    & 2.69          & 1.28          & 2.66          & 3.63          & 1.59          & 3.55          & 4.42          & 1.89          & 4.43          \\ \cline{2-11} 
                                    & \textbf{EIKF-Net}               & \textbf{1.65} & \textbf{0.81} & \textbf{1.58} & \textbf{2.51} & \textbf{1.22} & \textbf{2.36} & \textbf{3.07} & \textbf{1.68} & \textbf{3.13} \\ 
                                    & \textbf{w/Unc- EIKF-Net}               & 1.72 & 0.86 & 1.63 &   2.58        &    1.26       & 2.41 &   3.14        &   1.74        &     3.27      \\ \hline \hline
\end{tabular}
}
\vspace{-1mm}
\caption{Pointwise forecast and predictive uncertainty errors on METR-LA and PEMS-BAY.}
\label{tab:results2}
\end{table}

\vspace{-8mm}
\subsection{Hypergraph Attention Network(HgAT)}
\vspace{-4mm}
The HgAT generalizes the local- and global-attention-based convolution operation(\cite{velivckovic2017graph, brody2021attentive}) on the spatio-temporal hypergraphs. The hypergraph encoder(HgAT) performs inference on the hypergraph-structured MTS data characterized by the incidence matrix, \resizebox{.085\textwidth}{!}{$\mathbf{I} \in \mathbb{R}^{n \times m}$}, and feature matrix, \resizebox{.155\textwidth}{!}{$\bar{\mathbf{X}}_{(t - \tau : \hspace{1mm}t-1)} \in \mathbb{R}^{n \times d}$}, to compute the transformed feature matrix \resizebox{.095\textwidth}{!}{$\mathbf{H}_{t} \in \mathbb{R}^{n \times d}$}. Each row in \resizebox{.03\textwidth}{!}{$\mathbf{H}_{t}$} represents the hypernode representations \resizebox{.155\textwidth}{!}{$\mathbf{h}^{t}_{i} \in \mathbb{R}^{d}, \forall i \in \mathcal{HV}$} at time step t, where \resizebox{.0375\textwidth}{!}{$\mathcal{HV}$} is the set of hypernodes. The HgAT operator captures the interdependencies and relations among the time-series variables by encoding both the structural and feature characteristics of the spatio-temporal hypergraphs in the hypernode representations $\mathbf{h}^{t}_{i}$. The HgAT operator is designed to adapt to capture the changes in the dependencies of the time-series variables over time in the hypernode representations $\mathbf{h}^{t}_{i}$. Let \resizebox{.175\textwidth}{!}{$\mathcal{N}_{j, i}=\big\{i | \mathbf{I}_{i, j} = 1\big\}$} represent the subset of hypernodes $i$ incident with any hyperedge $j$. The intra-edge neighborhood of the incident hypernode $i$ is given by \resizebox{.05\textwidth}{!}{$\mathcal{N}_{j,i} \backslash i$}. It is a localized group of semantically-correlated time series variables and captures higher-order relationships. The inter-edge neighborhood of hypernode $i$, \resizebox{.175\textwidth}{!}{$\mathcal{N}_{i, j} = \big\{j|\mathbf{I}_{i, j} = 1\big\}$}, spans the spectrum of the set of hyperedges $j$ incident with hypernode $i$. The HgAT operator leverages the relational inductive bias encoded by the hypergraph's connectivity to perform intra-edge and inter-edge neighborhood aggregation schemes, which allows it to explicitly model the fine-grained spatial-temporal correlations between the time series variables. The intra-edge neighborhood aggregation focuses on the interrelations between a specific hypernode and its immediate neighboring hypernodes incident with a specific hyperedge, while inter-edge neighborhood aggregation takes into account the interrelations between a specific hypernode and all other hyperedges incident with it. We peform the attention-based intra-edge neighborhood aggregation for learning the latent hyperedge representations as follows,

\vspace{-4mm}
\resizebox{0.9\linewidth}{!}{
\begin{minipage}{\linewidth}
\begin{equation}
\mathbf{h}^{(t, \ell)}_{j} =  \sum_{z=1}^{\mathcal{Z}} \sigma \big( \hspace{-0.25mm}  \sum_{i \hspace{0.5mm}\in \hspace{0.5mm}{\mathcal{N}_{j, i}}} \hspace{-1mm}  \alpha^{(t, \ell, z)}_{j, i} \mathbf{W}^{(z)}_{0}\mathbf{h}^{(t, \ell-1, z)}_{i} \big),  \ell=1 \ldots \text{L}_{\text{HgAT}}\label{eq:hgcnn1}
\end{equation}
\end{minipage}
}

\vspace{-2mm}
where $\mathbf{h}_{j} \in \mathbb{R}^{d}$ represents the hyperedge representations. The superscript $\ell$ denotes the layer. We utilize a single HgAT layer, $\text{L}_{\text{HgAT}} = 1$. The initial representation of each hypernode is set to its feature vector i.e., $\mathbf{h}^{(t, 0, z)}_{i} = \hspace{1mm}\bar{\mathbf{x}}^{t}_{i}$, where each row in \resizebox{.155\textwidth}{!}{$\bar{\mathbf{X}}_{(t - \tau : \hspace{1mm}t-1)} \in \mathbb{R}^{n \times d}$} is denoted by $\bar{\mathbf{x}}^{t}_{i}$. $\sigma$ represents the sigmoid function. The HgAT operator computes multiple representations of the data, each with its own set of parameters, and then combines these representations by summing them up.
This is similar to multi-head self-attention mechanism. It allows the model to capture multiple different aspects underlying the hypergraph-structured MTS data. The attention coefficient $\alpha^{t}_{j, i}$ determines the relative importance of the hypernode $i$ incident with the hyperedge, $j$, at time t and is computed by,

\vspace{-9mm}
\resizebox{0.95\linewidth}{!}{
\hspace{1cm}\begin{minipage}{\linewidth}
\begin{align}
\alpha^{(t, \ell, z)}_{j, i} &= \frac{\exp \big(e^{(t, \ell, z)}_{j, i}\big)}{{\textstyle \sum_{i \hspace{0.5mm}\in \hspace{0.5mm}{\mathcal{N}_{j, i}}} \exp \big(e^{(t, \ell, z)}_{j, i}\big)}} ; e^{(t, \ell, z)}_{j, i} =\operatorname{ReLU}\big(\text{W}^{(z)}_{0} \mathbf{h}^{(t, \ell-1, z)}_{i}\big) 
\end{align}
\end{minipage}
}

\vspace{-1mm}
where $e_{j, i}$ denotes the unnormalized attention score. The model captures the complex dependencies and relations between hyperedges and hypernodes by using an attention-based inter-edge neighborhood aggregation scheme into the expressive hypernode representations as computed by,

\vspace{-3mm}
\resizebox{0.95\linewidth}{!}{
\begin{minipage}{\linewidth}
\begin{equation}
\mathbf{h}^{(t, \ell)}_{i}=\sum_{z=1}^{\mathcal{Z}} \operatorname{ReLU}\big(\text{W}^{(z)}_{0}\mathbf{h}^{(t, \ell-1, z)}_{i} + \sum_{j \in \mathcal{N}_{i, j}} \beta^{(t, \ell, z)}_{i, j} \text{W}^{(z)}_{1} \mathbf{h}^{(t, \ell, z)}_{j}\big),  \ell=1 \ldots L_{\text{HgAT}}
\end{equation}
\end{minipage}
} 

\vspace{-2mm}
where \resizebox{.185\textwidth}{!}{$\text{W}^{(z)}_{0}, \text{W}^{(z)}_{1} \in \mathbb{R}^{d \times d}$} are trainable weight matrices. We utilize the $\operatorname{ReLU}$ function to introduce non-linearity for updating the hypernode-level representations. The normalized attention scores $\beta_{i, j}$ specify the importance of hyperedge $j$ incident with  hypernode $i$, allowing the HgAT operator to focus on the most relevant hyperedges and are computed by,

\vspace{-3mm}
\resizebox{1\linewidth}{!}{
\hspace{1cm}\begin{minipage}{\linewidth}
\begin{align}
\beta^{(t, \ell, z)}_{i, j} =  \frac{\exp (\phi^{(t, \ell, z)}_{i, j})}{{\textstyle \sum_{j \hspace{0.5mm}\in \hspace{0.5mm}{\mathcal{N}_{i, j}}} \exp (\phi^{(\ell, z)}_{i, j})}}; \phi^{(t, \ell, z)}_{i, j} = \operatorname{ReLU}\big(\text{W}^{(z)}_{3} \cdot \big(\text{W}^{(z)}_{2}\mathbf{h}^{(t, \ell-1, z)}_{i} \oplus \text{W}^{(z)}_{2} \mathbf{h}^{(t, \ell, z)}_{j}\big)\big) \nonumber
\end{align}
\end{minipage}
} 

\vspace{-1mm}
where \resizebox{.125\textwidth}{!}{$\text{W}^{(z)}_{2} \in \mathbb{R}^{d \times d}$} and \resizebox{.115\textwidth}{!}{$\text{W}^{(z)}_{3} \in \mathbb{R}^{2d}$} are weight matrix and vector respectively. $\oplus$ denotes the concatenation operator. $\phi_{i, j}$ is the unnormalized attention score. We apply batch norm and dropout for regularization. We regulate the information flow by applying a gating mechanism to selectively fuse features from $\bar{\mathbf{x}}^{t}_{i}$ and $\mathbf{h}^{(t, \ell)}_{i}$ using gates through a differentiable approach as described below,

\vspace{-4mm}
\resizebox{1\linewidth}{!}{
\hspace{1cm}\begin{minipage}{\linewidth}
\begin{align}
g^{t}  &= \sigma \big( f_s(\mathbf{h}^{(t, \ell)}_{i}) + f_g(\bar{\mathbf{x}}^{t}_{i}) \big) \\
\mathbf{h}^{(t, \ell)}_{i}  &= \sigma \big( g^{t}(\mathbf{h}^{(t, \ell)}_{i}) + (1-g^{t})(\bar{\mathbf{x}}^{t}_{i}) \big)
\end{align}
\end{minipage}
} 

\vspace{-1mm}
where $f_s$ and $f_g$ are linear projections. This enables the model to better capture the relationships between the different time-series variables and how they change over time, in order to improve the forecast accuracy. As a result, the HgAT operator is a powerful tool for encoding and analyzing spatio-temporal hypergraphs, which provides a deeper understanding of the dependencies and relationships among the time-series variables in hypergraph-structured MTS data.

\vspace{-5mm}
\subsection{Hypergraph Transformer(HgT)}
\vspace{-3mm}
The proposed HgT operator is an extension of transformer networks(\cite{vaswani2017attention}) to handle arbitrary sparse hypergraph structures with full attention as a desired structural inductive bias. The HgT operator allows the framework to attend to all hypernodes and hyperedges in the hypergraph, which incentivizes learning the fine-grained interrelations unconstrained by domain-specific hierarchical structural information underlying the MTS data. This can facilitate the learning of optimal hypergraph representations by allowing the model to span large receptive fields for global reasoning of the hypergraph-structured data. The HgT operator does not rely on structural priors, unlike existing methods such as the method of stacking multiple HgNN layers with residual connections(\cite{fey2019just}, \cite{xu2018representation}), virtual hypernode mechanisms(\cite{gilmer2017neural, ishiguro2019graph, pham2017graph}), or hierarchical pooling schemes(\cite{rampavsek2021hierarchical}, \cite{gao2019graph}, and \cite{lee2019self}) to model the long-range correlations in the hypergraph-structured data. The permutation-invariant HgT module, by exploiting global contextual information can model the pairwise relations between all hypernodes in the hypergraph-structured data. As a result, the HgT module acts as a drop-in replacement to existing methods for modeling the hierarchical dependencies and relationships among time series variables in the spatio-temporal hypergraphs leading to more robust and generalizable representations on a wide range of downstream tasks. The transformer encoder(\cite{vaswani2017attention}) consists of alternating layers of multiheaded self-attention (MSA) and multi-layer perceptron(MLP) blocks to capture both the local and global contextual information. To improve the performance and regularize the model, we apply Layer normalization(LN(\cite{ba2016layer})) and residual connections after every block. The transformer encoder is inspired by ResNets(\cite{he2016deep}) and is designed to relieve vanishing gradients and over-smoothing issues by adding skip-connections through an initial connection strategy. This allows the model to learn more complex and deeper representations of the hypergraph-structured data.

\vspace{-4mm}
\resizebox{1\linewidth}{!}{
\hspace{1cm}\begin{minipage}{\linewidth}
\begin{align}
\mathbf{h^{\prime}_{i}}^{(t, \ell)} &=\operatorname{MSA}\big(\operatorname{LN}\big(\mathbf{h}^{(t, \ell-1)}_{i}  \big)\big) + \mathbf{h}^{(t, \ell-1)}_{i}, & &  \ell=1 \ldots \text{L}_{\text{HgT}} \\
\mathbf{h^{\prime}_{i}}^{(t, \ell)} &=\operatorname{MLP}\big(\operatorname{LN}\big(\mathbf{h^{\prime}_{i}}^{(t, \ell)}\big)\big)+ \bar{\mathbf{x}}^{t}_{i}, & &  \ell=1 \ldots \text{L}_{\text{HgT}} 
\end{align}
\end{minipage}
} 

\vspace{-1mm}
In this proposed method, we utilize a single HgT layer, $\text{L}_{\text{HgT}}$ = 1 and the initial representation of each hypernode is set to its feature vector, $\mathbf{h}^{(t, 0)}_{v_{i}} = \bar{\mathbf{x}}^{t}_{i} \in \mathbb{R}^{d}$. The HgT is an effective method for hypergraph-structured data summarization that addresses the limitation of HgAT's representational capacity. The model learns the task-specific inter-relations between hypernodes beyond the original sparse structure and distills the long-range information in the downstream layers to learn task-specific, expressive hypergraph representations.

\vspace{-4mm}
\subsubsection{Spatial-Temporal Graph Representation Learning}
\vspace{-2mm}
The Temporal Graph Convolutional Network(T-GCN, \cite{zhao2019t}) operates on a sequence of dynamic graphs, where graph structure is fixed, and node attributes change over time. Each graph represents the graph-structured MTS data at a specific time step. The T-GCN operator utilizes Gated Recurrent Units(GRU, \cite{cho2014learning}) to model the spatio-temporal dynamics of the input dynamic graph sequence. In a traditional GRU, the update gate, reset gate, and hidden state are computed using matrix multiplication with weight matrices. However in T-GCN, these matrix multiplications are replaced with Graph Convolutional Networks(GCN, \cite{kipf2016semi}). The T-GCN operator analyzes graph-structured MTS data over time. It propagates information between nodes across different time steps, which enables the model to capture the complex spatio-temporal dependencies between the graphs. The T-GCN operator utilizes the predefined graph to propagate information between nodes by averaging the node representations in their local neighborhood at each time step computed as follows,
 
\vspace{-5mm} 
\resizebox{0.935\linewidth}{!}{
\hspace{1cm}\begin{minipage}{\linewidth}
\begin{align}
\mathbf{U}_t=\sigma\left(\text{W}_u\left[f\left(\text{A}^{(0)}, \mathbf{X}_{(t - \tau : \hspace{1mm}t-1)}\right), \mathbf{H}^{\prime\prime\prime}_{t-1}\right]+ \mathbf{B}_u\right) \\
\mathbf{R}_t=\sigma\left(\text{W}_r\left[f\left(\text{A}^{(0)}, \mathbf{X}_{(t - \tau : \hspace{1mm}t-1)}\right), \mathbf{H}^{\prime\prime\prime}_{t-1}\right]+ \mathbf{B}_r\right) \\
\mathbf{C}_t=\tanh \left(\text{W}_c\left[f\left(\text{A}^{(0)}, \mathbf{X}_{(t - \tau : \hspace{1mm}t-1)}\right),\left(\mathbf{R}_t * \mathbf{H}^{\prime\prime\prime}_{t-1}\right)\right]+ \mathbf{B}_c\right) \\
\mathbf{H}^{\prime\prime\prime}_t= \mathbf{U}_t \otimes \mathbf{H}^{\prime\prime\prime}_{t-1}+\left(1-\mathbf{U}_t\right) \otimes \mathbf{C}_t
\end{align}
\end{minipage}
} 

\vspace{-1mm}
where \resizebox{.165\textwidth}{!}{$f\big(\text{A}^{(0)}, \mathbf{X}_{(t - \tau : \hspace{1mm}t-1)}\big)$} denote the GCN operator. \resizebox{.075\textwidth}{!}{$\mathbf{U}_t$, $\mathbf{R}_t$} denote the  update gate and reset gate at time $t$. \resizebox{.175\textwidth}{!}{$\text{W}_r, \text{W}_u$, and $\text{W}_c$} are learnable weight matrices and \resizebox{.075\textwidth}{!}{$\mathbf{B}_u, \mathbf{B}_r$}, and \resizebox{.035\textwidth}{!}{$\mathbf{B}_c$} are learnable biases. In summary, the node representation matrix, \resizebox{.035\textwidth}{!}{$\mathbf{H}^{\prime\prime\prime}_t$} captures the spatio-temporal dynamics at different scales underlying the discrete-time dynamic graphs, where each row in \resizebox{.035\textwidth}{!}{$\mathbf{H}^{\prime\prime\prime}_t$} represents the hypernode representations \resizebox{.165\textwidth}{!}{$\mathbf{h^{\prime\prime\prime}_{i}}^{(t)} \in \mathbb{R}^{d}, \forall i \in \mathcal{V}$}. Some of the key advantages of T-GCN operator over traditional methods include its ability to handle large and sparse spatio-temporal graphs.

\vspace{-5mm}
\subsection{Ablation Study}
\vspace{-3mm}
We conduct a comprehensive ablation study to determine the impact of each component of the \textbf{EIKF-Net} framework by removing or altering them and observing the effect on the overall performance. We evaluate the individual contributions of components on the overall model performance. This study provides insights and helps identify components critical for the framework to achieve better performance. The baseline for our ablation study is the EIKF-Net framework, which integrates the spatial and temporal learning components to capture the fine-grained inter and intra-time-series correlations for modeling the nonlinear dynamics of complex interconnected sensor networks. The spatial learning component comprises two main modules: the explicit graph and the implicit hypergraph learning modules. The former operates on the predefined graph topology to capture the pair-wise semantic relationships between the variables in the graph-structured data to model the dynamics of the interconnected networks. The latter operates on the implicit hypergraph topology to capture time-evolving high-order spatial correlations in the hypergraph-structured data to model the observations of the dynamical interacting networks. We systematically exclude the components under evaluation to derive a set of model variants for verifying the importance of different components.  We investigate the ablated variant model's performance compared with the baseline model to disentangle the relative gains of each learning component. The ablation study will shed light on the relationship between the variant and the baseline model and their generalization performance for MTSF, advancing the importance of underlying mechanisms. We provide the details about each ablated variant as follows.

\vspace{-2mm}
\begin{itemize}
    \item Specifically, ``\textbf{w/o Spatial}": A variant of \textbf{EIKF-Net} framework without the spatial inference component. It shows the significance of learning explicit graph and implicit hypergraph representations through message-passing schemes in capturing the inter-series correlations.
    \item ``\textbf{w/o Temporal}": A variant of \textbf{EIKF-Net} without the temporal inference component. It verifies the effectiveness of utilizing temporal learning component to capture the time-varying inter-series dependencies. 
    \item In addition, ``\textbf{w/o Explicit Graph}": A variant of \textbf{EIKF-Net} framework without the explicit graph obtained from domain knowledge. It shows the importance of learning on the predefined graphs. The variant operates on the implicit hypergraph-structured data for learning latent hypergraph representations for downstream multi-horizon forecasting tasks.
    \item  While, ``\textbf{w/o Implicit Hypergraph}": A variant of \textbf{EIKF-Net} framework without the learned implicit hypergraph. It supports the rationale of learning implicit hypergraph structure underlying the MTS data. The variant performs inference on the graph-structured data for learning latent graph representations for downstream multi-horizon forecasting tasks.
\end{itemize}

\vspace{-2mm}
Table \ref{tab:ablationstudy} presents the results of the ablation studies on the benchmark datasets. We utilize a variety of forecast accuracy metrics, such as Mean Absolute Error(MAE), Root Mean Squared Error(RMSE), and Mean Absolute Percentage Error(MAPE), to provide a comprehensive understanding of the variant model's performance compared to the baseline model. The results for multistep-ahead forecasting are computed based on the difference between the pointwise forecasts and the observed data across the prediction interval and reported in terms of these forecast accuracy metrics. Inside the parentheses, we additionally report the relative percentage difference between the performance of the variant model and the baseline model. We repeated experiments five times to report the average results. Additionally, the forecast horizon for these experiments is set at 12 to study the variant model's ability to handle the complexity of the long-term prediction compared to the baseline model. 
The variant models have lower forecast accuracy and significantly under perform when compared to the benchmark model, as shown in Table \ref{tab:ablationstudy}. We can observe that the spatial inference component of \textbf{EIKF-Net} framework is more important than the temporal inference component for achieving SOTA performance on the benchmark datasets. On the PeMSD8 dataset, the ``\textbf{w/o Spatial}" variant reports a drastic drop in performance w.r.t. baseline model with a significant increase of 15.0$\%$ in the RMSE, 15.17$\%$ in the MAE, and 15.02$\%$ in the MAPE, respectively. Contrariwise, we observe a slightly worse performance  of ``\textbf{w/o Temporal}" variant compared to the baseline model with a marginal rise of 1.59$\%$, 1.63$\%$, and 1.59$\%$ w.r.t RMSE, MAE, and MAPE, respectively. A higher percentage increase in the variant model's error metrics scores compared to the baseline model verifies the relative importance of the mechanism underlying the excluded components of the baseline model. Likewise, we observe similar trends on the PeMSD4 dataset. The ``\textbf{w/o Spatial}" variant compared to the benchmark model, performs significantly worse with an increase of $12.1\%$, $10.4\%$, and $8.5\%$ w.r.t RMSE, MAE, and MAPE-measure. On the contrary, ``\textbf{w/o Temporal}" variant performance declines slightly than the baseline model with an insignificant increase of $0.4\%$, $0.6\%$, $0.7\%$ w.r.t RMSE, MAE, and MAPE score. Thus, the predominant backbone underpinning the \textbf{EIKF-Net} framework is the spatial inference component, which captures the intricate interdependencies among the multiple time series variables while learning the dynamics of interacting systems. It consists of two modules, i.e., explicit graph learning and implicit hypergraph learning modules, which are the cornerstone of our neural forecast architecture.

\vspace{-3mm}
\begin{itemize}
\item The explicit graph representation learning encodes the explicit graph-structured data to capture the time-evolving spatial correlations among the multiple correlated time series variables for modeling the dynamics of complex systems. 
\item Similarly, the joint learning of the latent discrete hypergraph structure and the optimal hypergraph representations capture the higher-order spatial dependencies underlying the time-varying network data. In particular, the hypergraph inference module learns the complex hierarchical structural-dynamic dependencies underlying the MTS data in the sparse discrete hypergraph structure. While simultaneously, the hypergraph representation learning backbone encodes the structural spatio-temporal inductive biases for modeling the continuous-time nonlinear dynamics of interconnected network of sensors. 
\end{itemize}

\vspace{-3mm}
The spatial inference component performs a convex combination of node-level and hypernode-level representations of an explicit graph and an implicit hypergraph, respectively to capture the evolutionary and multi-scale interactions among the time series variables. The mixup representations generated by the spatial inference component disentangle the various latent aspects underneath the spatio-temporal data for improved multistep-ahead forecast accuracy. The substantial decrease in performance of ``\textbf{w/o Spatial}" variant w.r.t. baseline model demonstrates that the spatial inference component is indispensable.

\vspace{-4mm}
\begin{table}[ht!]
\setlength{\tabcolsep}{0.35em} 
\renewcommand\arraystretch{1.335} 
\centering
\hspace*{-0.5cm} \resizebox{1.06\textwidth}{!}{
\begin{tabular}{c|ccc|ccc|ccc|ccc}
\hline
\multirow{2}{*}{\textbf{Method}} & \multicolumn{3}{c|}{{\ul \textbf{PeMSD3}}}       & \multicolumn{3}{c|}{{\ul \textbf{PeMSD4}}}              \\ \cline{2-7} 
                                 & \textbf{RMSE}  & \textbf{MAE}   & \textbf{MAPE}  & \textbf{RMSE}  & \textbf{MAE}  & \textbf{MAPE}    \\ \hline
\textbf{EIKF-Net}              & \textbf{23.11} & \textbf{14.84} & \textbf{13.07} & \textbf{28.88} & \textbf{19.20} & \textbf{11.86}   \\ \hline
\textbf{w/o Spatial}           & 26.23${\color{black} (13.5 \% \uparrow)}$          & 16.73${\color{black} (12.8 \% \uparrow)}$          & 14.18${\color{black} (8.5 \% \uparrow)}$          & 32.38${\color{black} (12.1 \% \uparrow)}$          & 21.20${\color{black} (10.4 \% \uparrow)}$          & 12.87${\color{black} (8.5 \% \uparrow)}$                    \\
\textbf{w/o Temporal}            & 23.10${\color{black} (0.04 \% \downarrow)}$           & 14.95${\color{black} (0.74 \% \uparrow)}$          & 13.14${\color{black} (0.53 \% \uparrow)}$          & 29.01${\color{black} (0.4 \% \uparrow)}$        & 19.32${\color{black} (0.6 \% \uparrow)}$         & 11.95${\color{black} (0.7 \% \uparrow)}$                    \\
\textbf{w/o Explicit Graph}    & 24.63${\color{black} (6.6 \% \uparrow)}$          & 15.75${\color{black} (6.1 \% \uparrow)}$          & 13.27${\color{black} (1.5 \% \uparrow)}$          & 30.62${\color{black} (6.0 \% \uparrow)}$        & 20.16${\color{black} (5.0 \% \uparrow)}$        & 12.05${\color{black} (1.6 \% \uparrow)}$                   \\
\textbf{w/o Implicit Hypergraph}    & 23.84${\color{black} (3.2 \% \uparrow)}$          & 15.26${\color{black} (2.8 \% \uparrow)}$          & 13.47${\color{black} (3.1 \% \uparrow)}$          & 29.35${\color{black} (1.6 \% \uparrow)}$         & 19.50${\color{black} (1.5 \% \uparrow)}$        & 12.07${\color{black} (1.7 \% \uparrow)}$                    \\ \hline
\end{tabular}
}

\vspace{1.5mm}

\setlength{\tabcolsep}{0.35em} 
\renewcommand\arraystretch{1.335} 
\centering
\hspace*{-0.5cm} \resizebox{1.06\textwidth}{!}{
\begin{tabular}{c|ccc|ccc|ccc|ccc}
\hline
\multirow{2}{*}{\textbf{Method}} & \multicolumn{3}{c|}{{\ul \textbf{PeMSD7}}}      & \multicolumn{3}{c}{{\ul \textbf{PeMSD8}}}       \\ \cline{2-7} 
                                 & \textbf{RMSE}  & \textbf{MAE}   & \textbf{MAPE} & \textbf{RMSE}  & \textbf{MAE}   & \textbf{MAPE} \\ \hline
\textbf{EIKF-Net}              &  \textbf{31.85} & \textbf{21.32} & \textbf{8.96} & \textbf{23.34} & \textbf{15.36} & \textbf{8.72} \\ \hline
\textbf{w/o Spatial}            & 36.56${\color{black} (14.8 \% \uparrow)}$          & 24.13${\color{black} (13.18 \% \uparrow)}$          & 10.36${\color{black} (5.63 \% \uparrow)}$         & 26.84${\color{black} (15.0 \% \uparrow)}$          & 17.69${\color{black} (15.17 \% \uparrow)}$          & 10.03${\color{black} (15.02 \% \uparrow)}$         \\
\textbf{w/o Temporal}            & 31.81${\color{black} (0.13 \% \downarrow)}$         & 21.33${\color{black} (0.05 \% \uparrow)}$         & 8.98${\color{black} (0.2 \% \uparrow)}$         & 23.71${\color{black} (1.59 \% \uparrow)}$          & 15.61${\color{black} (1.63 \% \uparrow)}$          & 8.89${\color{black} (1.59 \% \uparrow)}$          \\
\textbf{w/o Explicit Graph}    & 34.10${\color{black} (7.0 \% \uparrow)}$         & 22.67${\color{black} (6.3 \% \uparrow)}$         & 9.42${\color{black} (5.1 \% \uparrow)}$         & 25.15${\color{black} (7.75 \% \uparrow)}$          & 16.46${\color{black} (7.16 \% \uparrow)}$          & 9.28${\color{black} (6.42 \% \uparrow)}$          \\
\textbf{w/o Implicit Hypergraph}       & 32.70${\color{black} (2.6 \% \uparrow)}$           & 21.86${\color{black} (2.5 \% \uparrow)}$          & 9.25${\color{black} (3.2 \% \uparrow)}$          & 23.89${\color{black} (2.36 \% \uparrow)}$          & 15.73${\color{black} (2.41 \% \uparrow)}$          & 8.93${\color{black} (2.41 \% \uparrow)}$          \\ \hline
\end{tabular}
}

\vspace{1.5mm}

\setlength{\tabcolsep}{0.35em} 
\renewcommand\arraystretch{1.335} 
\centering
\hspace*{-0.5cm} \resizebox{1.06\textwidth}{!}{
\begin{tabular}{c|ccc|ccc|ccc}
\hline
\multirow{2}{*}{\textbf{Method}} & \multicolumn{3}{c|}{{\ul \textbf{PeMSD7(M)}}} & \multicolumn{3}{c|}{{\ul \textbf{PEMS-BAY}}}    \\ \cline{2-7} 
                                 & \textbf{RMSE} & \textbf{MAE}  & \textbf{MAPE} & \textbf{RMSE} & \textbf{MAE}  & \textbf{MAPE}   \\ \hline
\textbf{EIKF-Net}              & \textbf{5.25} & \textbf{2.99} & \textbf{6.35} & \textbf{3.08} & \textbf{1.69} & \textbf{3.14}  \\ \hline
\textbf{w/o Spatial}           & 5.75${\color{black} (9.52 \% \uparrow)}$          & 3.21${\color{black} (7.35 \% \uparrow)}$          & 6.92${\color{black} (8.97 \% \uparrow)}$          & 3.43${\color{black} (11.36 \% \uparrow)}$          & 1.81${\color{black} (7.1 \% \uparrow)}$          & 3.34${\color{black} (6.36 \% \uparrow)}$                    \\
\textbf{w/o Temporal}            & 5.26${\color{black} (0.19 \% \uparrow)}$          & 3.02${\color{black} (1.0 \% \uparrow)}$           & 6.41${\color{black} (0.94 \% \uparrow)}$          & 3.41${\color{black} (10.71 \% \uparrow)}$          & 1.73${\color{black} (2.36 \% \uparrow)}$          & 3.21${\color{black} (2.22 \% \uparrow)}$                 \\
\textbf{w/o Explicit Graph}    & 5.19${\color{black} (1.14 \% \downarrow)}$          & 2.97${\color{black} (0.66 \% \downarrow)}$          & 6.30${\color{black} (0.78 \% \downarrow)}$           & 3.45${\color{black} (12.0 \% \uparrow)}$          & 1.73${\color{black} (2.36 \% \uparrow)}$          & 3.21${\color{black} (2.22 \% \uparrow)}$                   \\
\textbf{w/o Implicit Hypergraph}    & 5.36${\color{black} (2.09 \% \uparrow)}$          & 3.14${\color{black} (5.01 \% \uparrow)}$          & 6.68${\color{black} (5.19 \% \uparrow)}$          & 3.51${\color{black} (13.96 \% \uparrow)}$          & 1.79${\color{black} (5.91 \% \uparrow)}$          & 3.30${\color{black} (5.09 \% \uparrow)}$                     \\ \hline
\end{tabular}
}

\vspace{1.5mm}

\setlength{\tabcolsep}{0.35em} 
\renewcommand\arraystretch{1.335} 
\centering
\hspace*{-0cm} \resizebox{0.635\textwidth}{!}{
\begin{tabular}{c|ccc|ccc|ccc}
\hline
\multirow{2}{*}{\textbf{Method}} & \multicolumn{3}{c}{{\ul \textbf{METR-LA}}}    \\ \cline{2-4} 
                                 & \textbf{RMSE} & \textbf{MAE}  & \textbf{MAPE} \\ \hline
\textbf{EIKF-Net}              &  \textbf{8.76} & \textbf{4.87} & \textbf{9.99} \\ \hline
\textbf{w/o Spatial}          &  9.09${\color{black} (3.76 \% \uparrow)}$          & 5.04${\color{black} (3.49 \% \uparrow)}$          & 10.09${\color{black} (1.0 \% \uparrow)}$         \\
\textbf{w/o Temporal}                  & 8.80${\color{black} (0.45 \% \uparrow)}$           & 4.90${\color{black} (0.61 \% \uparrow)}$           & 10.05${\color{black} (0.6 \% \uparrow)}$         \\
\textbf{w/o Explicit Graph}             & 8.94${\color{black} (2.05 \% \uparrow)}$          & 5.01${\color{black} (2.87 \% \uparrow)}$          & 10.23${\color{black} (2.4 \% \uparrow)}$         \\
\textbf{w/o Implicit Hypergraph}    & 9.01${\color{black} (2.85 \% \uparrow)}$          & 5.07${\color{black} (4.1 \% \uparrow)}$          & 10.36${\color{black} (3.7 \% \uparrow)}$         \\ \hline
\end{tabular}
}
\vspace{-2mm}
\caption{Ablation study results on benchmark datasets}
\label{tab:ablationstudy}
\end{table}

\vspace{-5mm}
To better understand the learning mechanism of the spatial inference component of the \textbf{EIKF-Net} framework, we conduct additional experiments to demonstrate the effectiveness of graph and hypergraph learning modules. The ``\textbf{w/o Explicit Graph}" variant performed worse than the benchmark model with an increase of $6.0\%$, $5.0\%$, $1.6\%$ on PeMSD4; $7.75\%$, $7.16\%$, $6.42\%$ on PeMSD8 w.r.t the RMSE, MAE, and MAPE metrics, respectively. The results suggest that learning on a predefined graph structure helps in improving the framework performance. It might be attributed to the sole reason that the predefined graph structure provides important interrelations and constraints that assist the framework in learning more expressive representations, leading to better performance on the multi-horizon forecasting task. The ``\textbf{w/o Implicit HyperGraph}" variant performed worse than the baseline model with a marginal increase of $1.6\%$, $1.5\%$, and $1.7\%$ on PeMSD4; $2.36\%$, $2.41\%$, $2.41\%$ on PeMSD8 w.r.t. RMSE, MAE, and MAPE metrics, respectively. The results support the rationale of incorporating the hypergraph learning module in the framework can improve its ability to capture long-range scale-specific correlations among the time series variables, leading to more accurate forecasts. Interestingly, on PeMSD3, PeMSD4, PeMSD7, PeMSD7M, and PeMSD8 benchmark datasets, the ablation studies suggest learning on the predefined graph is more beneficial than on implicit hypergraph. In contrast, on PeMS-BAY and METR-LA benchmark datasets, the results suggest learning on implicit hypergraph structure is more helpful compared to learning on explicit graph. More results of the ablation study on benchmark datasets are provided in Table \ref{tab:ablationstudy}. The results reported in Table \ref{tab:ablationstudy} demonstrate that our proposed framework generalizes well despite the complexity of patterns across the broad spectrum of benchmark datasets and scales well on large-scale graph datasets. In summary, the ablation studies corroborate the hypothesis of the joint optimization of the spatial-temporal learning components for improved performance on the MTSF.

\vspace{-4mm}
\subsection{Additional study on the significance of hypergraph inference approach.}
\vspace{-3mm}
In this section, we study the impact of jointly learning the implicit hypergraph structure \& hypergraph representations from MTS data on the forecast accuracy compared to jointly learning the implicit graph structure and its corresponding graph representations. In the case of implicit graph structure learning, the goal is to capture hidden relationships and dependencies among the different variables in the MTS data represented in a graph structure that might not have been apparent with domain expertise knowledge. STGNNs perform message-passing schemes on the inferred implicit graph topology to learn the more expressive graph representations capturing the patterns and relationships underlying the MTS data for better forecasts. In the recent past, there has been an increased focus on developing methods for joint learning of discrete graph structure and representations for forecasting on MTS data. The existing methods include GTS(Graph for Time Series, \cite{shang2021discrete}), Graph Deviation Network(GDN, \cite{deng2021graph}), and MTS forecasting with GNNs(MTGNN, \cite{wu2020connecting}). We design a set of variant models by substituting the hypergraph inference and representation learning with the corresponding implicit graph inference and representation learning methods mentioned earlier. We conduct ablation experiments on the PeMSD4 and PeMSD8 benchmark datasets to evaluate the effectiveness of each design choice. We set the forecast horizon as 12. This study can help to understand the specific design choice's effectiveness, which can contribute to improved overall framework performance on long-term prediction tasks. The ablated variant, ``w/ GTS: EIKF-Net" refers to a variant of the \textbf{EIKF-Net} framework with the GTS module, but does not include the implicit hypergraph learning module. Similarly, ``w/ GDN: EIKF-Net" and ``w/ MTGNN: EIKF-Net" are also ablated variants.

\vspace{-3mm}
\begin{table}[ht!]
\setlength{\tabcolsep}{0.35em} 
\renewcommand\arraystretch{1.115} 
\centering
\hspace*{+0.25cm} \resizebox{0.65\textwidth}{!}{
\begin{tabular}{c|ccc|ccc|ccc|ccc}
\hline
\multirow{2}{*}{\textbf{Method}} & \multicolumn{3}{c|}{{\ul \textbf{PeMSD4}}}       & \multicolumn{3}{c|}{{\ul \textbf{PeMSD8}}}              \\ \cline{2-7} 
                                 & \textbf{RMSE}  & \textbf{MAE}   & \textbf{MAPE}  & \textbf{RMSE}  & \textbf{MAE}  & \textbf{MAPE}    \\ \hline
\textbf{EIKF-Net}              & \textbf{23.11} & \textbf{14.84} & \textbf{13.07} & \textbf{28.88} & \textbf{19.20} & \textbf{11.86}   \\ \hline
\textbf{w/ GTS : EIKF-Net}              & 24.93 & 15.93 & 13.83 & 30.07 & 20.13 & 12.43   \\ \hline
\textbf{w/ GDN: EIKF-Net}              & 25.07 & 16.23 & 14.29 & 31.68 & 21.34 & 12.93   \\ \hline
\textbf{w/ MTGNN : EIKF-Net}              & 25.42 & 16.37 & 14.78 & 31.93 & 21.76 & 13.15   \\ \hline
\end{tabular}
}
\vspace{-3mm}
\caption{Additonal ablation study results on PeMSD4 and PeMSD8}
\label{tab:add1_ablationstudy}
\end{table}

\vspace{-5mm}
Table \ref{tab:add1_ablationstudy} reports the experimental results, which indicates that the hypergraph learning module is indispensable. In summary, learning to simultaneously infer the latent hypergraph interaction structure and modeling the time-evolving dynamics through the hypergraph representation learning schemes in an end-to-end framework can be significant. It allows for modeling complex higher-order relationships among multiple variables, which leads to more accurate hypergraph representations for analyzing MTS data and thus has lower forecast errors.

\vspace{-4mm}
\subsection{Additional study on the impact of hypergraph learning paradigms.}
\vspace{-3mm}
Our framework performs joint learning of hypergraph structure and representations through two learning units: hypergraph structure learning (i.e., through embedding-based similarity metric learning) and hypergraph representation learning. The objective is to learn the optimal hypergraph structures and representations for the downstream forecasting task.
There exist different learning paradigms for optimizing the two separate learning units, that includes joint learning(JL),  adaptive learning(AL, \cite{wu2022graph, pilco2019graph}), and iterative learning(IL, \cite{chen2019deep}) of hypergraph structures and representations, respectively. We conduct ablation experiments to investigate the impact of learning paradigms on the overall performance of the framework. The ablated variants, ``w/ AL: EIKF-Net" and ``w/ IL: EIKF-Net" refers to variants of the \textbf{EIKF-Net} framework that utilizes the Adaptive Learning(AL)  and Iterative Learning(IL) module respectively instead of the Joint Learning(JL) paradigm. Table \ref{tab:add2_ablationstudy} reports the experimental results. However, the results indicate the strong performance of JL paradigm and highlights its advantage compared to other learning paradigms.

\vspace{-4mm}
\begin{table}[ht]
\setlength{\tabcolsep}{0.35em} 
\renewcommand\arraystretch{1.105} 
\centering
\hspace*{-0.5cm} \resizebox{0.625\textwidth}{!}{
\begin{tabular}{c|ccc|ccc|ccc|ccc}
\hline
\multirow{2}{*}{\textbf{Method}} & \multicolumn{3}{c|}{{\ul \textbf{PeMSD4}}}       & \multicolumn{3}{c|}{{\ul \textbf{PeMSD8}}}              \\ \cline{2-7} 
                                 & \textbf{RMSE}  & \textbf{MAE}   & \textbf{MAPE}  & \textbf{RMSE}  & \textbf{MAE}  & \textbf{MAPE}    \\ \hline
\textbf{EIKF-Net}              & \textbf{23.11} & \textbf{14.84} & \textbf{13.07} & \textbf{28.88} & \textbf{19.20} & \textbf{11.86}   \\ \hline
\textbf{w/ AL : EIKF-Net}              & 24.37 & 15.42 & 14.25 & 30.07 & 20.31 & 12.47   \\ \hline
\textbf{w/ IL : EIKF-Net}              & 25.61 & 16.07 & 14.79 & 31.24 & 21.19 & 13.05   \\ \hline
\end{tabular}
}
\vspace{-2mm}
\caption{Additonal ablation study results on learning paradigms}
\label{tab:add2_ablationstudy}
\end{table}

\vspace{-7mm}
\subsection{Pointwise prediction error for multi-horizon forecasting}
\begin{figure}[ht!]
\centering
\hspace*{-0.05cm}\resizebox{0.98\textwidth}{!}{
\subfloat[MAE on PeMSD3]{\includegraphics[width=50mm]{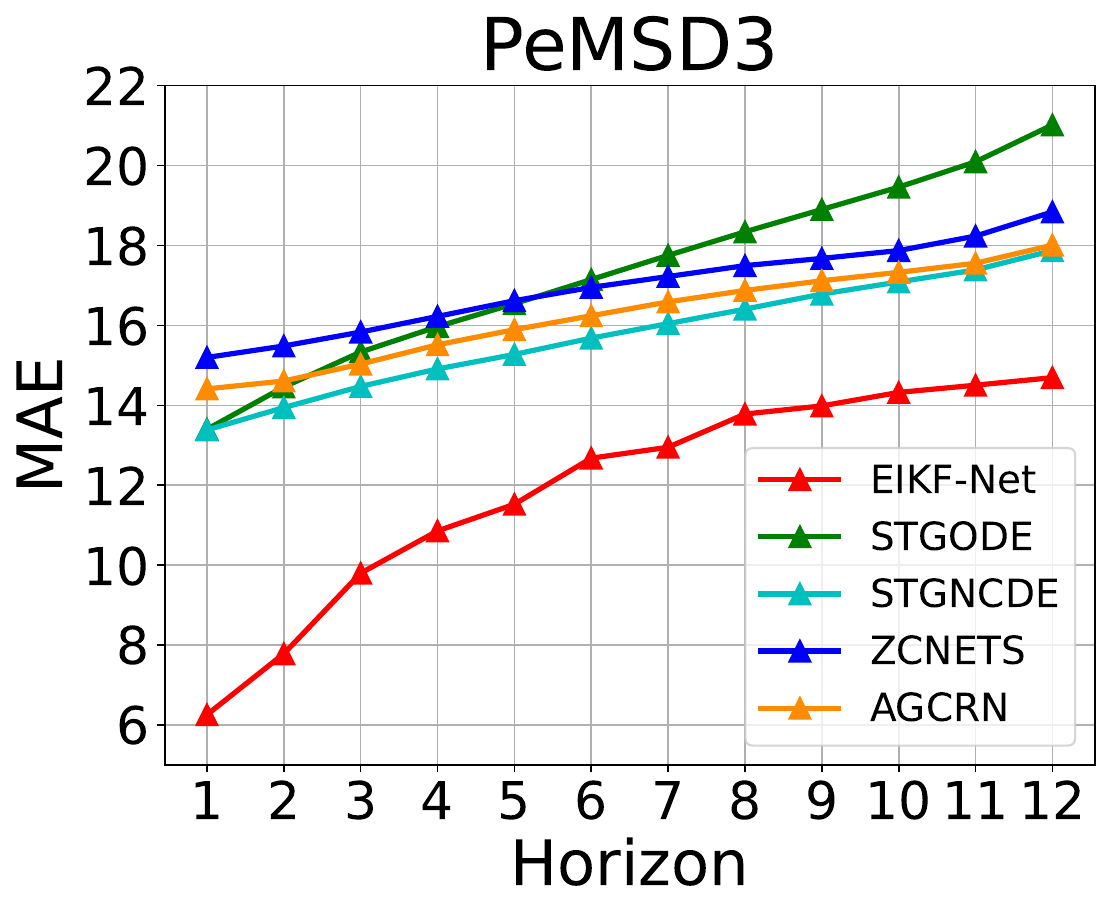}}
\subfloat[MAPE on PeMSD3]{\includegraphics[width=50mm]{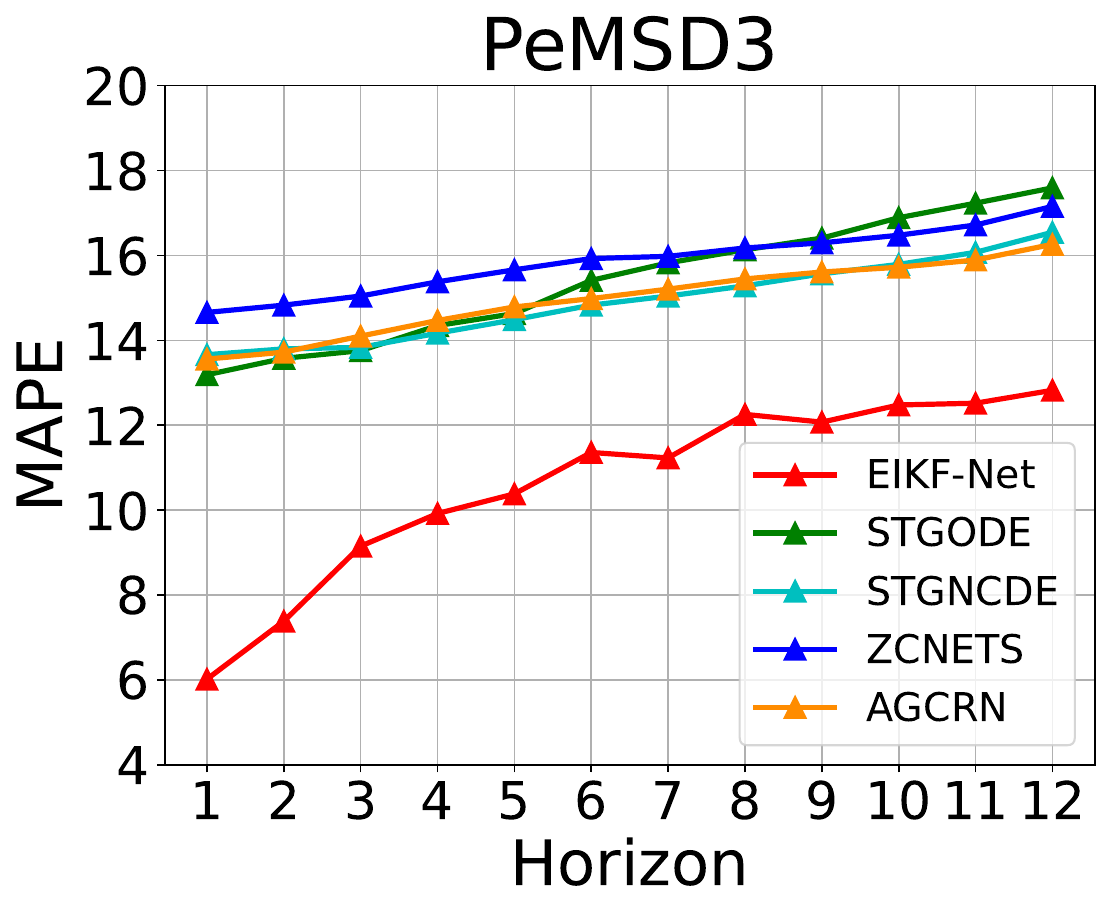}}
\subfloat[RMSE on PeMSD3]{\includegraphics[width=50mm]{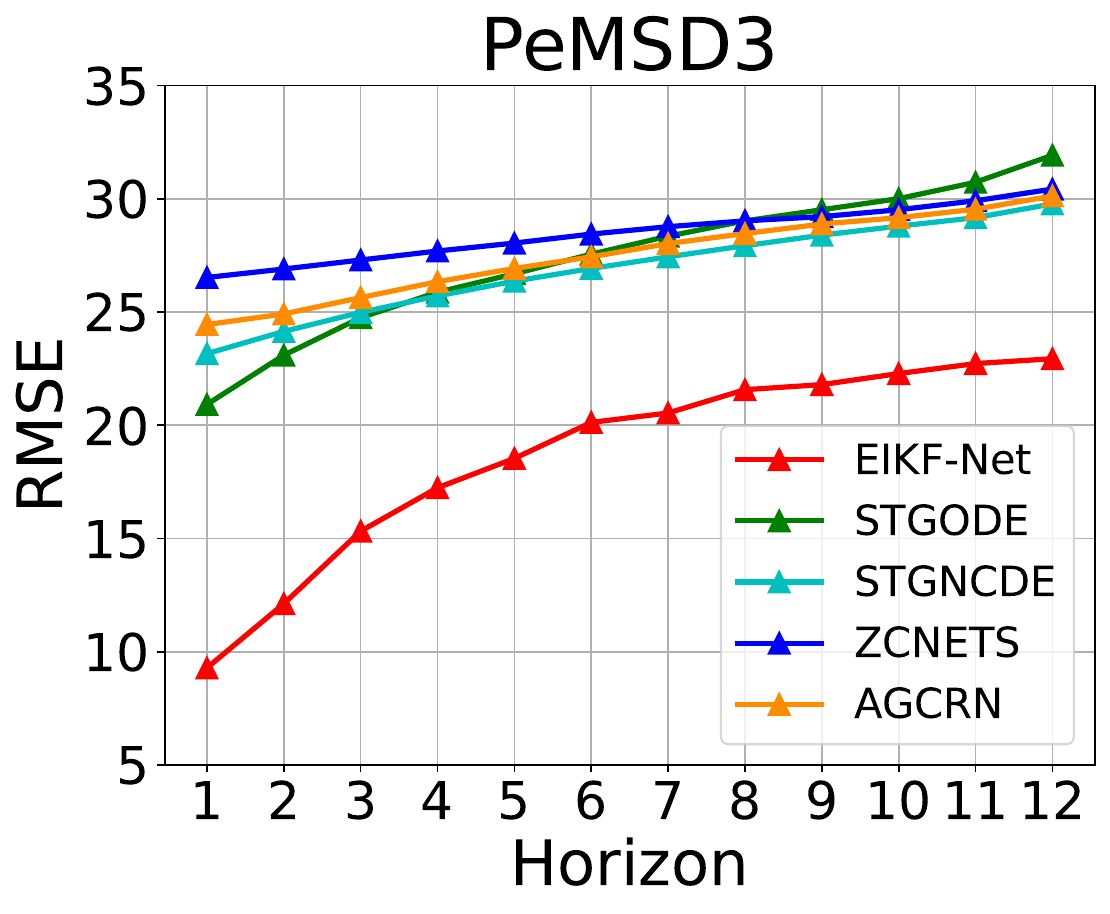}}
}\\[-2ex]
\hspace*{-0.05cm}\resizebox{0.98\textwidth}{!}{
\subfloat[MAE on PeMSD4]{\includegraphics[width=50mm]{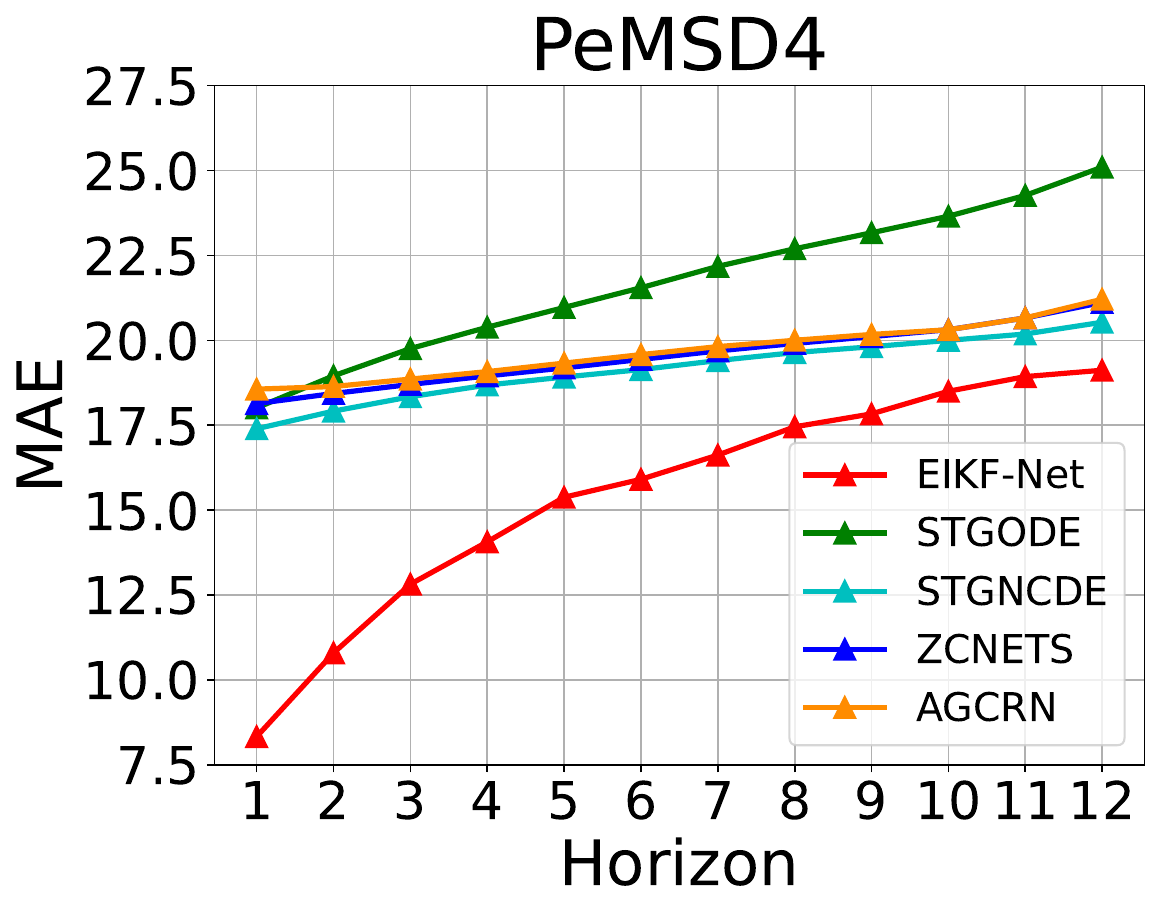}}
\subfloat[MAPE on PeMSD4]{\includegraphics[width=50mm]{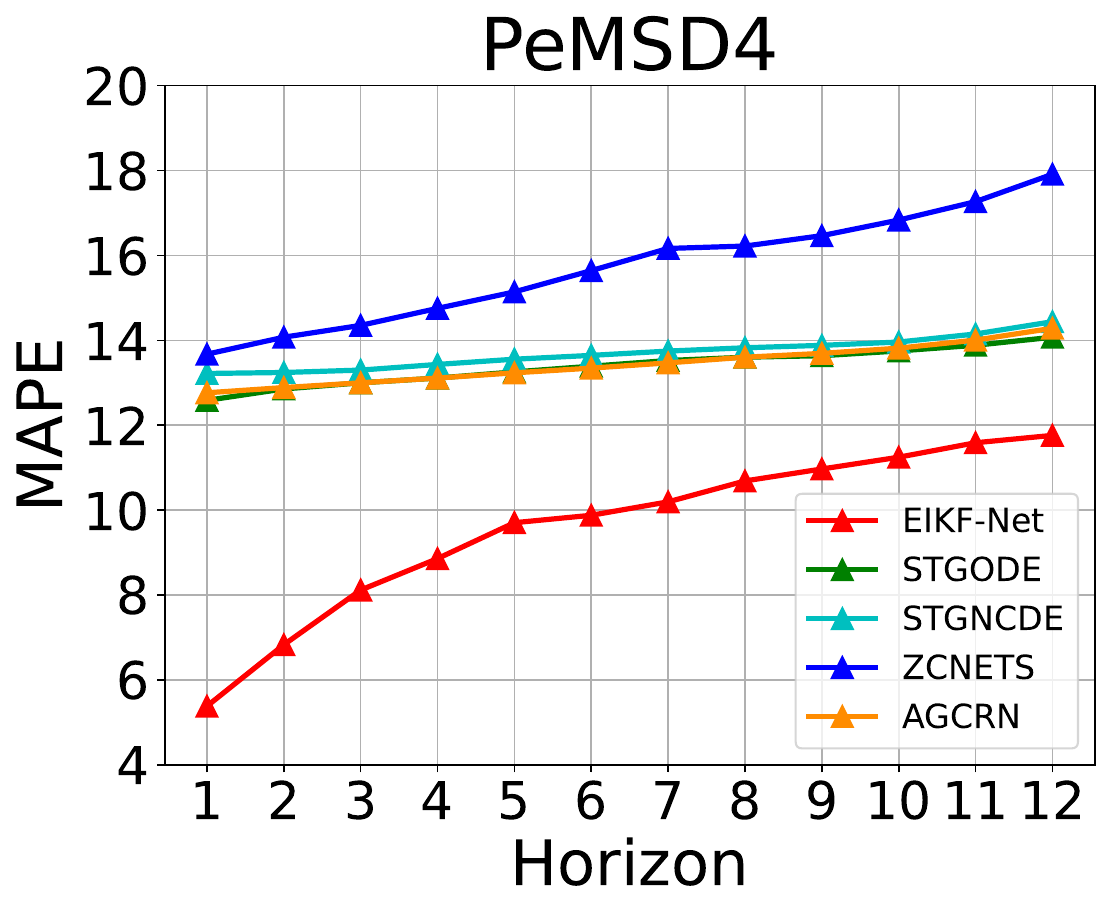}}
\subfloat[RMSE on PeMSD4]{\includegraphics[width=50mm]{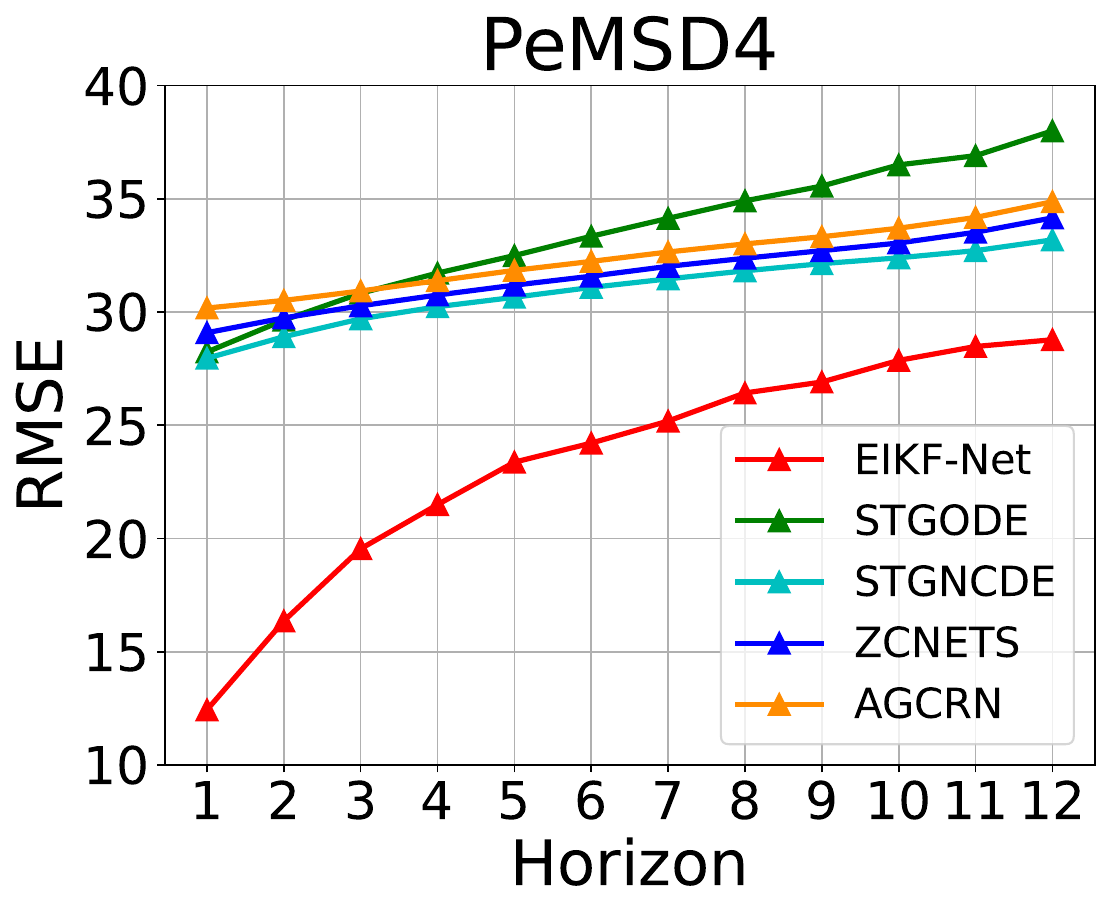}}
}\\[-2ex]
\hspace*{-0.05cm}\resizebox{0.98\textwidth}{!}{
\subfloat[RMSE on PeMSD7]{\includegraphics[width=50mm]{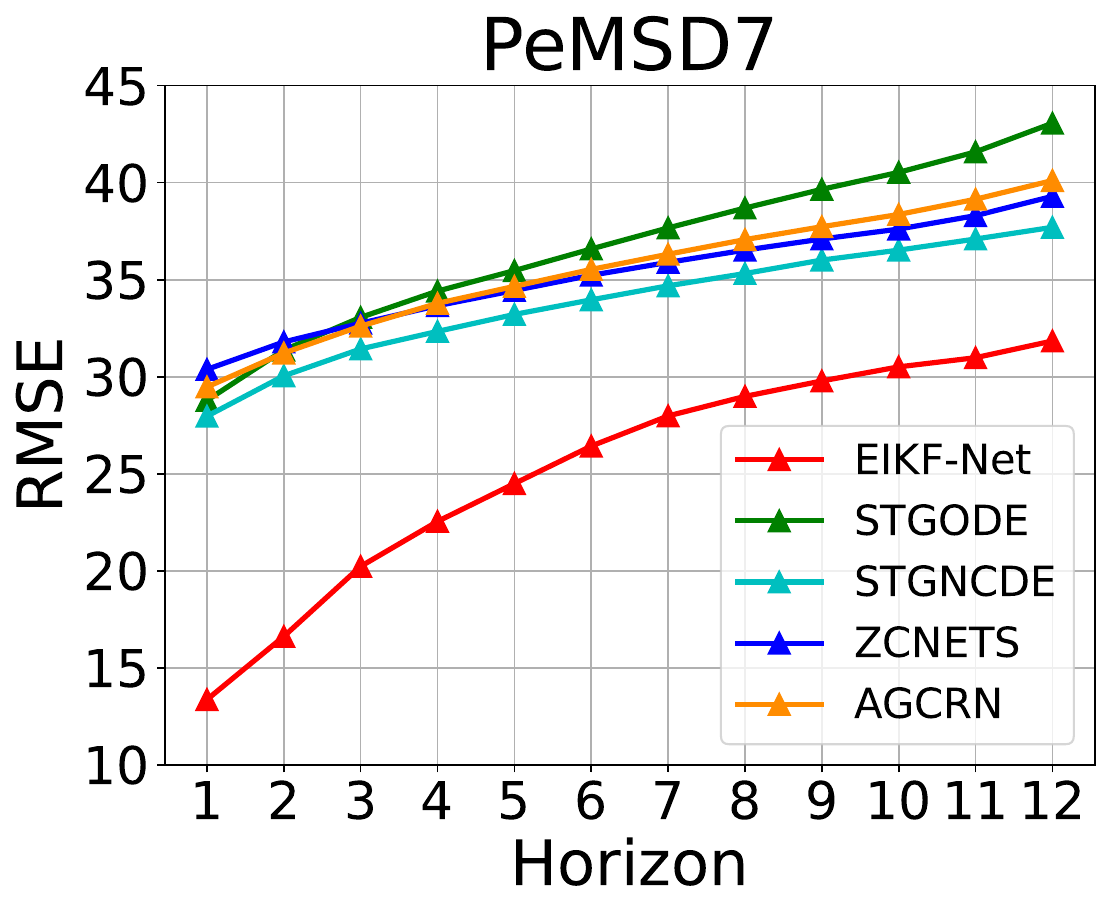}}
\subfloat[MAE on PeMSD7]{\includegraphics[width=50mm]{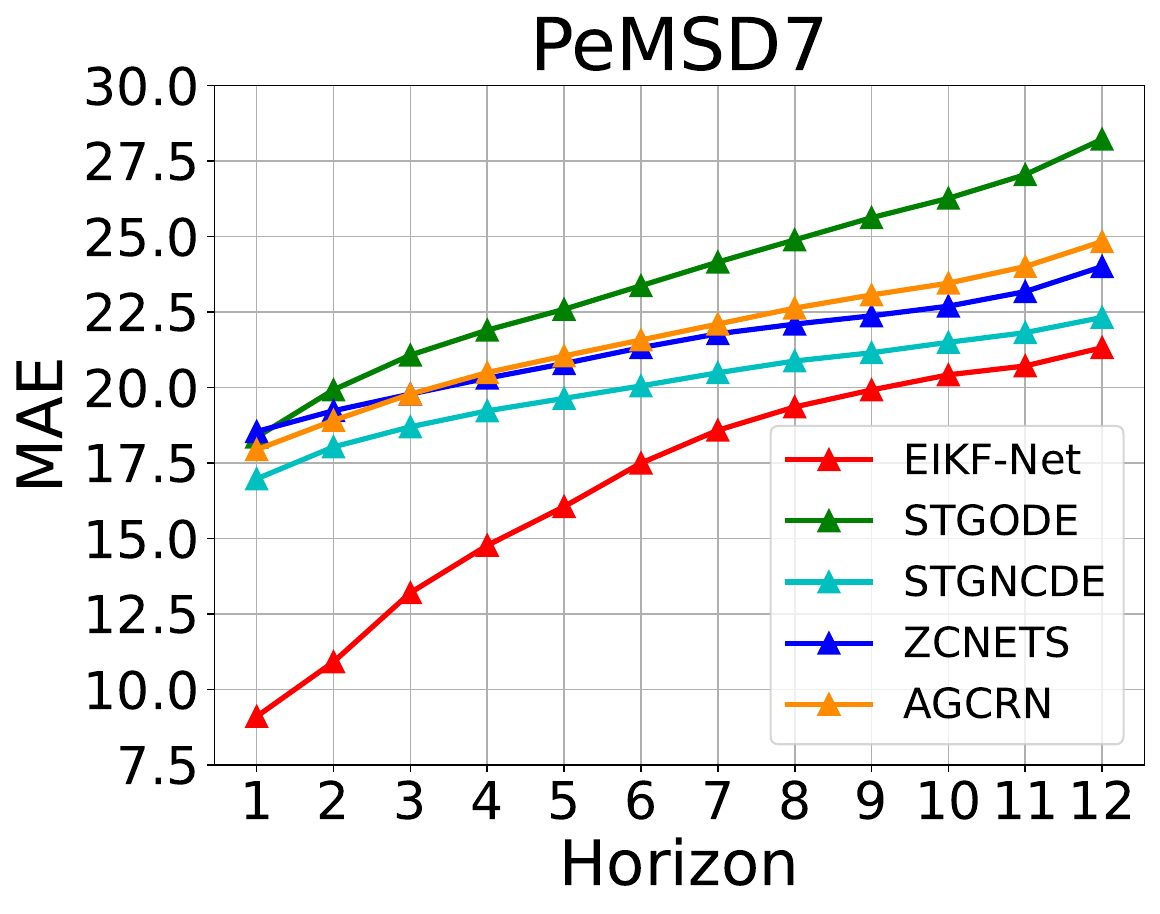}}
\subfloat[MAPE on PeMSD7]{\includegraphics[width=50mm]{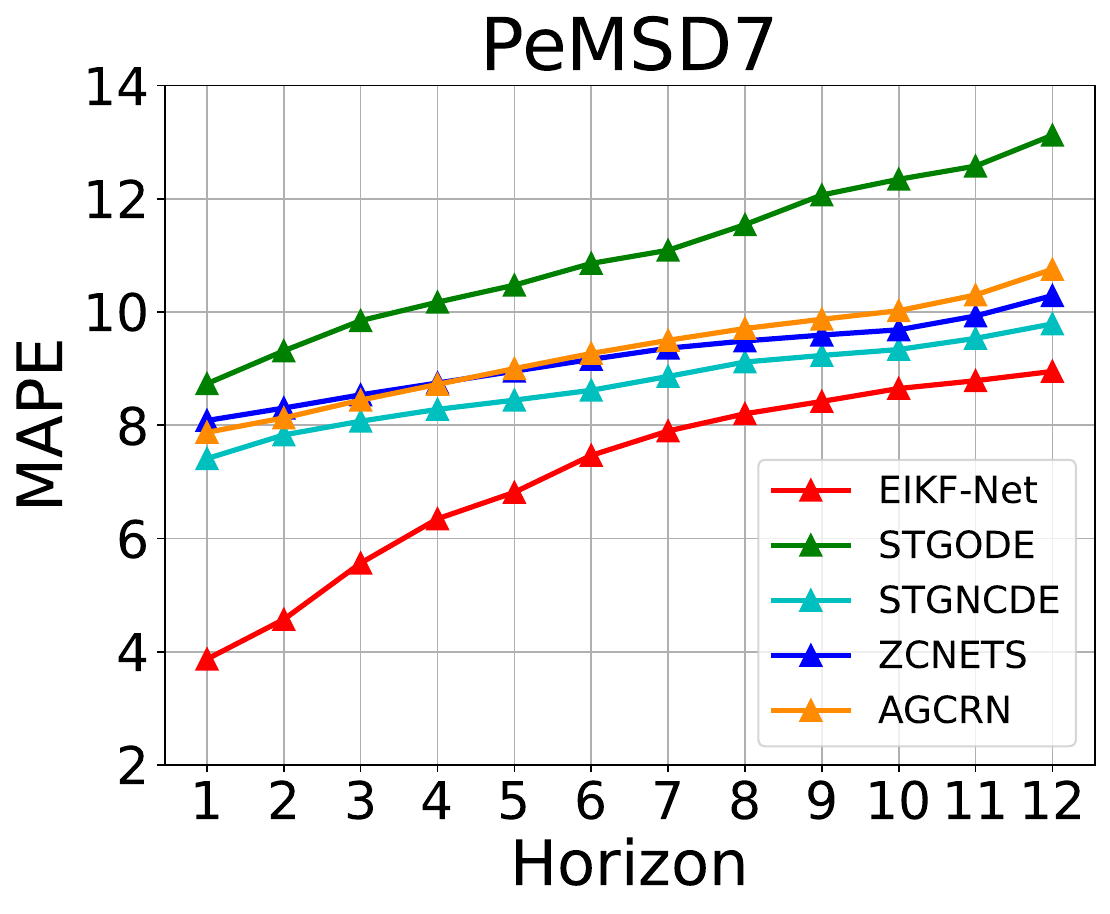}}
}\\[-2ex]
\hspace*{-0.05cm}\resizebox{0.975\textwidth}{!}{
\subfloat[RMSE on PeMSD7(M)]{\includegraphics[width=50mm]{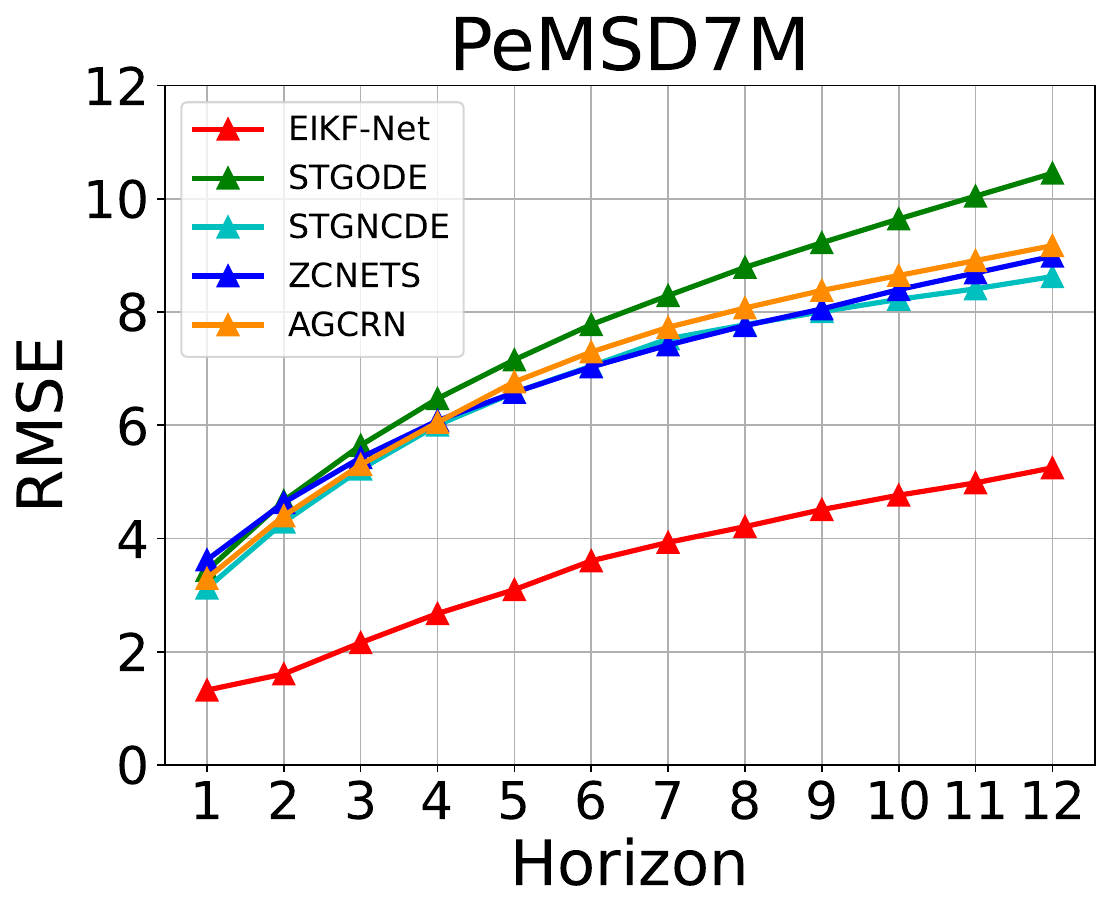}}
\subfloat[MAE on PeMSD7(M)]{\includegraphics[width=50mm]{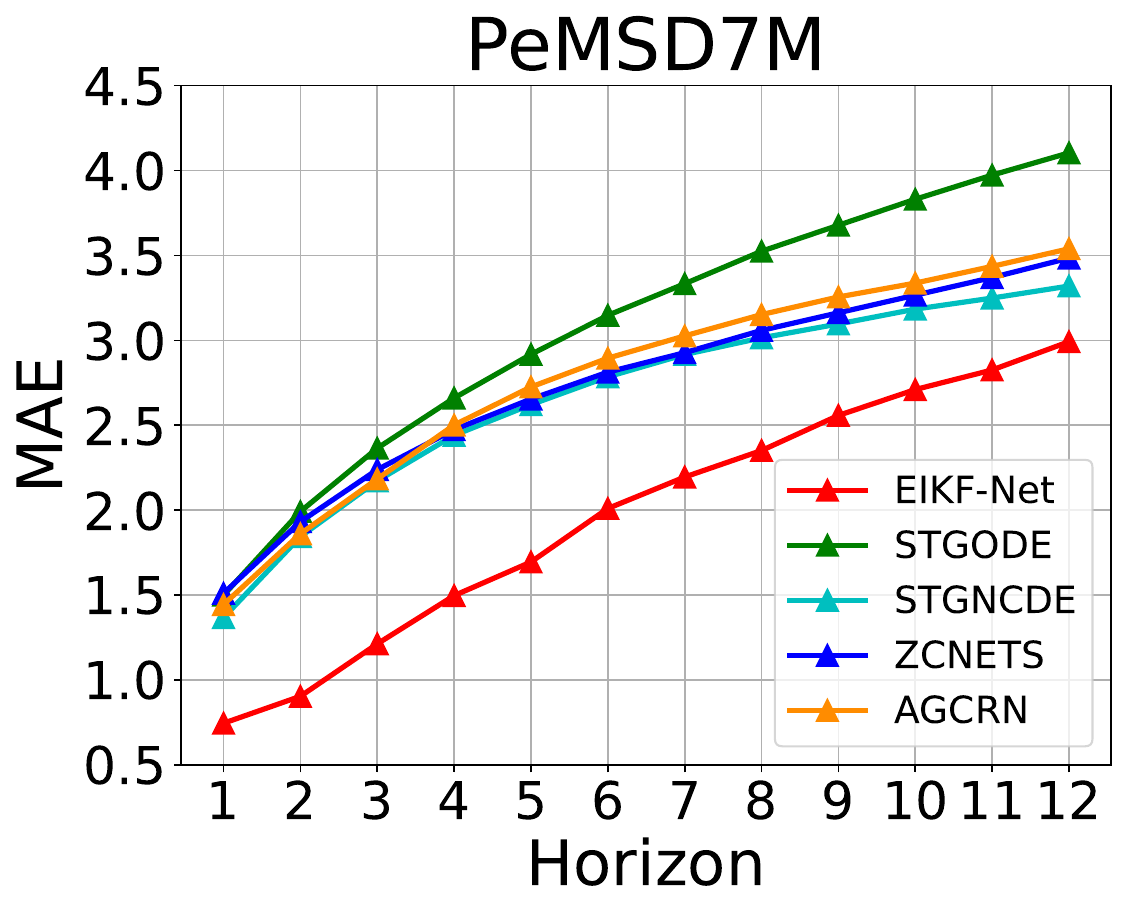}}
\subfloat[MAPE on PeMSD7(M)]{\includegraphics[width=50mm]{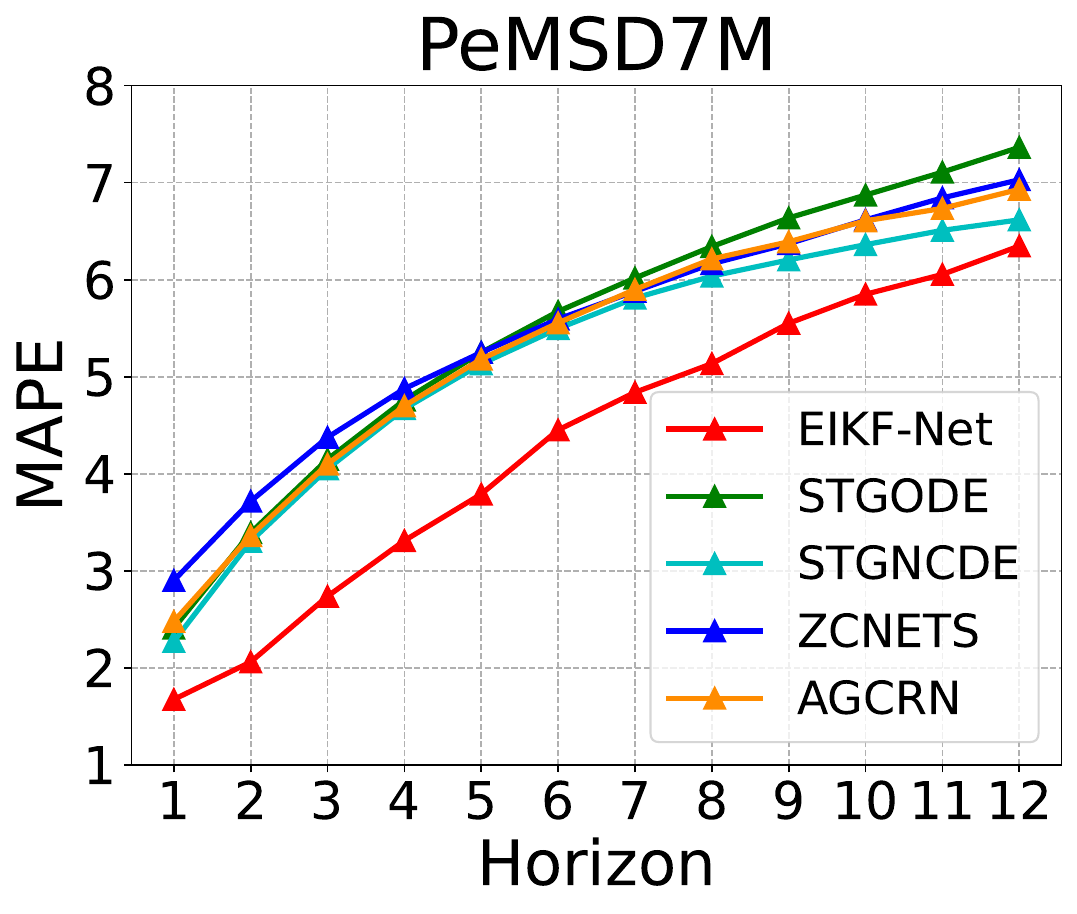}}
}\\[-1ex]
\hspace*{-0.05cm}\resizebox{0.975\textwidth}{!}{
\subfloat[MAE on PeMSD8]{\includegraphics[width=50mm]{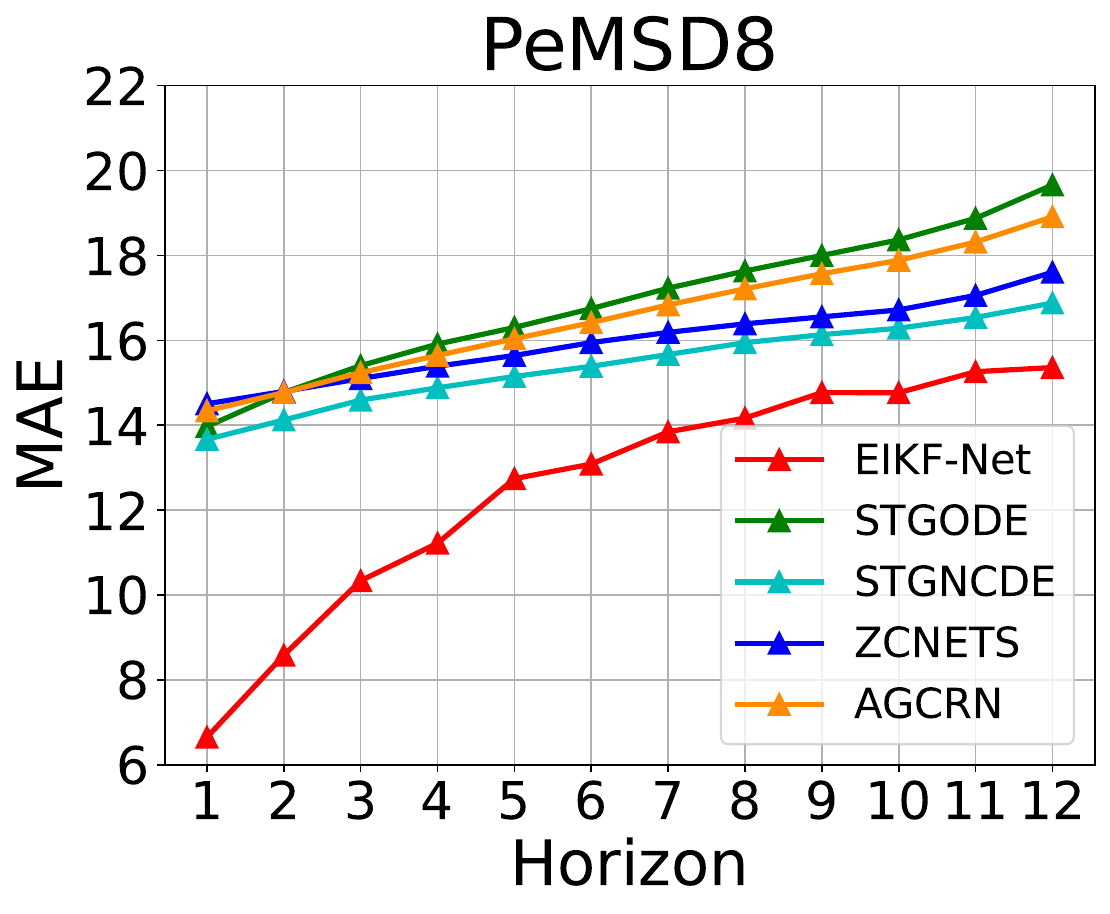}}
\subfloat[MAPE on PeMSD8]{\includegraphics[width=50mm]{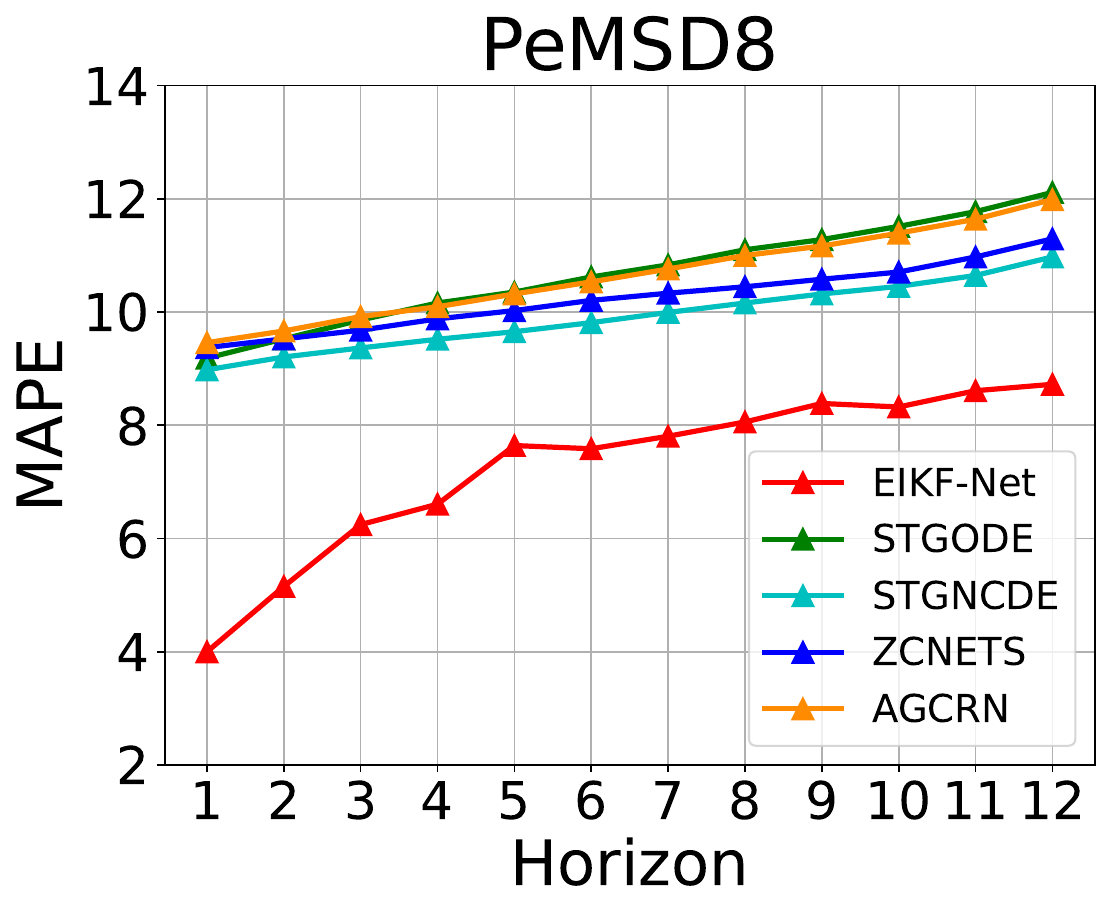}}
\subfloat[RMSE on PeMSD8]{\includegraphics[width=50mm]{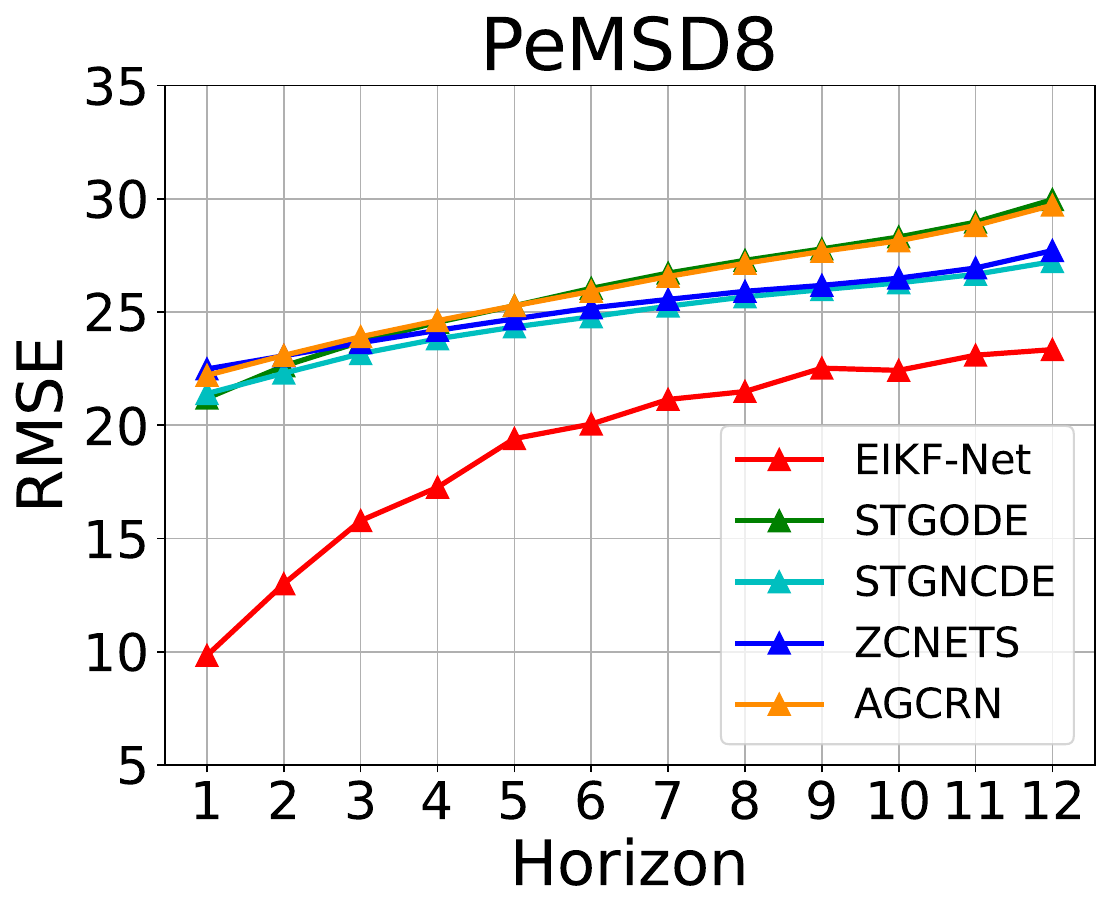}}
}\\[-1.25ex]
\caption{Pointwise prediction error for multi-horizon forecasting tasks on benchmark datasets.}
\label{fig:ppeh1}
\end{figure}

\vspace{-3mm}
Figure \ref{fig:ppeh1} shows the proposed neural forecasting framework(\textbf{EIKF-Net}) multistep-ahead forecasts error at each horizon on the benchmark datasets. Lower RMSE, MAPE, and MAE indicate better model performance on MTSF task. Across all the prediction horizons, our neural forecasting architecture has reduced forecast errors more than the baselines. Our proposed neural forecast framework is particularly well-suited to effectively exploit the relational inductive biases to capture the nonlinear spatio-temporal dependencies in the structured data to improve forecast accuracy.

\vspace{-7mm}

\vspace{-4mm}
\subsection{Irregular Time Series Forecasting}
\vspace{-2mm}
Large, complex interconnected sensor networks span a diverse range of real-world applications and suffer from inherent drawbacks of low-data quality due to inevitable and pervasive intermittent sensor failure, faulty sensors, etc., during the data acquisition process. We simulate the data availability/missingness with two different types of missingness patterns(\cite{roth2022forecasting, cini2021multivariate}) to evaluate the \textbf{EIKF-Net} framework effectiveness in handling the missing data. The simulation methods mimic the missingness patterns occurring in continuous time with asynchronous temporal patterns commonly observed in real-world data of large, complex sensor networks. Firstly, the point-missing pattern simulates sensor failure by randomly dropping observations of each variable within a given historical window, with missing ratios ranging from $10\%$ to $50\%$. Secondly, the block-missing pattern simulates sensor failure by randomly masking out available data for each sensor within a given historical window, with missing ratios ranging from $10\%$ to $50\%$. In addition, we simulate a sensor failure with probability $p_{failure} = 0.15\%$ for generating blocks of missing MTS data. Our tailored deep learning-based framework consists of spatial and temporal inference components to process and analyze MTS data characterized by complex dependencies and relationships between variables that change over time. We conduct further research to investigate the importance of different components of the \textbf{EIKF-Net} framework on multi-horizon forecasting accuracy when dealing with irregular and missing data. We conduct an additional ablation study on the benchmark model(\textbf{EIKF-Net}) by removing either the spatial or temporal learning components to evaluate their individual contribution to the forecasting accuracy in missing data scenarios. These studies demonstrate the robustness and reliability of the \textbf{EIKF-Net} framework in real-world applications where missing data are ubiquitous. 

\vspace{-3mm}
\begin{table}[ht!]
\setlength{\tabcolsep}{0.3em} 
\renewcommand\arraystretch{1.35} 
\centering
 \resizebox{1.035\textwidth}{!}{
\hspace*{-0.5cm}\begin{tabular}{c|c|ccc|ccc|ccc|ccc}
\hline
\multirow{2}{*}{\textbf{Missing Scheme}} & \multirow{2}{*}{\textbf{Missing Rate}} & \multicolumn{3}{c|}{\textbf{PeMSD3}} & \multicolumn{3}{c|}{\textbf{PeMSD4}} & \multicolumn{3}{c|}{\textbf{PeMSD7}} & \multicolumn{3}{c}{\textbf{METR-LA}} \\ \cline{3-14} 
 &  & \textbf{RMSE} & \textbf{MAE} & \textbf{MAPE} & \textbf{RMSE} & \textbf{MAE} & \textbf{MAPE} & \textbf{RMSE} & \textbf{MAE} & \textbf{MAPE} & \textbf{RMSE} & \textbf{MAE} & \textbf{MAPE} \\ \hline
\textbf{EIKF-Net} & \textbf{0\%} & \textbf{23.107} & \textbf{14.837} & \textbf{13.070} & \textbf{28.879} & \textbf{19.206} & \textbf{11.856} & \textbf{31.847} & \textbf{21.321} & \textbf{8.959} & \textbf{8.764} & \textbf{4.867} & 9.987 \\ \hline
\multirow{3}{*}{Point} & 10\% & 23.112 & 14.945 & 13.768 & 31.026 & 20.955 & 12.808 & 33.852 & 22.954 & 9.687 & 9.154 & 5.286 & \textbf{9.618} \\
 & 30\% & 23.317 & 15.328 & 13.941 & 33.101 & 22.511 & 13.707 & 35.377 & 25.054 & 10.325 & 9.633 & 5.769 & 10.259 \\
 & 50\% & 24.972 & 16.697 & 14.568 & 34.628 & 23.613 & 14.468 & 37.486 & 25.681 & 11.027 & 10.090 & 6.188 & 10.746 \\ \hline
\multirow{3}{*}{\begin{tabular}[c]{@{}c@{}}Point\\ (Only Spatial)\end{tabular}} & 10\% & 23.104 & 15.332 & 13.350 & 31.575 & 21.472 & 13.366 & 34.315 & 23.251 & 9.920 & 9.259 & 5.389 & 9.760 \\
 & 30\% & 23.687 & 16.030 & 14.004 & 33.330 & 22.793 & 14.005 & 36.167 & 24.678 & 10.586 & 9.740 & 5.910 & 10.472 \\
 & 50\% & 25.058 & 17.012 & 14.860 & 35.053 & 23.969 & 14.687 & 38.082 & 26.156 & 11.261 & 10.197 & 6.368 & 11.005 \\ \hline
\multirow{3}{*}{\begin{tabular}[c]{@{}c@{}}Point\\ (Only Temporal)\end{tabular}} & 10\% & 35.172 & 23.898 & 19.144 & 42.701 & 29.756 & 17.862 & 49.022 & 34.684 & 15.084 & 9.993 & 5.869 & 9.968 \\
 & 30\% & 41.843 & 29.899 & 22.898 & 50.987 & 37.087 & 21.516 & 58.650 & 43.920 & 18.836 & 10.841 & 6.872 & 10.763 \\
 & 50\% & 47.987 & 34.661 & 24.885 & 58.017 & 42.876 & 23.462 & 66.751 & 50.904 & 20.852 & 11.624 & 7.676 & 11.434 \\ \hline
\multirow{3}{*}{Block} & 10\% & 22.470 & 14.969 & 12.936 & 31.004 & 20.990 & 12.842 & 33.806 & 22.876 & 9.688 & 9.150 & 5.284 & 9.627 \\
 & 30\% & 23.524 & 15.864 & 13.702 & 33.379 & 22.677 & 13.891 & 35.371 & 24.057 & 10.290 & 9.739 & 5.920 & 10.477 \\
 & 50\% & 25.060 & 17.014 & 14.731 & 35.234 & 24.067 & 14.751 & 37.093 & 25.332 & 10.809 & 10.208 & 6.392 & 11.003 \\ \hline 
\multirow{3}{*}{\begin{tabular}[c]{@{}c@{}}Block\\ (Only Spatial)\end{tabular}} & 10\% & 23.082 & 15.343 & 13.344 & 31.583 & 21.478 & 13.406 & 34.214 & 23.177 & 9.880 & 9.263 & 5.399 & 9.774 \\
 & 30\% & 23.666 & 16.002 & 13.908 & 35.092 & 24.123 & 14.859 & 36.021 & 24.560 & 10.541 & 9.741 & 5.923 & 10.479 \\
 & 50\% & 25.157 & 17.032 & 14.733 & 36.064 & 24.696 & 15.162 & 37.808 & 25.903 & 11.139 & 10.207 & 6.392 & 11.003 \\ \hline
 \multirow{3}{*}{\begin{tabular}[c]{@{}c@{}}Block\\ (Only Temporal)\end{tabular}} & 10\% & 34.909 & 23.616 & 18.862 & 42.444 & 29.493 & 17.645 & 48.421 & 33.975 & 14.626 & 9.982 & 5.855 & 9.961 \\
 & 30\% & 42.345 & 30.206 & 23.037 & 50.954 & 37.089 & 21.521 & 59.060 & 44.193 & 18.920 & 10.819 & 6.863 & 10.752 \\
 & 50\% & 50.357 & 36.535 & 25.491 & 58.695 & 43.416 & 23.579 & 70.617 & 54.398 & 21.700 & 11.657 & 7.753 & 11.484 \\ \hline
\end{tabular}
}
\vspace{-2mm}
\caption{Pointwise forecasting error on irregular PeMSD3, PeMSD4, PeMSD7 and METR-LA}
\label{tab:pfe1}
\end{table}

\vspace{-3mm}
We split multiple benchmark datasets in chronological order with a ratio of 7:1:2 for the METR-LA and PEMS-BAY datasets and a ratio of 6:2:2 for the other datasets into training, validation, and test sets, respectively. We evaluate the model performance w.r.t the forecasting error metrics to assess the model performance on the simulated data. The model performance helps to understand its ability to handle MTS data with missingness or irregular intervals and how the model performance is affected as the percentage of missing data increases. We choose \textbf{EIKF-Net} framework trained on fully observed data(i.e., $0\%$ missingness) as the SOTA benchmark for the MTSF task. Tables \ref{tab:pfe1} and \ref{tab:pfe2} report the irregular-time-series forecasting results on the benchmark datasets. The model performance slightly degrades compared to the benchmark model for a lower percentage of missing data. Nevertheless, as the percentage of incomplete data increases, the model performance degrades resulting in a lower forecast accuracy across the datasets, regardless of the specific pattern of the missing data. The proposed framework is relatively robust to missing data by conditioning the pointwise forecasts on the available observations rather than relying on the imputed values of the missing data to capture the complex dependencies and patterns underlying the MTS data, which can help to improve the forecast accuracy. In addition, it generates more reliable out-of-sample forecasts by effectively capturing the nonlinear spatial-temporal dynamic dependencies existing within the networks of interconnected sensors. Furthermore, based on the results, in different missing data scenarios, the ablation model only with the spatial inference component significantly outperforms that only with the temporal inference component. However, \textbf{EIKF-Net}, which utilizes both the temporal and the spatial learning components, still outperforms all the ablated models. It suggests that the joint-optimization of temporal and spatial inference components is necessary for achieving optimal performance in missing data scenarios. The experimental results suggest that our framework can jointly learn the spatial-temporal dependencies from the partial data in various missingness patterns resulting in less forecast error.

\vspace{-2mm}
\begin{table}[ht!]
\setlength{\tabcolsep}{0.45em} 
\renewcommand\arraystretch{1.275} 
\centering
 \resizebox{0.935\textwidth}{!}{
\begin{tabular}{c|c|ccc|ccc|ccc}
\hline
\multirow{2}{*}{\textbf{Missing Scheme}} & \multirow{2}{*}{\textbf{Missing Rate}} & \multicolumn{3}{c|}{\textbf{PeMSD7(M)}} & \multicolumn{3}{c|}{\textbf{PeMSD8}} & \multicolumn{3}{c}{\textbf{PEMS-BAY}} \\ \cline{3-11} 
 &  & \textbf{MAE} & \textbf{RMSE} & \textbf{MAPE} & \textbf{MAE} & \textbf{RMSE} & \textbf{MAPE} & \textbf{MAE} & \textbf{RMSE} & \textbf{MAPE} \\ \hline
\textbf{EIKF-Net} & \textbf{0\%} & \textbf{5.249} & \textbf{2.994} & \textbf{6.349} & \textbf{23.337} & \textbf{15.355} & \textbf{8.721} & \textbf{3.119} & \textbf{1.699} & \textbf{3.154} \\ \hline
\multirow{3}{*}{Point} & 10\% & 5.742 & 3.474 & 7.334 & 25.598 & 17.352 & 9.808 & 3.316 & 1.870 & 3.447 \\
 & 30\% & 6.115 & 3.838 & 8.134 & 26.860 & 18.311 & 10.416 & 3.458 & 2.001 & 3.680 \\
 & 50\% & 6.450 & 4.151 & 8.736 & 28.908 & 19.911 & 11.352 & 3.624 & 2.130 & 3.898 \\ \hline
\multirow{3}{*}{\begin{tabular}[c]{@{}c@{}}Point\\ (Spatial Only)\end{tabular}} & 10\% & 5.775 & 3.528 & 7.436 & 25.928 & 17.642 & 10.081 & 3.325 & 1.894 & 3.489 \\
 & 30\% & 6.199 & 3.956 & 8.365 & 27.715 & 19.117 & 10.866 & 3.459 & 2.003 & 3.685 \\
 & 50\% & 6.576 & 4.317 & 9.027 & 29.421 & 20.349 & 11.690 & 3.615 & 2.128 & 3.896 \\ \hline 
\multirow{3}{*}{\begin{tabular}[c]{@{}c@{}}Point\\ (Temporal Only)\end{tabular}} & 10\% & 6.415 & 3.771 & 7.869 & 35.769 & 24.917 & 14.379 & 3.793 & 2.081 & 3.735 \\
 & 30\% & 7.041 & 4.381 & 8.996 & 42.985 & 31.323 & 17.771 & 4.096 & 2.336 & 4.135 \\
 & 50\% & 7.610 & 4.879 & 9.809 & 49.357 & 36.442 & 19.760 & 4.376 & 2.548 & 4.431 \\ \hline  
\multirow{3}{*}{Block} & 10\% & 5.769 & 3.500 & 7.368 & 25.585 & 17.325 & 9.750 & 3.311 & 1.866 & 3.442 \\
 & 30\% & 6.142 & 3.872 & 8.187 & 26.842 & 18.309 & 10.468 & 3.455 & 1.988 & 3.660 \\
 & 50\% & 6.465 & 4.176 & 8.754 & 29.020 & 19.962 & 11.421 & 3.621 & 2.124 & 3.883 \\ \hline
\multirow{3}{*}{\begin{tabular}[c]{@{}c@{}}Block\\ (Spatial Only)\end{tabular}} & 10\% & 5.759 & 3.514 & 7.428 & 26.016 & 17.710 & 10.137 & 3.327 & 1.897 & 3.498 \\
 & 30\% & 6.160 & 3.902 & 8.269 & 27.526 & 18.891 & 10.729 & 3.465 & 2.010 & 3.701 \\
 & 50\% & 6.523 & 4.256 & 8.933 & 29.124 & 20.106 & 11.524 & 3.630 & 2.147 & 3.926 \\ \hline 
\multirow{3}{*}{\begin{tabular}[c]{@{}c@{}}Block\\ (Temporal Only)\end{tabular}} & 10\% & 6.417 & 3.758 & 7.824 & 35.501 & 24.646 & 14.186 & 3.781 & 2.068 & 3.711 \\
 & 30\% & 7.064 & 4.390 & 9.000 & 42.856 & 31.237 & 17.738 & 4.092 & 2.334 & 4.128 \\
 & 50\% & 7.668 & 4.909 & 9.859 & 49.808 & 36.785 & 19.797 & 4.447 & 2.613 & 4.526 \\ \hline
\end{tabular}
}\\[-1ex]
\caption{Pointwise forecasting error on irregular PeMSD7(M), PeMSD8 and PEMS-BAY}
\label{tab:pfe2}
\end{table}

\vspace{-7mm} 
\subsection{Sensitivity Analysis}
\label{Sanalysis}
\begin{figure}[ht]
\centering
\hspace*{-0.25cm}\resizebox{1.065\textwidth}{!}{
\subfloat[PeMSD3]{\includegraphics[width=60mm]{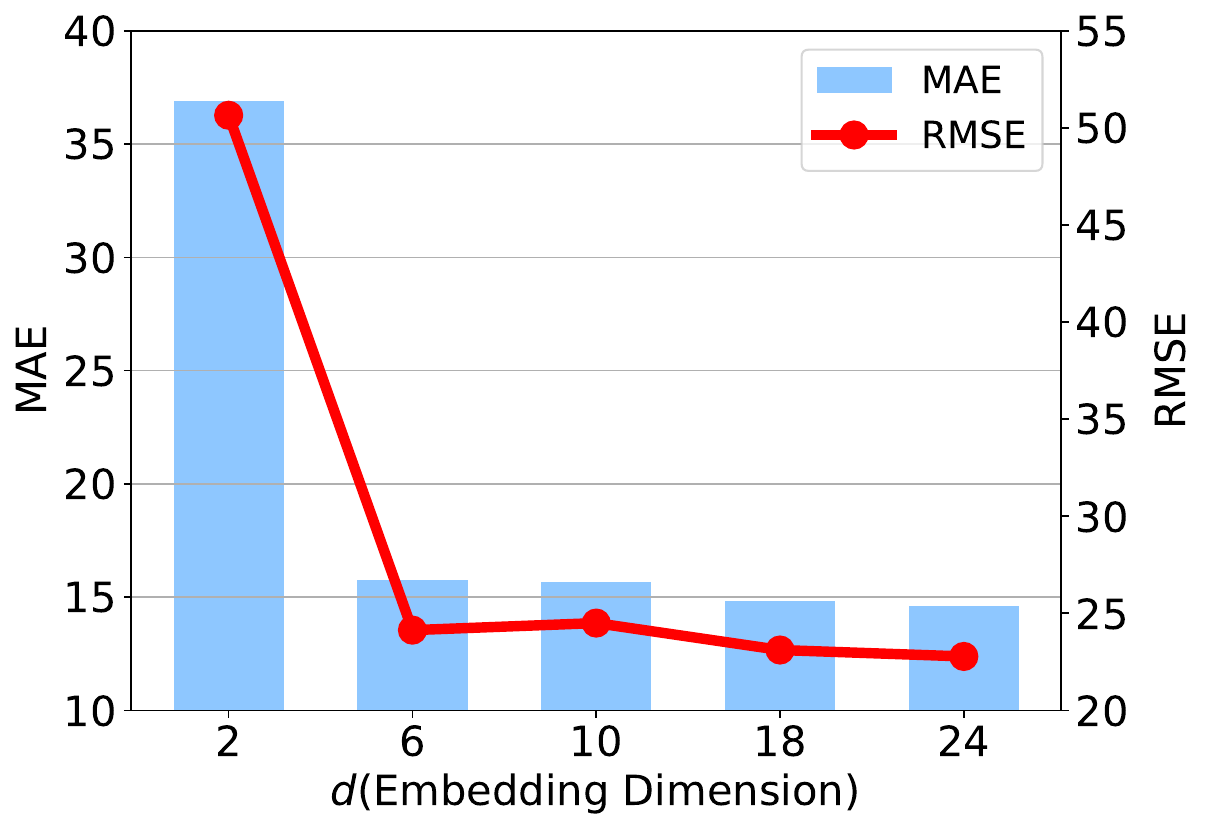}}
\subfloat[PeMSD4]{\includegraphics[width=60mm]{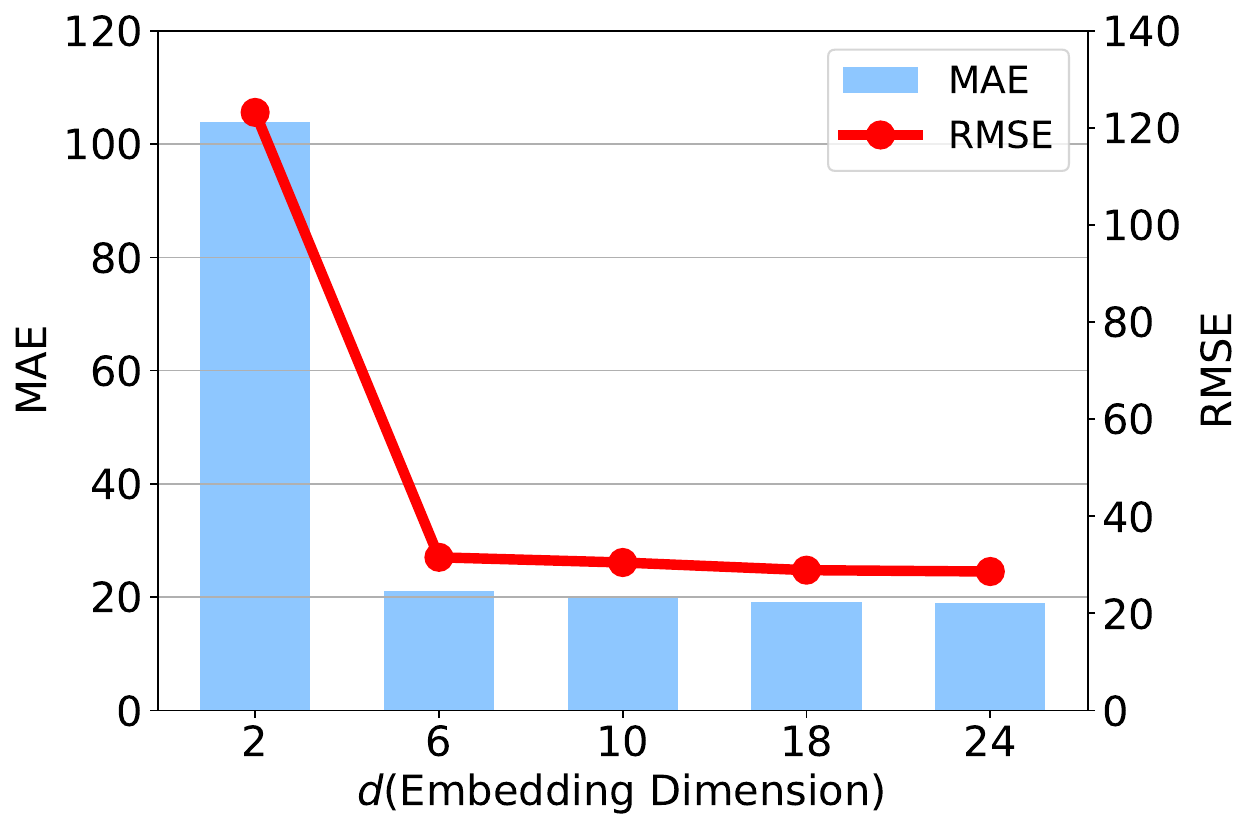}}
\subfloat[PeMSD7]{\includegraphics[width=60mm]{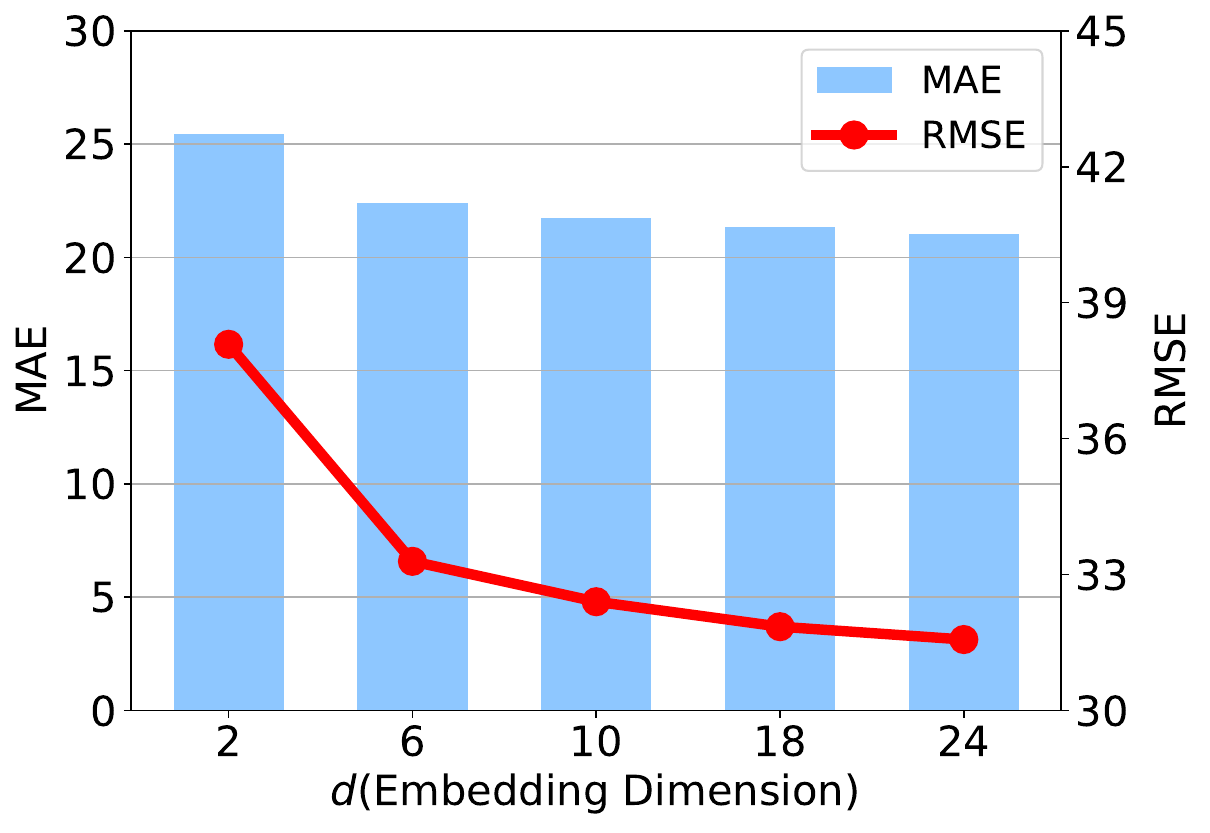}}
}\\[-2ex]
\hfill
\hspace*{-0.25cm}\resizebox{1.065\textwidth}{!}{
\subfloat[PeMSD7(M)]{\includegraphics[width=60mm]{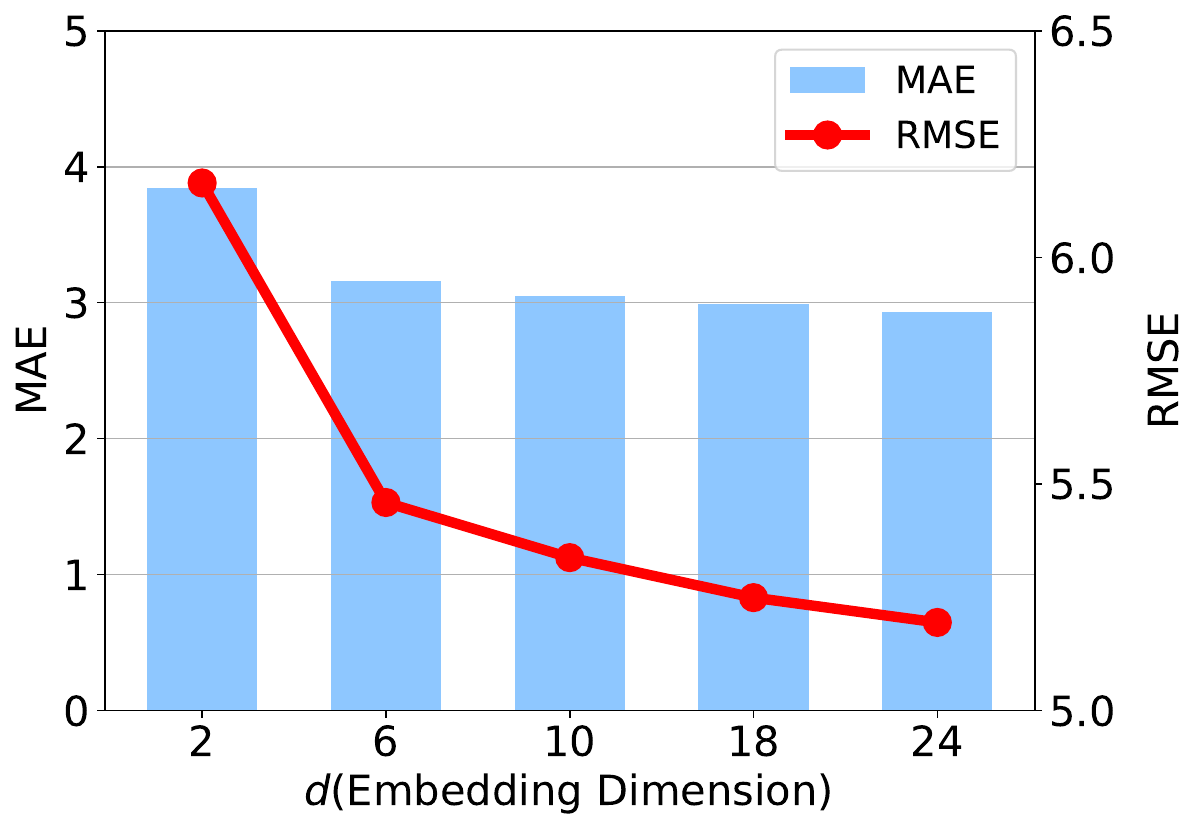}}
\subfloat[PeMSD8]{\includegraphics[width=60mm]{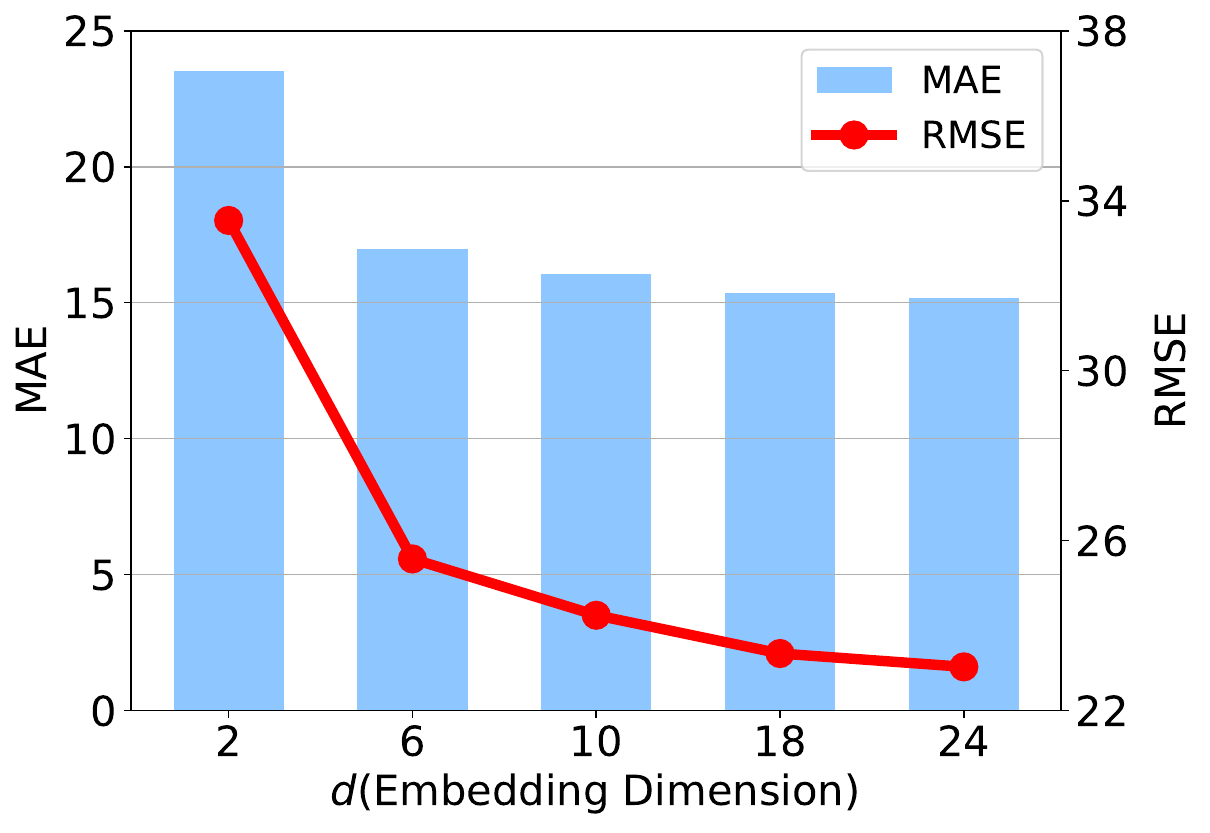}}
\subfloat[PeMSD3]{\includegraphics[width=60mm]{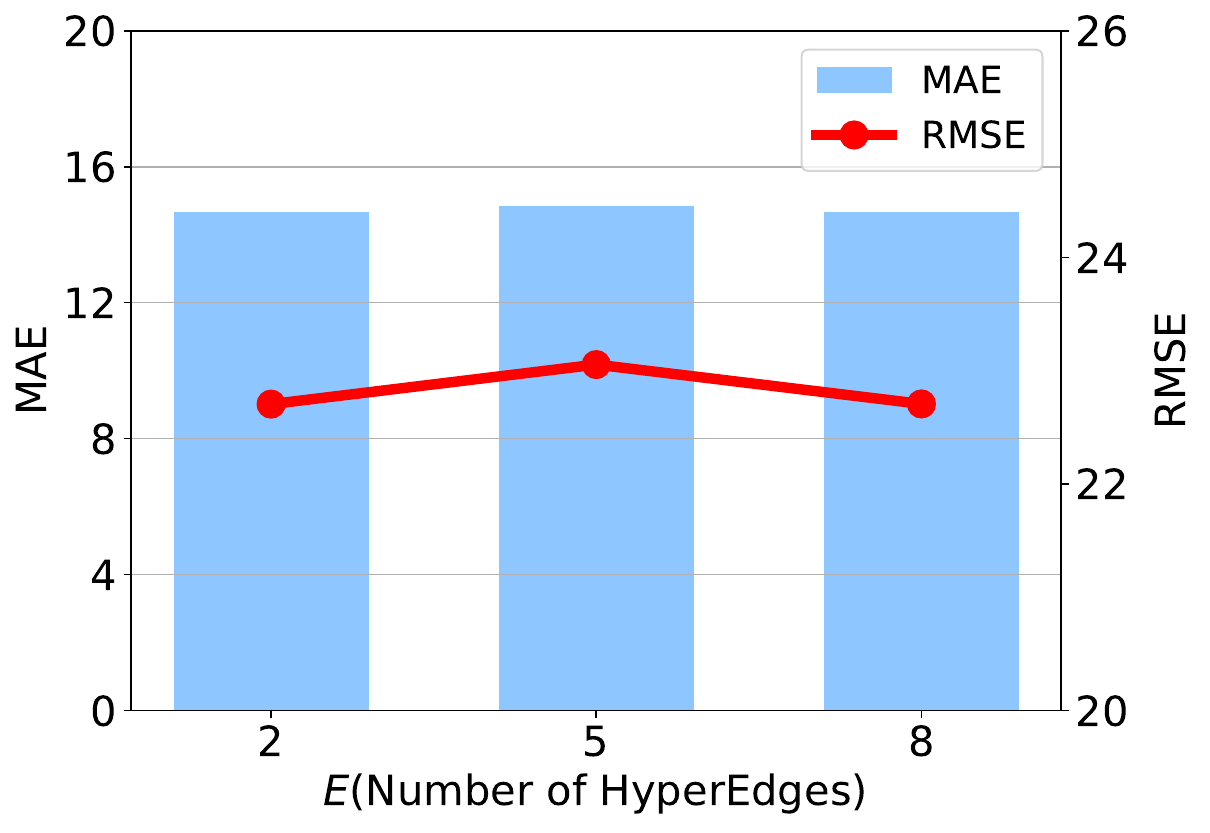}}
} \\[-2ex]
\hfill
\hspace*{-0.25cm}\resizebox{1.065\textwidth}{!}{
\subfloat[PeMSD4]{\includegraphics[width=60mm]{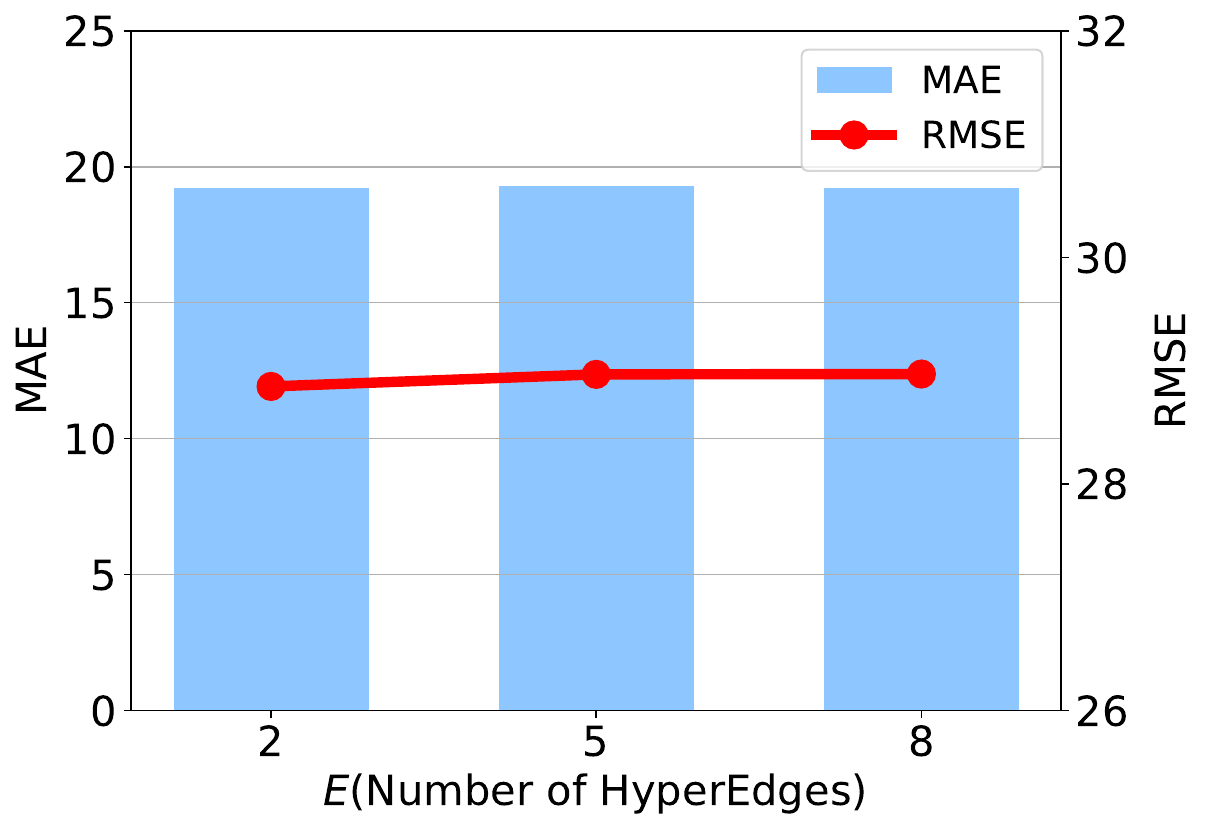}}
\subfloat[PeMSD7]{\includegraphics[width=60mm]{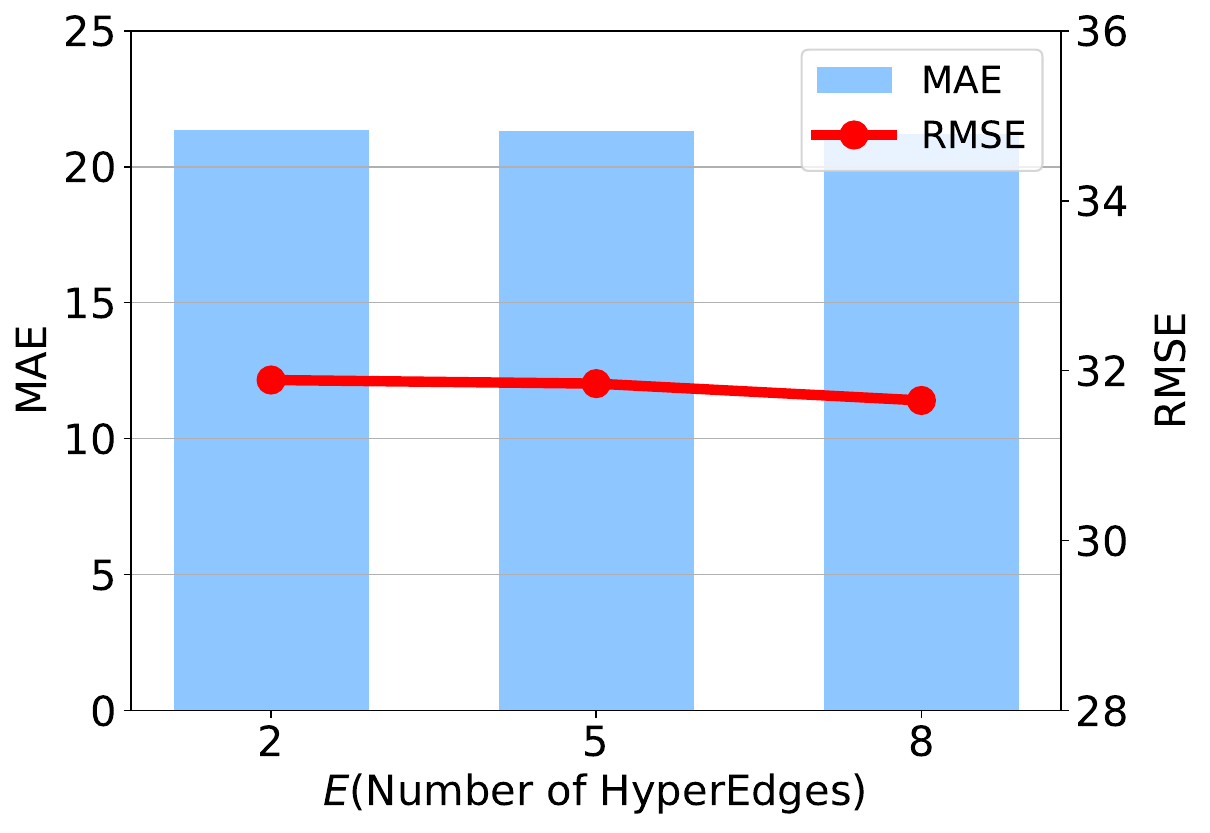}}
\subfloat[PeMSD7(M)]{\includegraphics[width=60mm]{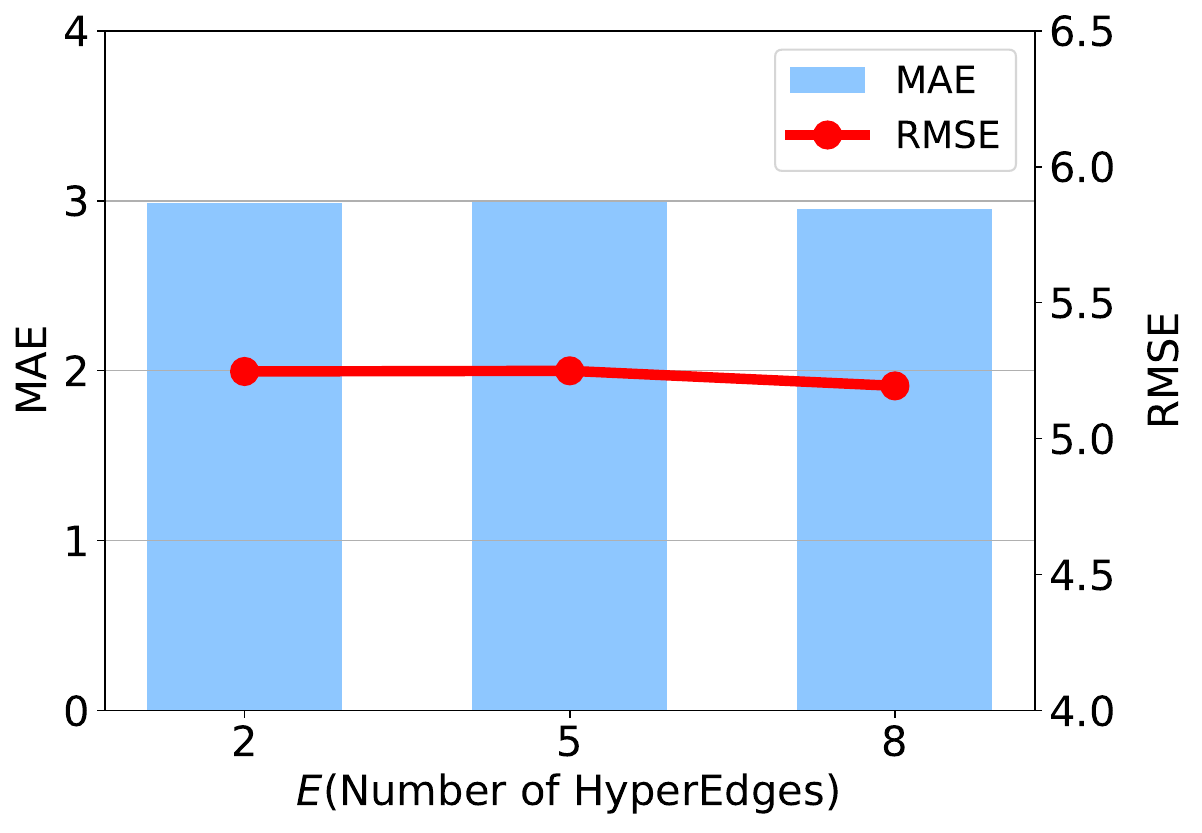}}
} \\[-2ex]
\hspace*{-0.25cm}\resizebox{0.725\textwidth}{!}{
\subfloat[PeMSD8]{\includegraphics[width=60mm]{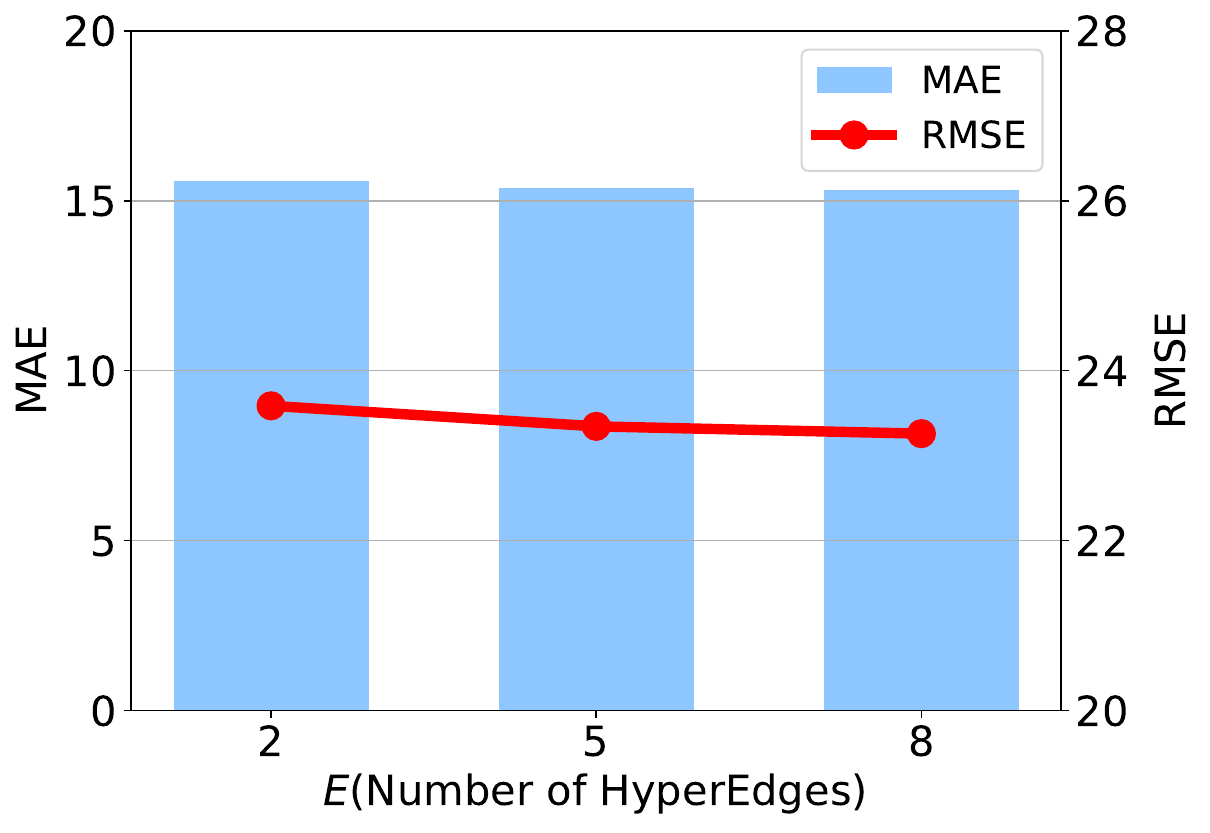}} 
\subfloat[Fraction of Incident Hypernodes]{\includegraphics[width=60mm]{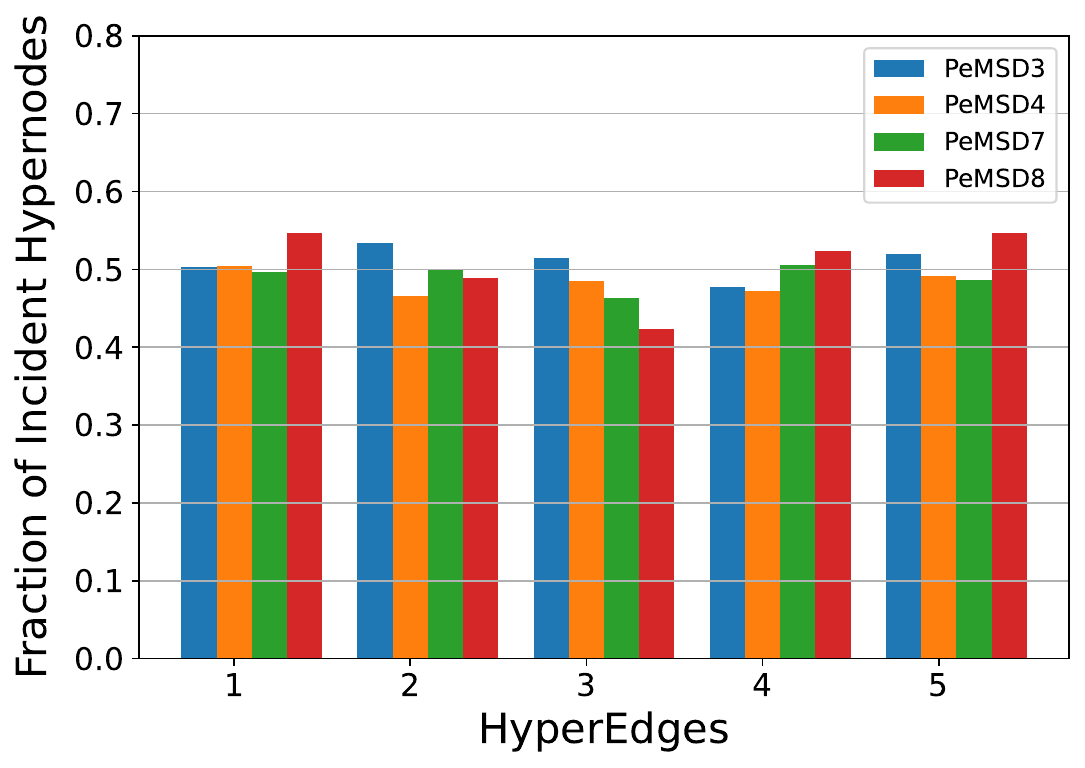}}
} \\[-0.5ex]
\caption{Sensitivity to embedding dimension and number of hyperedges}
\label{fig:sa}
\end{figure}

\vspace{-3mm}
We conduct a hyperparameter study to find the impact of hyperparameters on the performance of proposed framework. The goal is to find the set of hyperparameter values that lead to the optimal performance of the framework on the benchmark datasets, respectively. The algorithm hyperparameters are the embedding size($\textit{d}$), batch size($\textit{b}$), learning rate($\textit{lr}$), and the number of hyperedges($|\mathcal{HE}|$). We have tuned the following hyperparameters over specific ranges of values: embedding dimension(d) $\in \{2, 6, 10, 18, 24\}$, the number of hyperedges($|\mathcal{HE}|$) $\in \{2, 5, 8\}$, batch size($\textit{b}$) $ \in \{2, 6, 10, 18, 24, 32, 64\}$, and the learning rate($\textit{lr}$) $ \in \{\num{1e-1}, \num{1e-2}, \num{1e-3}, \num{1e-4}\}$. We have chosen the hyperparameter ranges to avoid Out-Of-Memory(OOM) errors on GPUs by limiting the model size and the memory requirements. We utilize the grid-search technique for hyperparameter optimization. Across all the datasets, we find the set of hyperparameter values that results in the lowest model forecast error on the validation set on a chosen evaluation metric as the optimal set of values. We utilize MAE and RMSE to evaluate the framework performance, with lower values indicating better performance. Figure \ref{fig:sa} shows how the \textbf{EIKF-Net} framework performance changes as the embedding size($\textit{d}$) and the number of hyperedges($|\mathcal{HE}|$) vary in a predefined range across all the datasets. It helps to understand the effect of these hyperparameters on the framework performance. We discuss the optimal set of hyperparameter configurations for each dataset as described below,

\vspace{-2mm}
\begin{itemize}
    \item For PeMSD3, we set the batch size($\textit{b}$) to 18, the initial learning rate($\textit{lr}$) to \num{1e-3}, and the embedding size($\textit{d}$) to 18. The number of hyperedges($|\mathcal{HE}|$) is 5.    
    \item For PeMSD4, we set the batch size($\textit{b}$) to 48, the initial learning rate($\textit{lr}$) to \num{1e-3}, and the embedding size($\textit{d}$) to 18. The number of hyperedges($|\mathcal{HE}|$) is 5.    
    \item For PeMSD7, we set the batch size($\textit{b}$) to 12, the initial learning rate($\textit{lr}$) to \num{1e-3}, and the embedding size($\textit{d}$) to 18. The number of hyperedges($|\mathcal{HE}|$) is 5.    
    \item For PeMSD8, we set the batch size($\textit{b}$) to 48, the initial learning rate($\textit{lr}$) to \num{1e-3}, and the embedding size($\textit{d}$) to 18. The number of hyperedges($|\mathcal{HE}|$) is 5.     
    \item For PeMSD7(M), we set the batch size($\textit{b}$) to 48, the initial learning rate($\textit{lr}$) to \num{1e-3}, and the embedding size($\textit{d}$) to 18. The number of hyperedges($|\mathcal{HE}|$) is 5. 
    \item For METR-LA, we set the batch size($\textit{b}$) to 48, the initial learning rate($\textit{lr}$) to \num{1e-3}, and the embedding size($\textit{d}$) to 18. The number of hyperedges($|\mathcal{HE}|$) is 5. 
    \item For PEMS-BAY, we set the batch size($\textit{b}$) to 12, the initial learning rate($\textit{lr}$) to \num{1e-3}, and the embedding size($\textit{d}$) to 18. The number of hyperedges($|\mathcal{HE}|$) is 5. 
\end{itemize}

\vspace{-6mm}

\vspace{-2mm}
The fraction of hypernodes connected with the hyperedges provides “edge density” information of a hypergraph. A higher fraction of hypernodes incident with hyperedges would suggest a denser network. In contrast, a lower fraction would indicate a sparser network. The density information helps to understand the structure of the hypergraph and can compare different hypergraphs. We control the density of the hypergraph by changing the number of hyperedges. We conducted an additional experiment for which we varied the number of predefined hyperedges to examine its impact on the learned hypergraph structures to potentially identify the optimal number of hyperedges for an MTSF task. This study helps to understand how the density of hypergraph changes as the number of hyperedges increases or decreases for a given dataset on MTSF task. Figures \ref{fig:sa}(f), \ref{fig:sa}(g), \ref{fig:sa}(h), \ref{fig:sa}(i), \ref{fig:sa}(j) shows the experimental results. Additionally, in another experiment, we have prefixed the number of hyperedges to 5. We study how the pattern of the fraction of incident hypernodes varies w.r.t to each hyperedge across the datasets for the MTSF task. Figure \ref{fig:sa}(k) shows the experimental results. We observe that the fraction of hypernodes connected to the hyperedges remains the same across the datasets. There is a need and necessity to dive deep to gain better insights for further understanding of the trends.

\vspace{-4mm}
\subsection{Time Series Forecasting Visualization}
\begin{figure}[ht]
\centering
\hspace*{-0.5cm}\resizebox{1.10\textwidth}{!}{
\subfloat[Node 12 in PeMSD3]{\includegraphics[width=60mm]{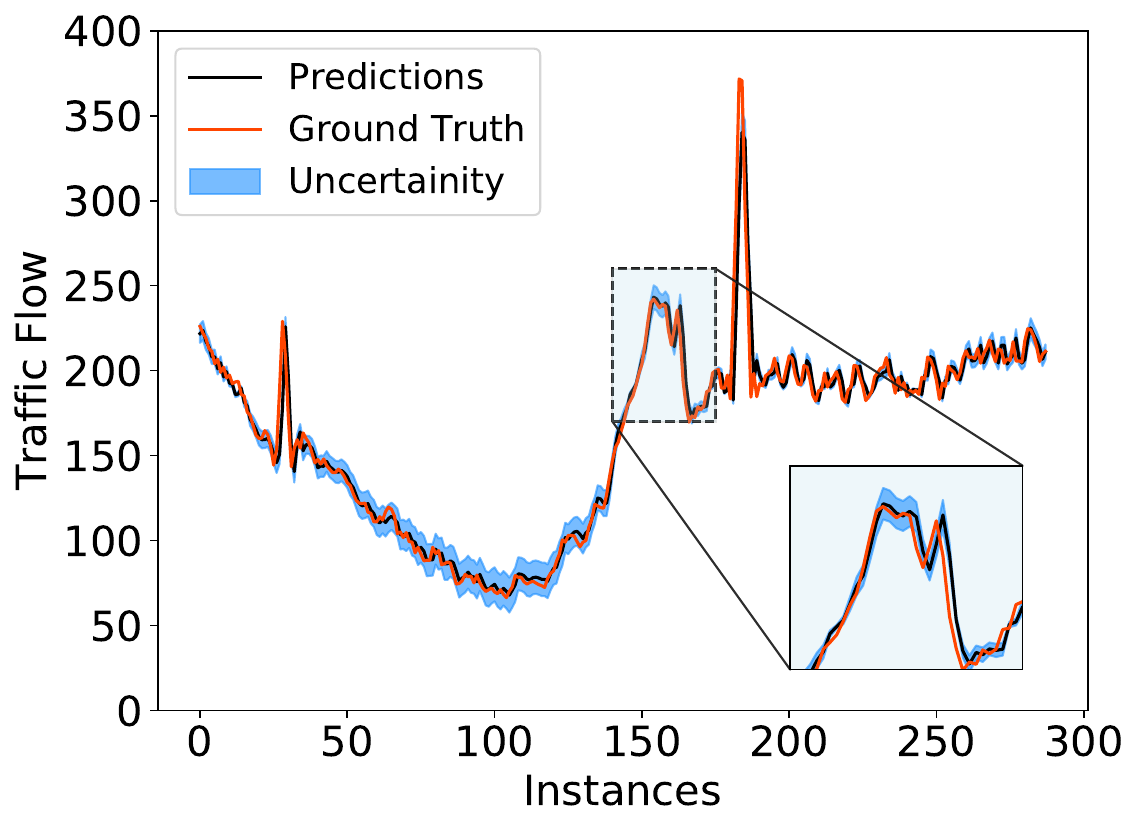}}
\subfloat[Node 99 in PeMSD3]{\includegraphics[width=60mm]{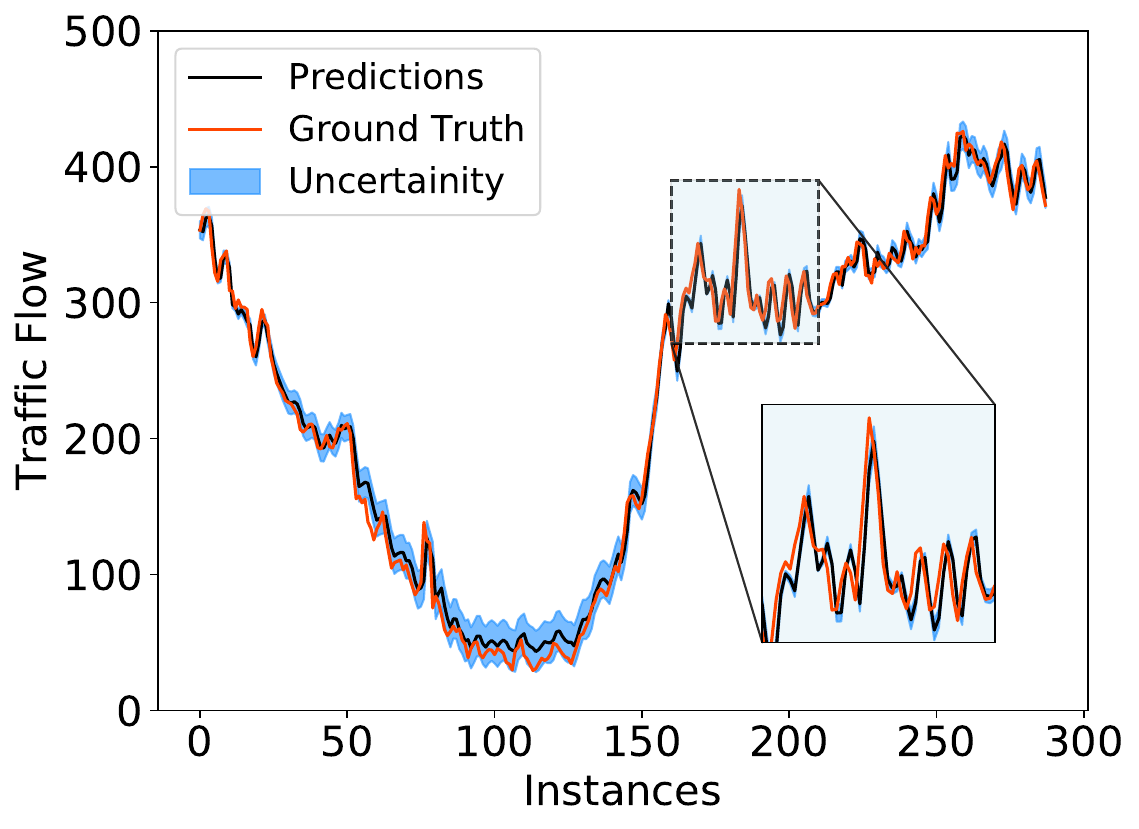}}
\subfloat[Node 108 in PeMSD3]{\includegraphics[width=60mm]{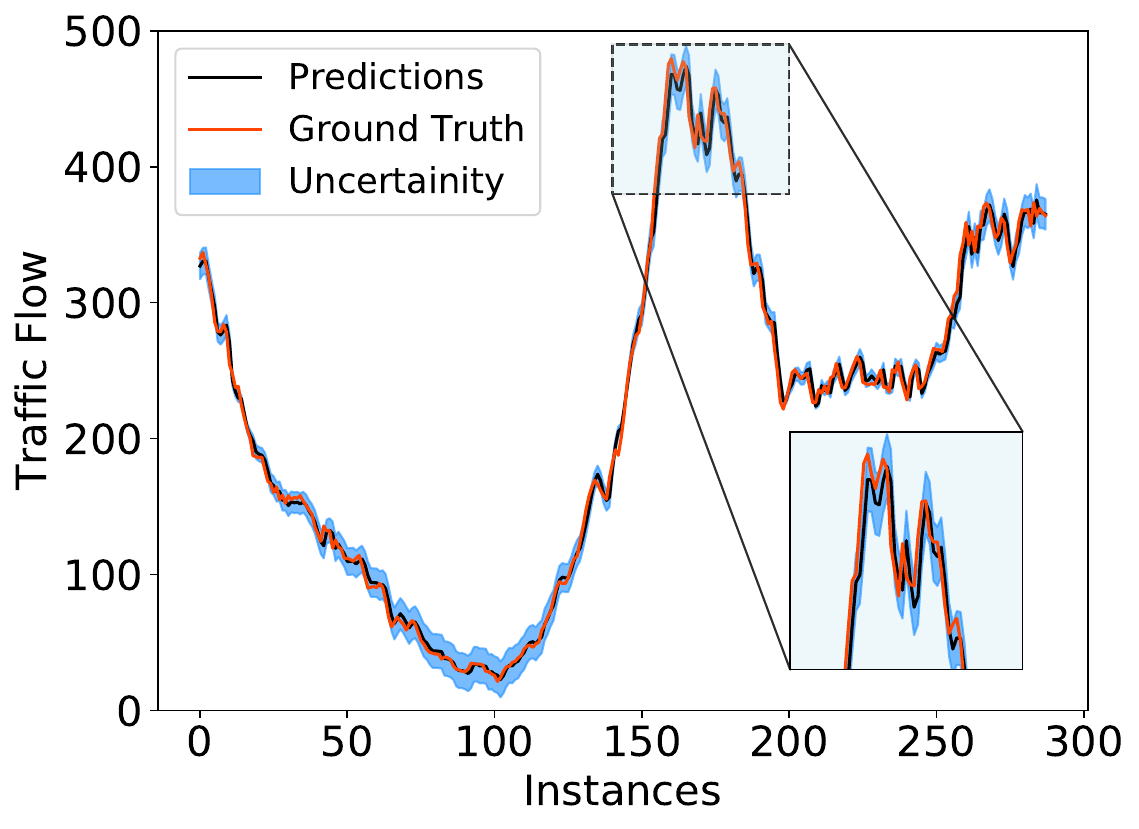}}
}\\[-2ex]
\hfill
\hspace*{-0.5cm}\resizebox{1.10\textwidth}{!}{
\subfloat[Node 141 in PeMSD3]{\includegraphics[width=60mm]{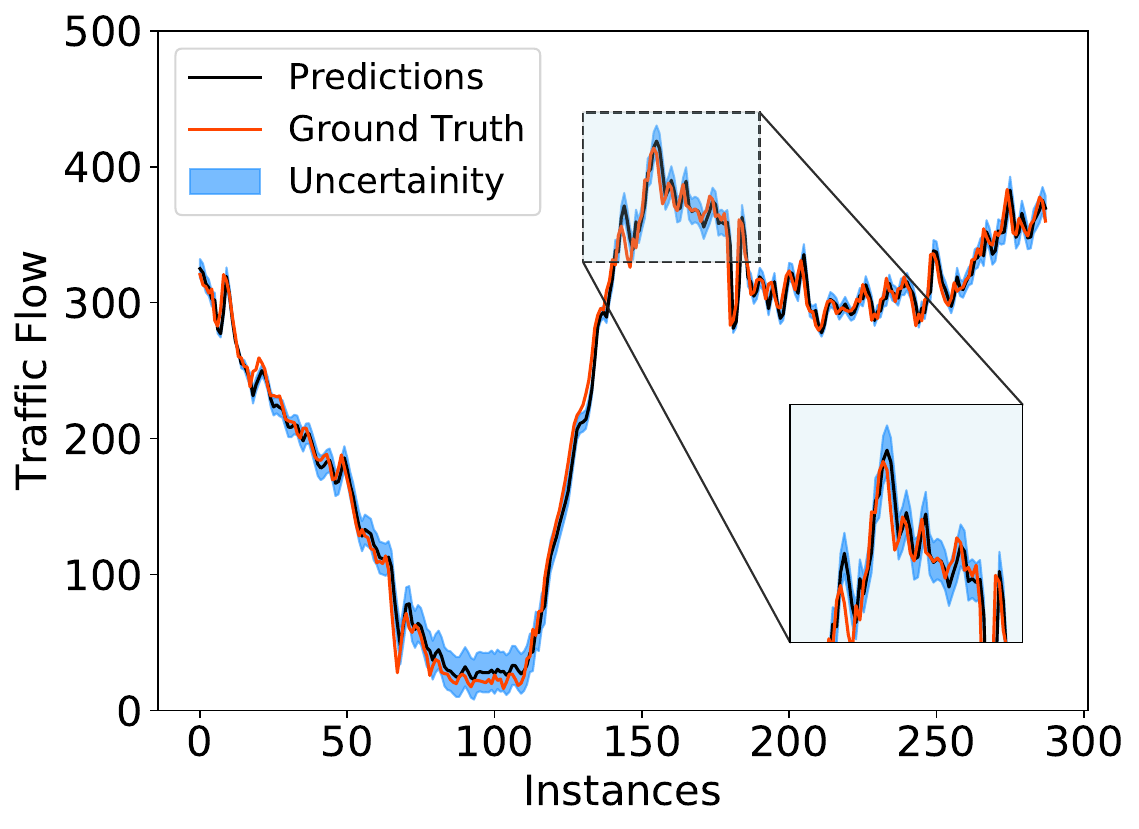}} 
\subfloat[Node 149 in PeMSD4]{\includegraphics[width=60mm]{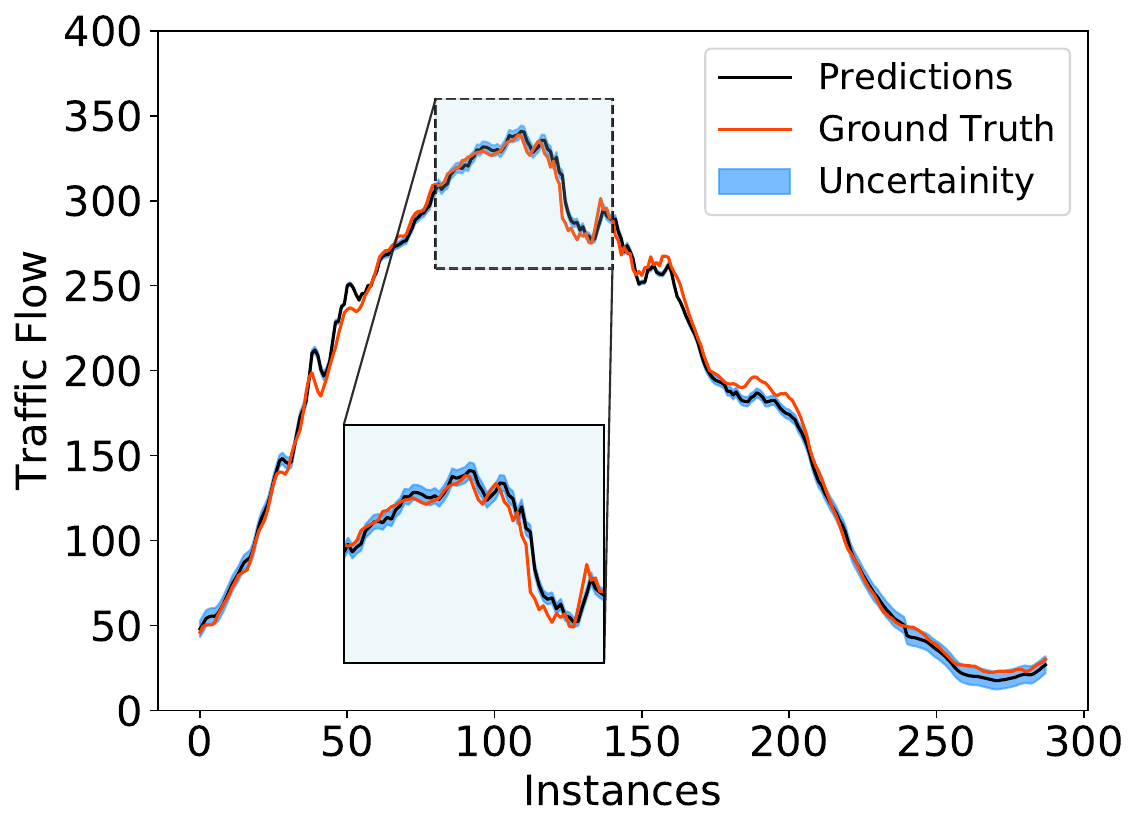}}
\subfloat[Node 170 in PeMSD4]{\includegraphics[width=60mm]{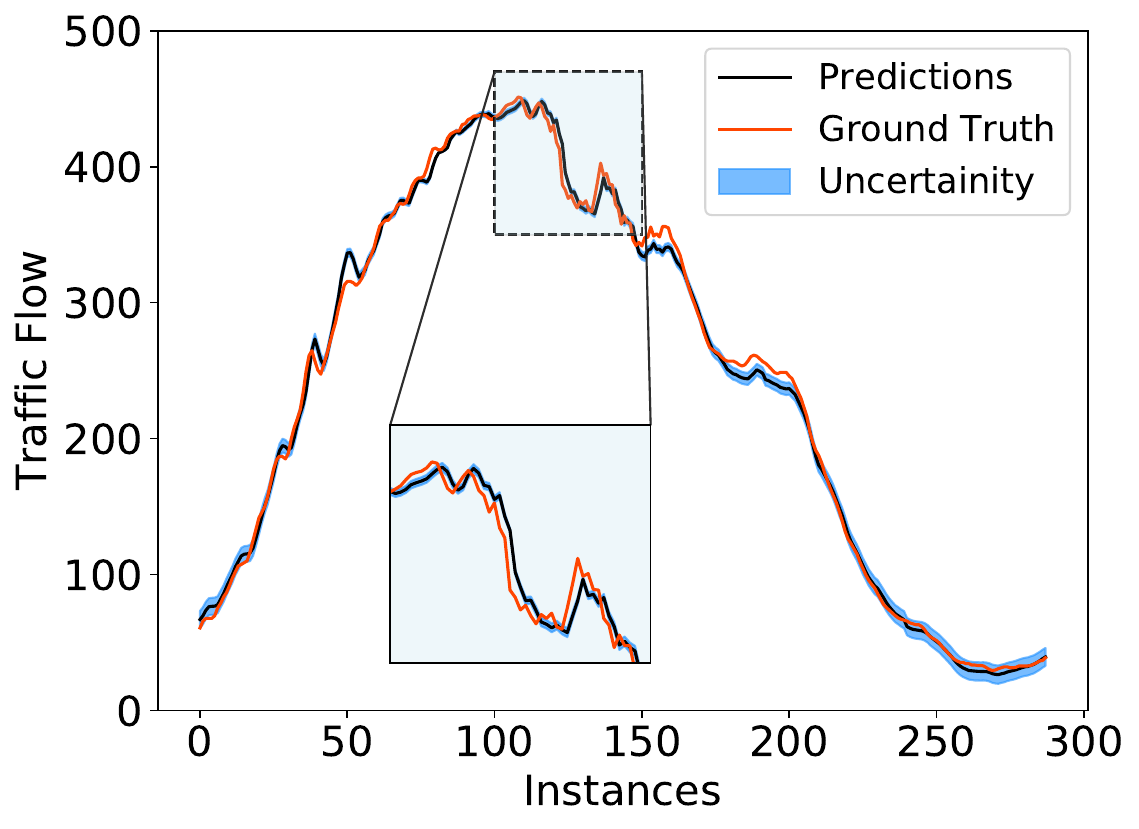}}
}\\[-1ex]
\hspace*{-0.5cm}\resizebox{1.10\textwidth}{!}{
\subfloat[Node 211 in PeMSD4]{\includegraphics[width=60mm]{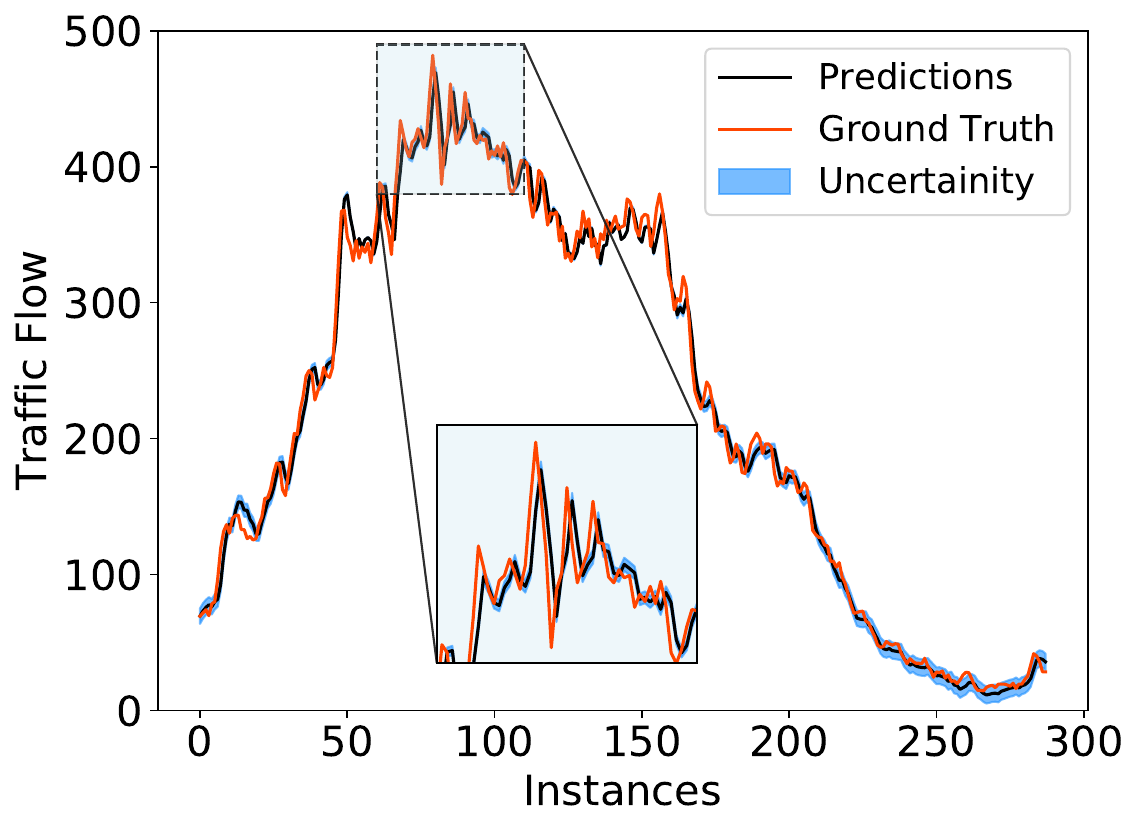}}
\subfloat[Node 287 in PeMSD4]{\includegraphics[width=60mm]{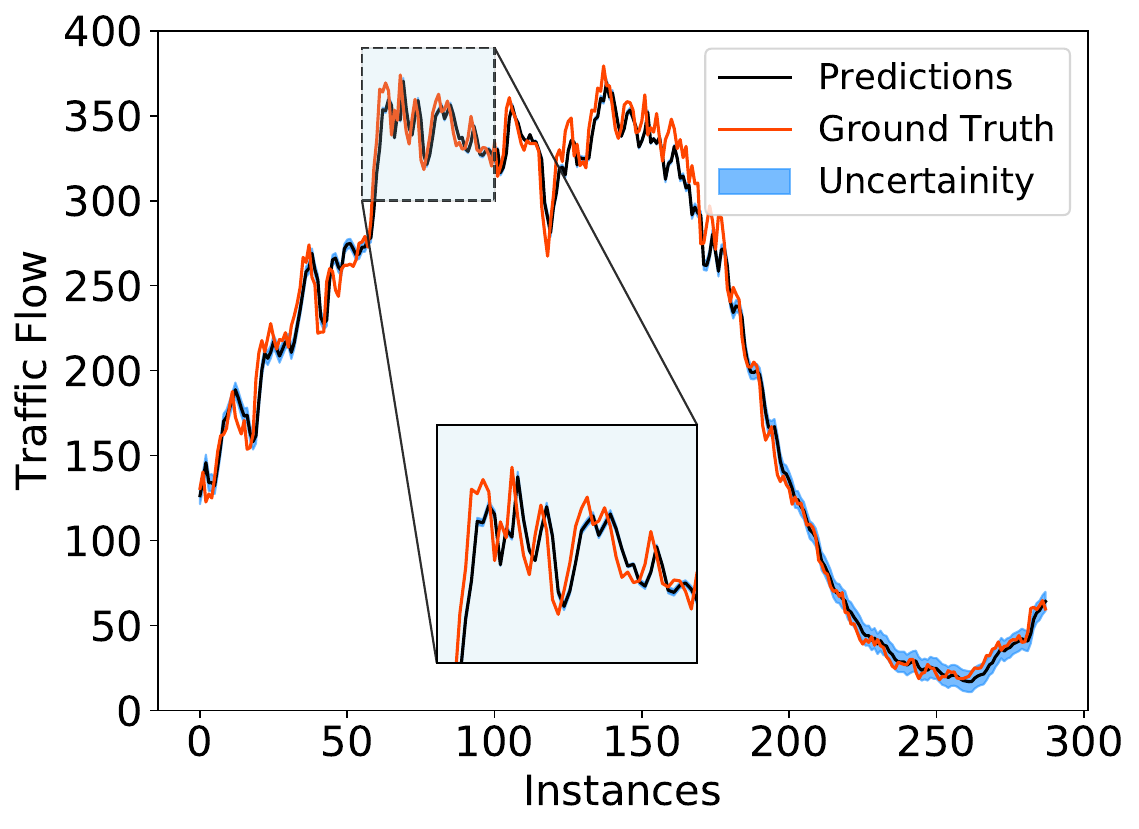}}
\subfloat[Node 85 in PeMSD8]{\includegraphics[width=60mm]{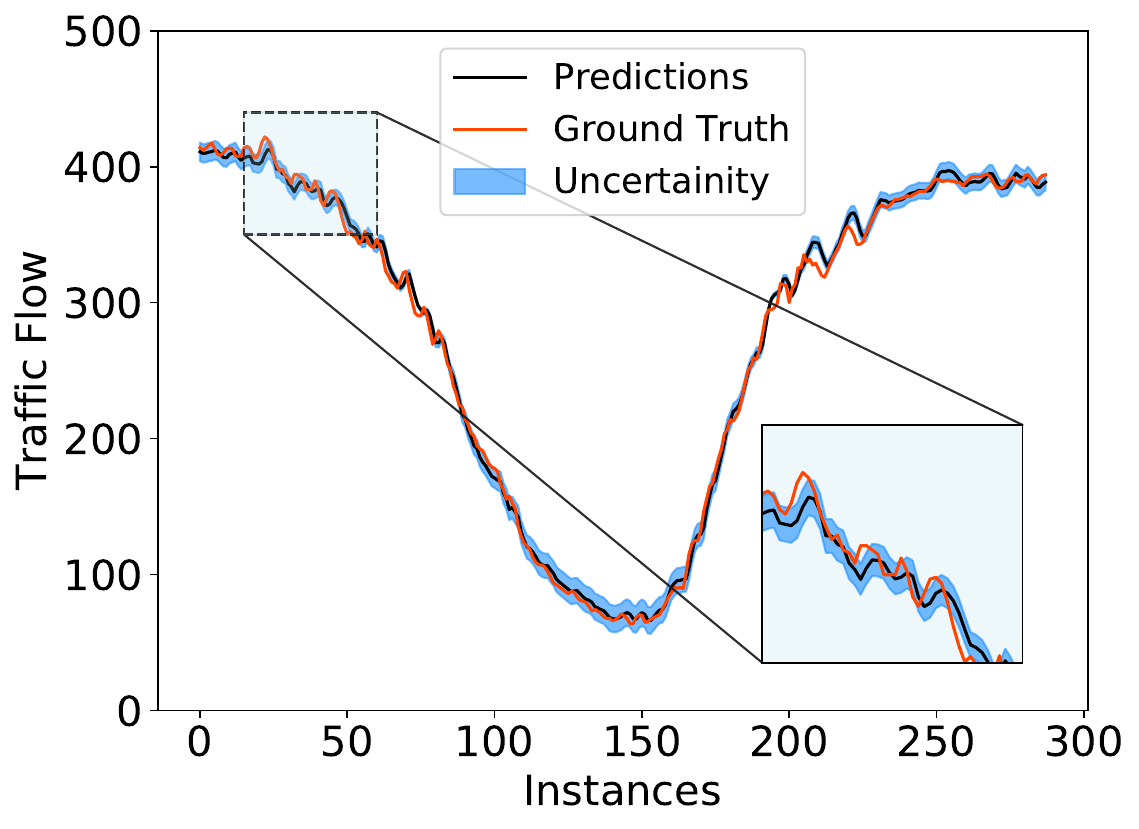}}
}\\[-1ex]
\hfill
\hspace*{-0.5cm}\resizebox{1.10\textwidth}{!}{
\subfloat[Node 104 in PeMSD8]{\includegraphics[width=60mm]{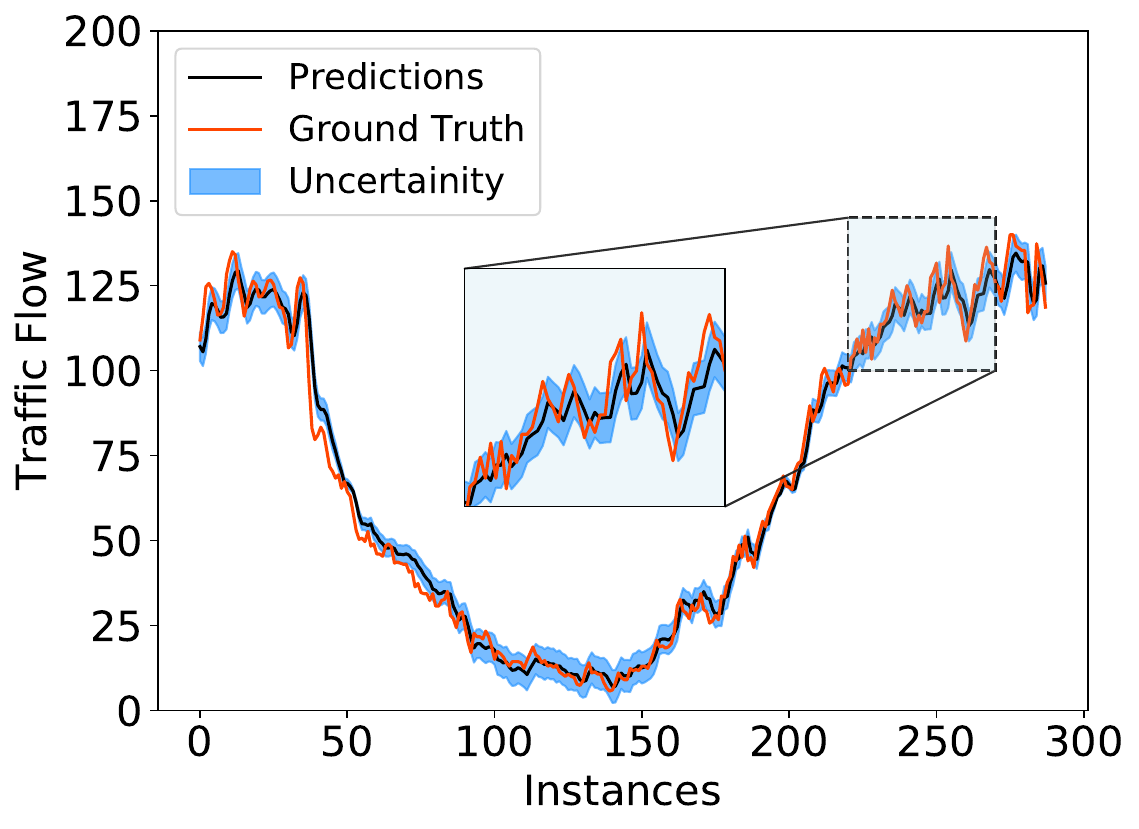}}
\subfloat[Node 155 in PeMSD8]{\includegraphics[width=60mm]{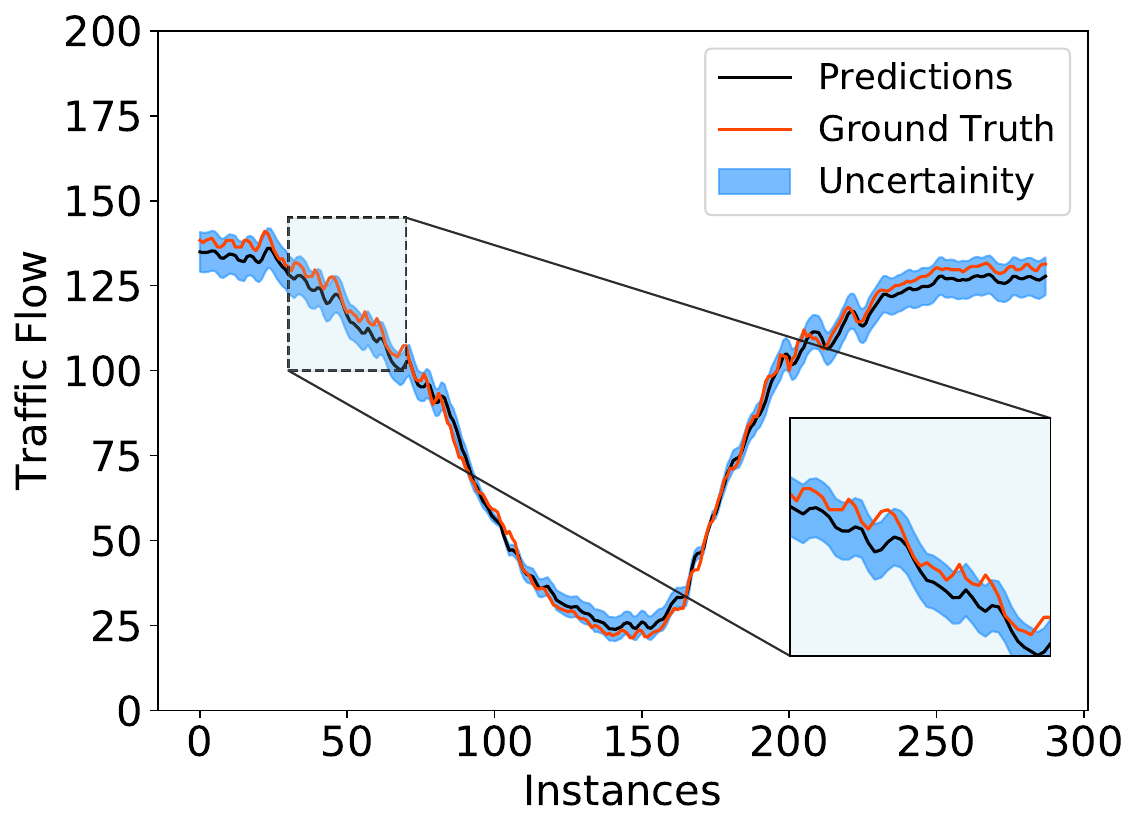}}
\subfloat[Node 162 in PeMSD8]{\includegraphics[width=60mm]{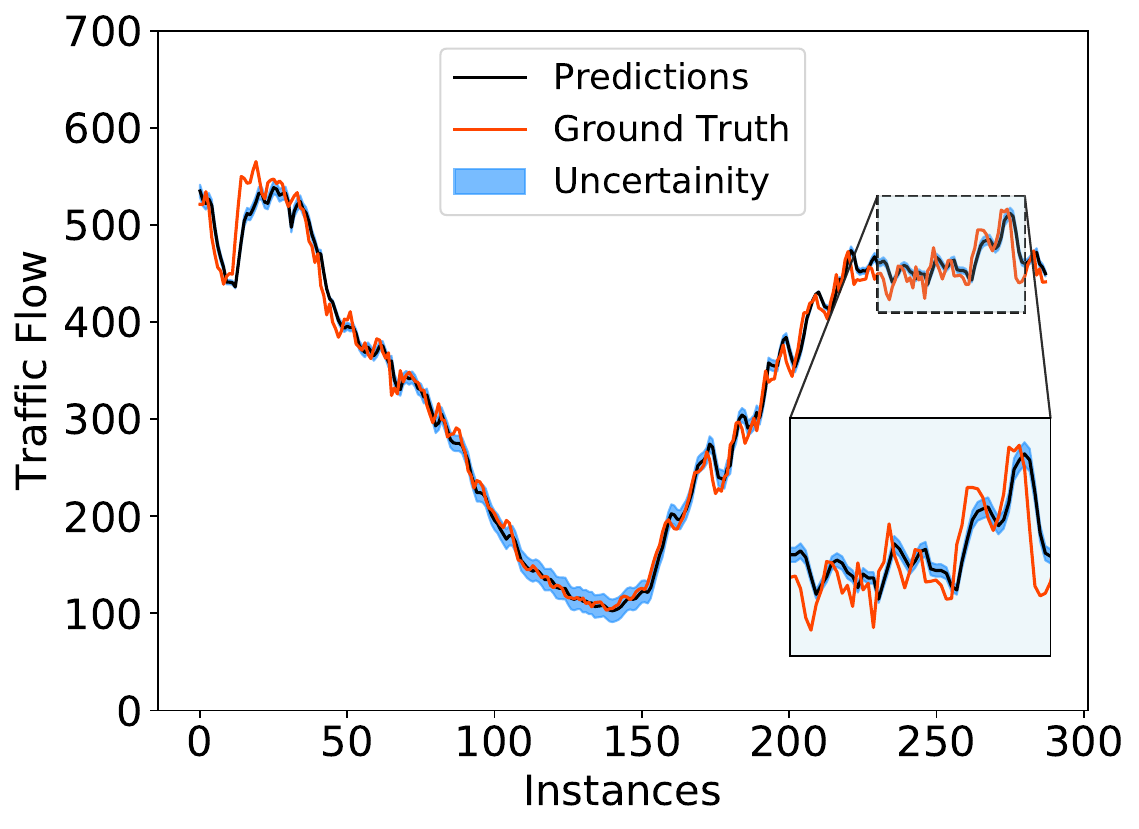}} 
}\\[-1.5ex]
\caption{Traffic forecasting visualization on PeMSD3, PeMSD4 and PeMSD8.}
\label{fig:tfv1}
\end{figure}

\begin{figure}[ht]
\center
\resizebox{0.40\textwidth}{!}{
\subfloat[Node 37 in PeMSD7(M)]{\includegraphics[width=60mm]{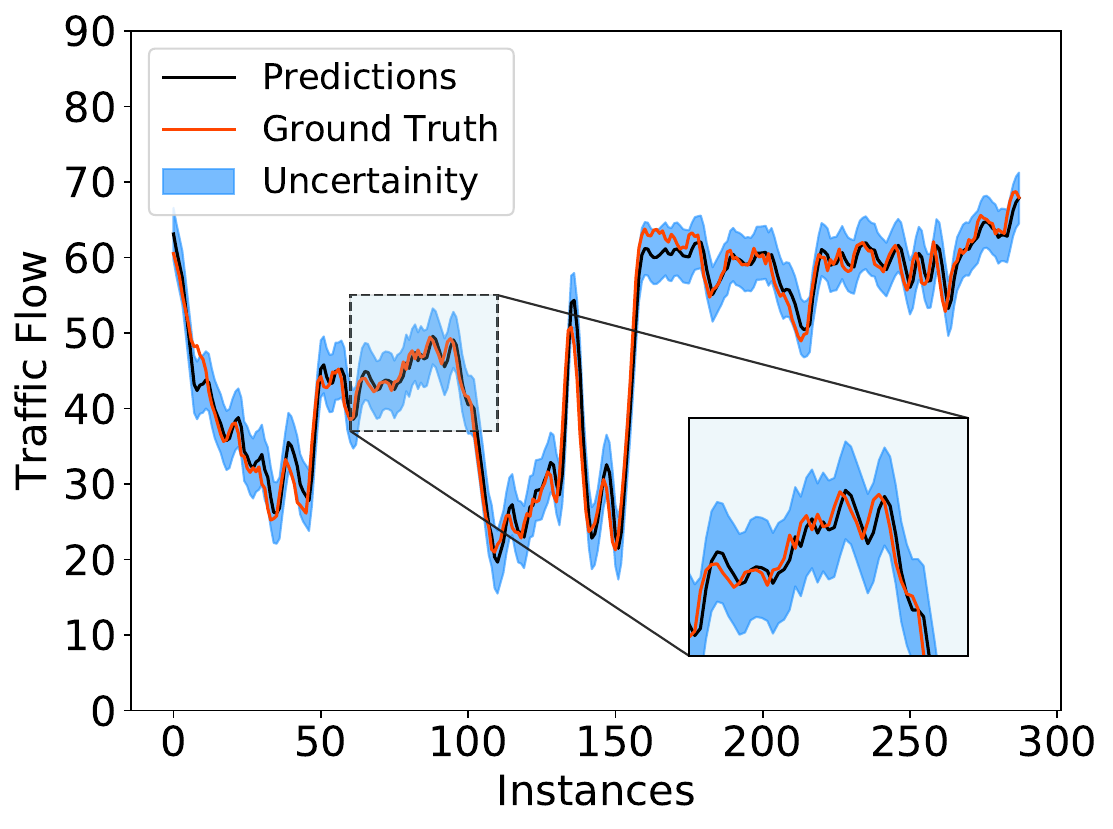}}
}
\caption{Traffic forecasting visualization on PEMSD7M.}
\label{fig:tfv2}
\end{figure}

\vspace{-3mm}
We visualize the ground truth, pointwise forecasts, and time-varying uncertainty estimates of our framework predictions shown in Figures \ref{fig:tfv1} and \ref{fig:tfv2}. 
However, existing methods for MTSF provide pointwise forecasts by modeling the nonlinear spatial-temporal dependencies existing within networks of interconnected sensors and fall short in measuring uncertainty.

\vspace{-6mm}

\vspace{-4mm}
In this context, our proposed \textbf{w/Unc- EIKF-Net} framework(\textbf{EIKF-Net} with local uncertainty estimation) effectively exploits the available – and often strong – relational information through the spatio-temporal propagation architecture to estimate uncertainty quantitatively. The multifaceted visualizations further show the efficacy of \textbf{EIKF-Net}, \textbf{w/Unc- EIKF-Net} in time series representation learning for the MTSF task.

\vspace{-4mm}
\begin{figure}[ht]
\centering
\hfill
\hspace*{-0.5cm}\resizebox{1.10\textwidth}{!}{
\subfloat[Node 14 in PeMSD7(M)]{\includegraphics[width=60mm]{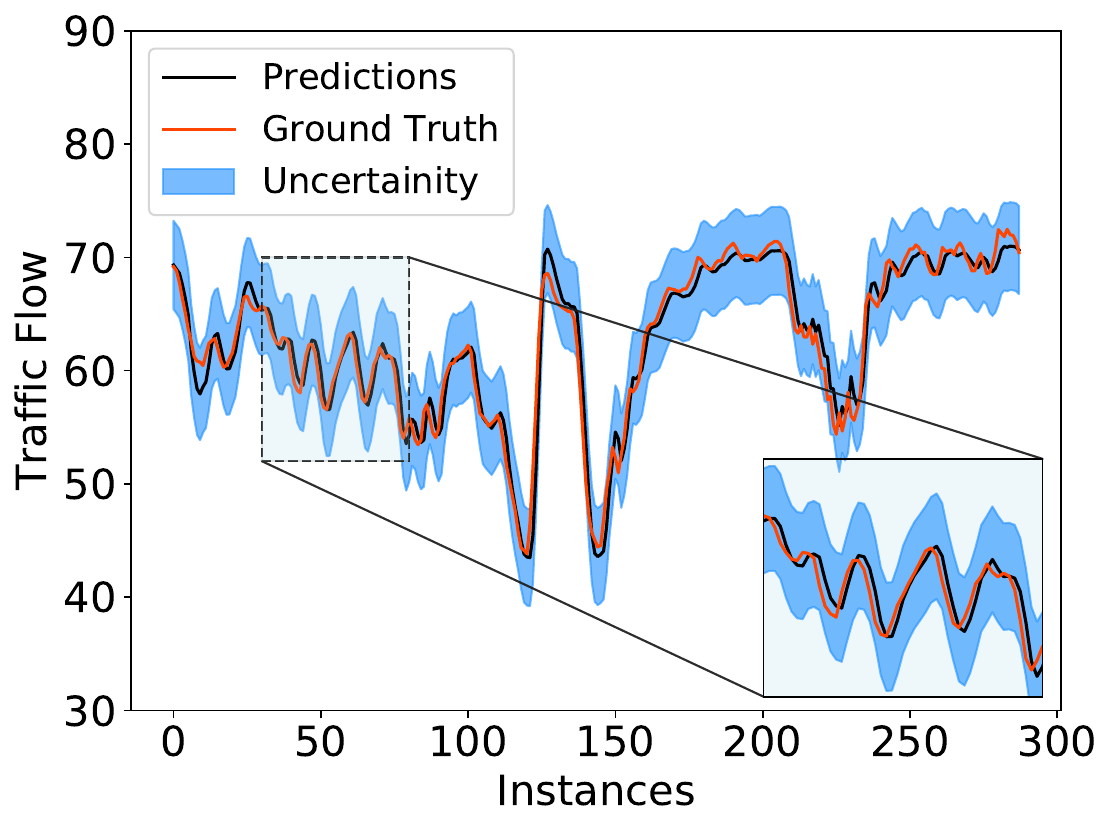}}
\subfloat[Node 15 in PeMSD7(M)]{\includegraphics[width=60mm]{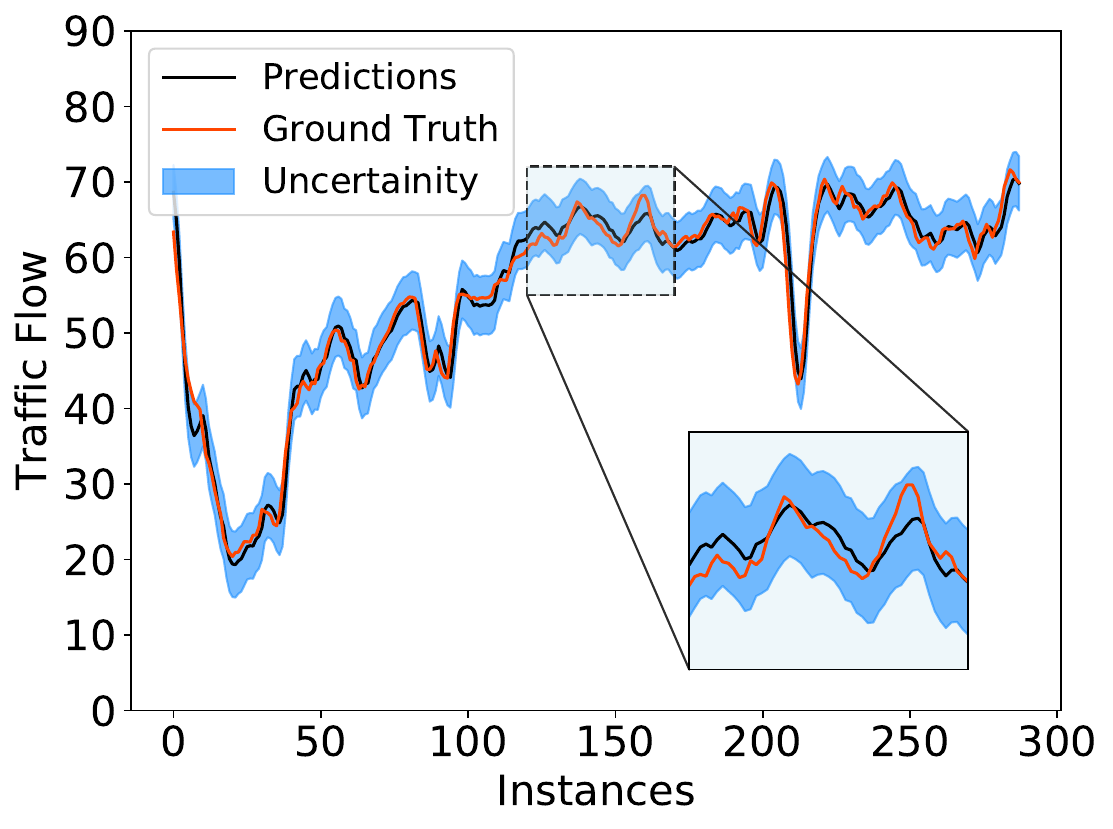}}
\subfloat[Node 18 in PeMSD7(M)]{\includegraphics[width=60mm]{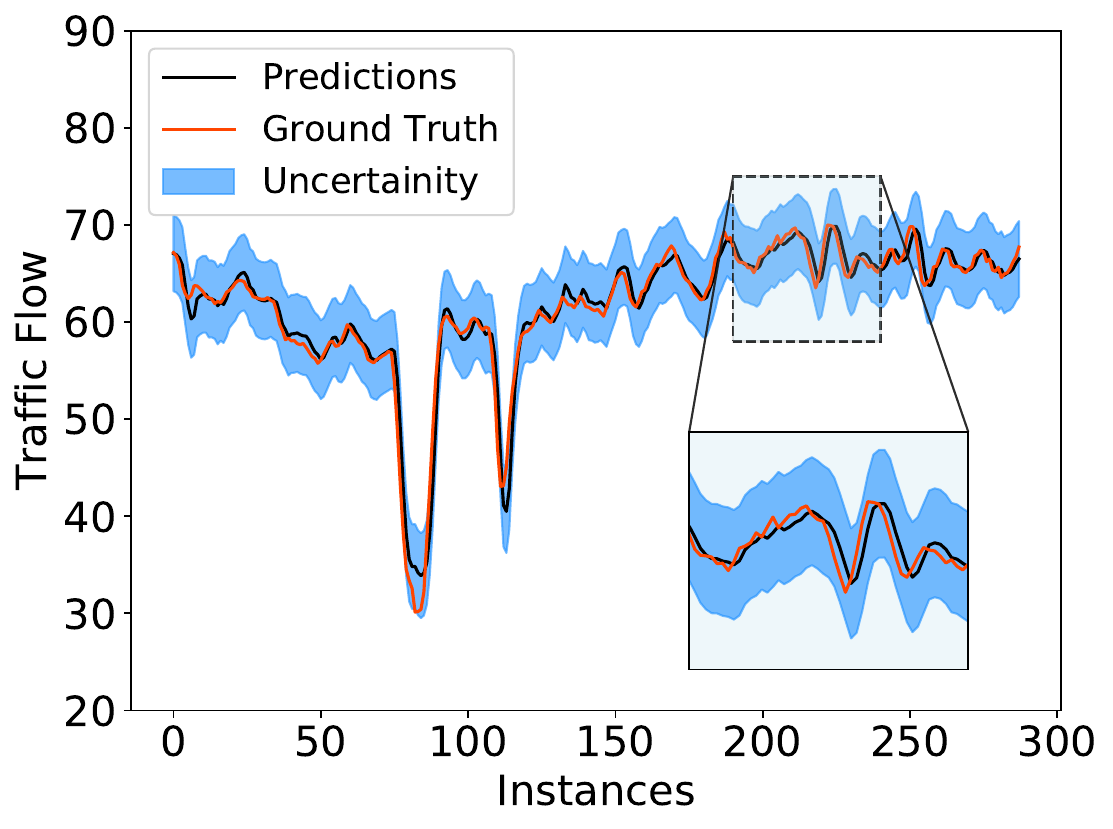}}
}
\end{figure}

\vspace{-3mm}

\vspace{-4mm}
\subsection{Forecasting Uncertainty}  
\vspace{-1mm}
The \textbf{EIKF-Net} framework is a supervised learning algorithm that utilizes the mean absolute error(MAE) as the loss function to train the model. We compute the MAE error between the model pointwise forecasts(represented by \resizebox{.1\textwidth}{!}{$\hat{\mathbf{X}}_{(t  : t + \upsilon-1)}$}) and ground-truth data (represented by \resizebox{.1\textwidth}{!}{$\mathbf{X}_{(t  : t + \upsilon-1)}$}) as described below,

\vspace{-5mm}
\resizebox{0.935\linewidth}{!}{
\begin{minipage}{\linewidth}
\begin{align}
\mathcal{L}_{\text{MAE}}\left(\theta\right) =\frac{1}{\upsilon}\left|\mathbf{X}_{(t  : t + \upsilon-1)}-\hat{\mathbf{X}}_{(t  : t + \upsilon-1)}\right| \label{eq:UCE}
\end{align}
\end{minipage}
} 

\vspace{-1mm}
The training process aims to minimize the MAE loss function, represented by $\mathcal{L}_{\text{MAE}}\left(\theta\right)$, by adjusting the model parameters $\theta$. The \textbf{w/Unc- EIKF-Net} is a variation of the \textbf{EIKF-Net} designed to estimate the uncertainty of the model predictions and enables more reliable decision-making. The \textbf{w/Unc- EIKF-Net} framework is a method for predicting the time-varying uncertainty in multistep-ahead forecasts. The predicted forecasts(\resizebox{.1025\textwidth}{!}{$\hat{\mathbf{X}}_{(t  : t + \upsilon-1)}$}) are modeled as a heteroscedastic Gaussian distribution with mean and variance given by \resizebox{.135\textwidth}{!}{$\mu_\phi\big(\bar{\mathbf{X}}_{(t - \tau : \hspace{1mm}t-1)}\big)$} and \resizebox{.135\textwidth}{!}{$\sigma_\phi^2\big(\bar{\mathbf{X}}_{(t - \tau : \hspace{1mm}t-1)}\big)$}, respectively, where \resizebox{.085\textwidth}{!}{$\bar{\mathbf{X}}_{(t - \tau : \hspace{1mm}t-1)}$} is the input time series. It is described as follows, 

\vspace{-5mm}
\resizebox{0.935\linewidth}{!}{
\begin{minipage}{\linewidth}
\begin{align}
\hat{\mathbf{X}}_{(t  : t + \upsilon-1)} \hspace{0.5mm} \sim \hspace{1mm}\mathcal{N}\big(\mu_\phi\big(\bar{\mathbf{X}}_{(t - \tau : \hspace{1mm}t-1)}\big), \sigma_\phi^2\big(\bar{\mathbf{X}}_{(t - \tau : \hspace{1mm}t-1)}\big)\big) 
\end{align}
\end{minipage}
}

The predicted mean and standard deviation are obtained as follows,

\vspace{-4mm}
\resizebox{0.935\linewidth}{!}{
\begin{minipage}{\linewidth}
\begin{align}
\mu_\phi\big(\bar{\mathbf{X}}_{(t - \tau : \hspace{1mm}t-1)}\big), \sigma_\phi^2\big(\bar{\mathbf{X}}_{(t - \tau : \hspace{1mm}t-1)}\big) &= f_\theta\big( \mathbf{H^{\prime\prime\prime\prime}_{t}}\big) 
\end{align}
\end{minipage}
}

\vspace{0mm}
The neural network($f_{\theta}$) takes in the output of the temporal inference component, represented by \resizebox{.045\textwidth}{!}{$\mathbf{H^{\prime\prime\prime\prime}_{t}}$}, as input. The network then predicts the mean\resizebox{.165\textwidth}{!}{$\big(\mu_\phi\big(\bar{\mathbf{X}}_{(t - \tau : \hspace{1mm}t-1)}\big)\big)$} and standard deviation\resizebox{.165\textwidth}{!}{$\big(\sigma_\phi^2\big(\bar{\mathbf{X}}_{(t - \tau : \hspace{1mm}t-1)}\big)\big)$} of a normal distribution for future observations(\resizebox{.105\textwidth}{!}{$\mathbf{X}_{(t  : t + \upsilon-1)}$}). The maximum likelihood estimate(MLE) of the predicted Gaussian distribution denoted as \resizebox{.105\textwidth}{!}{$\hat{\mathbf{X}}_{(t  : t + \upsilon-1)}$}. It is obtained as follows,

\vspace{-6mm}
\resizebox{0.945\linewidth}{!}{
\begin{minipage}{\linewidth}
\begin{align}
\hat{\mathbf{X}}_{(t  : t + \upsilon-1)} = \mu_\phi\big(\bar{\mathbf{X}}_{(t - \tau : \hspace{1mm}t-1)}\big) 
\end{align}
\end{minipage}
}

\vspace{-1mm}
This means that \resizebox{.145\textwidth}{!}{$\mu_\phi\big(\bar{\mathbf{X}}_{(t - \tau : \hspace{1mm}t-1)}\big)$} is an estimate of the expected value of the Gaussian distribution of future values given the observed values from time points \resizebox{.045\textwidth}{!}{$t - \tau$} to \resizebox{.045\textwidth}{!}{$t-1$}. \resizebox{.145\textwidth}{!}{$\sigma_\phi^2\big(\bar{\mathbf{X}}_{(t - \tau : \hspace{1mm}t-1)}\big)$} denotes the model uncertainty predictions for the next $\upsilon$ time steps from the current $t^{th}$ time point. The uncertainty modeling framework optimizes the negative Gaussian log likelihood(\cite{nix1994estimating}) of the observations for current $t^{th}$ time point based on its  mean and variance estimates as described by,

\vspace{-4mm}
\resizebox{0.875\linewidth}{!}{
\begin{minipage}{\linewidth}
\begin{equation*}
\mathcal{N}(\hat{\mathbf{X}}_{(t  : t + \upsilon-1)};\mu_\phi\big(\bar{\mathbf{X}}_{(t - \tau : \hspace{1mm}t-1)}\big),\sigma_\phi\big(\bar{\mathbf{X}}_{(t - \tau : \hspace{1mm}t-1)}\big))={\frac {1}{\sigma_\phi\big(\bar{\mathbf{X}}_{(t - \tau : \hspace{1mm}t-1)}\big) {\sqrt {2\pi }}}}\hspace{1.5mm} e^{-{\dfrac {1}{2}}\left({\dfrac {\mathbf{X}_{(t  : t + \upsilon-1)}-\mu_\phi\big(\bar{\mathbf{X}}_{(t - \tau : \hspace{1mm}t-1)}\big) }{\sigma_\phi\big(\bar{\mathbf{X}}_{(t - \tau : \hspace{1mm}t-1)}\big)}}\right)^{2}}
\end{equation*} 
\end{minipage} 
}

The negative Gaussian log likelihood is a measure of the likelihood of the observations, given the estimated mean and variance of the Gaussian distribution. The lower the negative Gaussian log likelihood, the better the fit of the Gaussian distribution to the observed values. We apply logarithm transformation on both sides of the equation, 

\vspace{-5mm}
\resizebox{1.175\linewidth}{!}{
\begin{minipage}{\linewidth}
\begin{align*}
\log\ \mathcal{N}(\hat{\mathbf{X}}_{(t  : t + \upsilon-1)};\mu_\phi\big(\bar{\mathbf{X}}_{(t - \tau : \hspace{1mm}t-1)}\big),\sigma_\phi\big(\bar{\mathbf{X}}_{(t - \tau : \hspace{1mm}t-1)}\big)) &= \log\left[{\frac {1}{\sigma_\phi\big(\bar{\mathbf{X}}_{(t - \tau : \hspace{1mm}t-1)}\big) {\sqrt {2\pi }}}}\right] + \log \left[e^{-{\dfrac {1}{2}}\left({\dfrac {\mathbf{X}_{(t  : t + \upsilon-1)}-\mu_\phi\big(\bar{\mathbf{X}}_{(t - \tau : \hspace{1mm}t-1)}\big) }{\sigma_\phi\big(\bar{\mathbf{X}}_{(t - \tau : \hspace{1mm}t-1)}\big)}}\right)^{2}}\right] \\
 &= \log\ {\frac {1}{\sigma_\phi\big(\bar{\mathbf{X}}_{(t - \tau : \hspace{1mm}t-1)}\big)}} + \log\ {\frac {1}{{\sqrt {2\pi }}}} -{\frac {1}{2}}\left(\dfrac {\mathbf{X}_{(t  : t + \upsilon-1)}-\mu_\phi\big(\bar{\mathbf{X}}_{(t - \tau : \hspace{1mm}t-1)}\big) }{\sigma_\phi\big(\bar{\mathbf{X}}_{(t - \tau : \hspace{1mm}t-1)}\big)}\right)^{2} \\
 &= -\log\ \sigma_\phi\big(\bar{\mathbf{X}}_{(t - \tau : \hspace{1mm}t-1)}\big) + C -{\frac {1}{2}}\left(\dfrac {\mathbf{X}_{(t  : t + \upsilon-1)}-\mu_\phi\big(\bar{\mathbf{X}}_{(t - \tau : \hspace{1mm}t-1)}\big) }{\sigma_\phi\big(\bar{\mathbf{X}}_{(t - \tau : \hspace{1mm}t-1)}\big)}\right)^{2}
\end{align*}
\end{minipage}
}

We drop the constant(C) and the Gaussian negative log likelihood loss(i.e., negative log gaussian probability density function(pdf)) for the full training dataset is described by, 

\vspace{-2mm}
\resizebox{0.875\linewidth}{!}{
\begin{minipage}{\linewidth}
\begin{align}
\hspace{5mm}\mathcal{L}_{\text{GaussianNLLLoss}} =  \sum_{t=1}^{\text{T}} \left[\frac{\log \sigma_\phi\big(\bar{\mathbf{X}}_{(t - \tau : \hspace{1mm}t-1)}\big)^2}{2}+\frac{\left(\mathbf{X}_{(t  : t + \upsilon-1)}-\mu_\phi\left(\bar{\mathbf{X}}_{(t - \tau : \hspace{1mm}t-1)}\right)\right)^2}{2 \sigma_\phi\left(\bar{\mathbf{X}}_{(t - \tau : \hspace{1mm}t-1)}\right)^2}\right] \label{eq:GUCE}
\end{align}
\end{minipage}
}

\vspace{1mm}
In summary, the \textbf{EIKF-Net} framework minimizes $\mathcal{L}_{\text{MSE}}$(refer Equation \ref{eq:UCE}) and the \textbf{w/Unc- EIKF-Net} framework(i.e., \textbf{EIKF-Net} with local uncertainty estimation) minimizes $\mathcal{L}_{\text{GaussianNLLLoss}}$(refer Equation \ref{eq:GUCE}) respectively. Simply, the \textbf{w/Unc- EIKF-Net} framework uses a Gaussian likelihood function in which the mean and variance are modeled by the neural network. The mean is the predicted value and the variance represents the uncertainty of the predictions. Minimizing the negative log likelihood of the Gaussian distribution finds the set of model parameters that provide the best fit to the data and also provides the uncertainty estimates.

\vspace{-2mm}
\subsection{Baselines}
\vspace{-2mm}

We briefly discuss the well-known and well-understood algorithms which provide a benchmark for evaluating the proposed neural forecasting models(\textbf{w/Unc- EIKF-Net}, \textbf{w/Unc- EIKF-Net}) performance on the MTSF task.

\vspace{-1mm}
\begin{itemize}
\item HA~\cite{hamilton2020time} uses the average of the predefined historical window based observations to predict the next value.
\item ARIMA is a statistical analysis model for handling non-stationary time series data.
\item VAR(\cite{hamilton2020time}) is an extension of the univariate autoregressive model(AR) and a linear multivariate time series model that captures the inter-dependencies among multiple time series variables.
\item TCN(~\cite{BaiTCN2018}) is an adaptive model for handling the sequential data for multistep-ahead time series prediction. It comprises a stack of causal convolutions and dilation layers to increase the receptive field of the network exponentially. Causal convolutions takes past information into account, while dilation layers increase the receptive field of the convolutional filters to learn long-range correlations in the MTS data.
\item FC-LSTM(~\cite{sutskever2014sequence}) is an encoder-decoder framework using Long Short-term Memory(LSTM) units with a peephole connection for multistep-ahead time series prediction.
\item GRU-ED(~\cite{cho2014grued}) is an encoder-decoder framework using a GRU-based baseline for multistep-ahead time series prediction.
\item DSANet(~\cite{Huang2019DSANet}) does not rely on recurrence to capture temporal dependencies in the MTS data. It is a correlated time series prediction model using CNN networks for capturing long-range intra-temporal dependencies of multiple time series and self-attention block that adaptively captures the inter-dependencies for multistep-ahead time series prediction.
\item DCRNN(~\cite{li2018dcrnn_traffic}) utilizes bidirectional random walks on graphs and combines graph convolution with recurrent neural networks to capture the spatial-temporal dynamics through an encoder-decoder approach for multistep-ahead time series prediction.
\item STGCN(~\cite{bing2018stgcn}) combines graph convolution with gated temporal convolution for learning spatial-temporal correlations of multiple time series variables for  multistep-ahead time series prediction.
\item GraphWaveNet(~\cite{wu2019graphwavenet}) jointly learns the adaptive dependency matrix using a wave-based propagation mechanism and the graph representations with dilated casual convolution to capture spatial-temporal dependencies for multistep-ahead time series prediction.
\item ASTGCN(~\cite{guo2019astgcn}) utilizes an attention-based spatio-temporal graph convolutional convolutional network to capture inter- and intra-dependencies for multi-step time series prediction.
\item STG2Seq(~\cite{bai2019STG2Seq}) uses a combination of gated graph convolutional networks(GGCNs) and a sequence-to-sequence(seq2seq) architecture with an attention mechanism to model the dynamic temporal and cross-channel information for multistep-ahead time series prediction.
\item STSGCN(~\cite{song2020stsgcn}) stacks multiple layers of spatial-temporal graph convolutional network layers to capture the localized intra- and inter-dependencies from the  graph-structured MTS data for multistep-ahead time series prediction.
\item LSGCN(~\cite{huang2020lsgcn}) integrates a graph attention mechanism into a spatial gated block to capture the dynamic spatial-temporal dependencies for multistep-ahead time series prediction.
\item AGCRN(~\cite{NEURIPS2020_ce1aad92}) utilizes data-adaptive graph structure learning in the absence of a predefined graph to capture the node-specific intra- and inter-correlations to capture complex spatial-temporal dependencies for multistep-ahead time series prediction.
\item STFGNN(~\cite{li2021stfgnn}) learns the spatial-temporal correlations by fusing representations obtained from the temporal graph module and the gated convolution module, which are operated for different time periods in parallel for multistep-ahead time series prediction.
\item Z-GCNETs(~\cite{chen2021ZGCNET}) is a model for multistep-ahead time series prediction that learns the salient time-conditioned topological information of the MTS data by integrating a time-aware zigzag topological layer into time-conditioned graph convolutional networks(GCNs) to capture hidden spatial-temporal dependencies. 
\item STGODE(~\cite{fang2021STODE}) utilizes a tensor-based ordinary differential equation(ODE) to capture inter- and intra-dependency dynamics of MTS data for multistep-ahead prediction.  
\end{itemize}

\vspace{-4mm} 
\subsection{Datasets} 
We compare the performance of our proposed \textbf{EIKF-Net} framework to benchmark models on several real-world datasets(PeMSD3, PeMSD4, PeMSD7, PeMSD7(M), PeMSD8, METR-LA, and PEMS-BAY).  We utilize kernel-based similarity metric functions to generate the pre-defined graphs for the traffic-related benchmark datasets(\cite{chen2001freeway, choi2022graph, LiYS018}) based on the distance between sensors on the road network.  Table \ref{tab:summarydatasets} reports more details of the benchmark datasets.

\vspace{-1mm}
\begin{table}[tbhp]
\center
\setlength{\tabcolsep}{0.25em} 
\renewcommand\arraystretch{1.325} 
\centering
 \resizebox{0.7\textwidth}{!}{
\begin{tabular}{c|c|c|c|c|c}
\hline
\textbf{Dataset} & \textbf{Nodes} & \textbf{Timesteps} & \textbf{Time-Range} & \multicolumn{1}{l|}{\textbf{Data Split}} & \multicolumn{1}{l}{\textbf{Granularity}} \\ \hline
PeMSD3 & 358 & 26,208 & 09/2018 - 11/2018 & \multirow{5}{*}{6 / 2 / 2} & \multirow{7}{*}{\rotatebox[origin=c]{270}{5 mins}} \\
PeMSD4 & 307 & 16,992 & 01/2018 - 02/2018 &  &  \\
PeMSD7 & 883 & 28,224 & 05/2017 - 08/2017 &  &  \\
PeMSD8 & 170 & 17,856 & 07/2016 - 08/2016 &  &  \\
PeMSD7(M) & 228 & 12,672 & 05/2012 - 06/2012 &  &  \\ \cline{1-5}
METR-LA & 207 & 34,272 & 03/2012 - 06/2012 & \multirow{2}{*}{7 / 1 / 2} &  \\
PEMS-BAY & 325 & 52,116 & 01/2017 - 05/2017 &  &  \\ \hline
\end{tabular}
}
\vspace{-1mm}
\caption{Summary of the various traffic-related datasets for the MTSF task.}
\label{tab:summarydatasets}
\end{table}
 
\vspace{-1mm} 
\subsection{Experiment setup} 
\vspace{-1mm}  
The traffic-related benchmark datasets are split into a training set to learn from data and train the model, a validation set to fine-tune the model hyperparameters, and test sets to evaluate the performance of the models on unseen data with different ratios. We split the PEMS-BAY and METR-LA datasets(\cite{LiYS018}) with a 7/1/2 ratio and for all other datasets(\cite{choi2022graph}) with a 6/2/2 ratio into training, validation, and test sets, respectively. The time series data preprocessing involves scaling each time-series variable to have zero mean and unit variance. We compute various forecast accuracy metrics such as MAE, RMSE, and MAPE on the original scale of the MTS data during the training and evaluation of forecasting models. We train the \textbf{w/Unc- EIKF-Net} framework for a predefined number of 30 epochs to iteratively update the learnable parameters of the model to learn from the training set with the aim of minimizing the forecast error. We utilize validation MAE for early stopping to prevent overfitting. This is done to avoid a suboptimal solution and select the best model that can generalize well to unseen data, improving the overall performance of the model. We evaluate the model performance on the test set to assess its ability to generalize to new, unseen data. We utilize a learning rate scheduler to drop the learning rate by half if the evaluation metrics show no improvement on the validation set for a patience number of 5 epochs. We train our models using the Adam optimizer to fine-tune the learnable parameters and improve convergence with an initial learning rate of \num{1e-3} to minimize the (a) MAE loss for the \textbf{EIKF-Net} model and (b) the negative Gaussian log-likelihood for the \textbf{w/Unc- EIKF-Net} model between the ground truth and the model predictions. We train our model on powerful GPUs(NVIDIA Tesla T4, Nvidia Tesla v100, and GeForce RTX 2080 GPUs) to significantly speed up the training process and enable the use of larger models and larger datasets built upon the PyTorch framework. Multiple independent experimental runs are performed. We report the ensemble average to provide a reliable evaluation of the models. The hyperparameters of our learning  algorithm are the embedding size($\textit{d}$), the number of hyperedges($|\mathcal{HE}|$), batch size($\textit{b}$), and the learning rate($\textit{lr}$). Section \ref{Sanalysis} reports the optimum hyperparameter values for each dataset.

\vspace{-4mm}
\subsection{Related Work}
\vspace{-2mm}
Forecasting multivariate time series is a crucial task that is widely used in various domains to drive strategic decision-making. It allows making informed decisions about strategy formulation, resource allocation, and other critical business processes by predicting multiple related variables that change over time. MTSF has a wide range of applications across different domains, such as in the manufacturing sector for production planning and inventory management; in finance, it can assist in investment planning and risk management, while in retail it helps in inventory management, product pricing, and so on. The traditional time series forecasting methods for handling correlated time series data such as VAR(\cite{watson1994vector}), ARIMA(\cite{makridakis1997arma}), and state-space models(SSMs, \cite{hyndman2008forecasting}) are limited in their ability to capture nonlinear relationships and complex dependencies among variables in MTS data. In recent years, deep learning-based methods, particularly FC-LSTM(\cite{sutskever2014sequence}) and TPA-LSTM(\cite{shih2019temporal}) have been proposed to overcome the inherent limitations of statistical methods to model complex spatial-temporal dependencies for MTSF task. These methods utilize an implicit recurrent process to model intra-temporal correlations. However, these methods may not explicitly consider the inter-correlations among different time series variables, which can lead to limitations in their ability to capture the full complexity of MTS data. Some recent research has proposed various methods that explicitly model the inter-correlations among the different time series variables in addition to the intra-correlations. These methods include  convolutional LSTM(ConvLSTM), recurrent attention networks(RAN), dynamic self-attention(DSA), temporal convolutional networks(TCN), transformer-based architectures, etc. LSTNet is a popular hybrid method for MTSF tasks proposed by \cite{lai2018modeling}. It uses a combination of convolutional layers(CNNs) and recurrent neural networks(RNNs) to capture the inter-dependencies among the variables and complex intra-temporal dependencies within time series variables.  However, global aggregation of pair-wise semantic relations is inefficient in modeling to precisely capture the complex relationships and dependencies within the different variables of MTS data. In recent years, transformer-based forecasting methods such as Autoformer(\cite{wu2021autoformer}), FEDformer(\cite{zhou2022fedformer}), Pyraformer(\cite{liu2021pyraformer}), LogTrans(\cite{li2019enhancing}), Informer(\cite{zhou2021informer}), and TFT(\cite{lim2021temporal}) have gained popularity for forecasting in dynamic systems characterized by complex long-range dependencies and interactions. More recently, MTSF has embraced Graph neural networks(GNNs, \cite{zonghanwu2019, Bai2020nips, wu2020connecting, YuYZ18, tampsgcnets2022, LiYS018}) and have become increasingly popular that involve modeling the structural dependencies between variables in the MTS data. These structural dependence approaches for MTSF require a predefined graph that represents the dependencies between variables as edges in the graph and predicts the dynamics with Graph Neural Networks(GNNs). However, in most real-world scenarios, the explicit graph structure is often unavailable or inaccurate. In addition, the predefined graph structure can be a limitation for modeling high-dimensional dynamics of the interconnected system for which it may not be able to capture the full complexity of the underlying relational graph structure between variables, leading to suboptimal forecasting. Additionally, the forecast models relying on predefined graphs, which typically assume a fixed relationship between variables, neglect the potential to incorporate the relational graph structure accounting for the dynamic relationships among the variables underlying the spatio-temporal data to make more accurate predictions. Of late, a class of GNN-based models(\cite{shang2021discrete, deng2021graph, wu2020connecting}) utilizes a time-varying adaptive attention mechanism(\cite{deng2021graph}) for modeling the complex dynamic correlations among variables in MTS data to improve forecast accuracy. These GNN-based models are focused on jointly learning the implicit graph structure and representations of the spatio-temporal graph data rather than relying on predefined graphs. However, incorporating the prior or domain knowledge of the interconnected networks further improves the model performance, especially when the explicit graph structure is completely available and reliable. Additionally, incorporating relevant prior knowledge overcomes the limitations of neural relational inference methods which rely solely on the data, especially in scenarios when the data is limited or noisy for improved forecast accuracy and robustness. It can also help to constrain the search space, regularize the model and improve its robustness. Nonetheless, the GNN-based models combined with graph-pooling operations model multi-hop relationships among the time series variables but suffer from over-smoothing and under-learning issues. Furthermore, the higher-order graph convolutions or modeling the graph as a simplicial complex are ineffective in capturing the higher-order correlations in the spatio-temporal graph data. Based on the above shortcomings, GNN-based models suffer from inherent limitations in their ability to model higher-order interdependencies in graph-structured data which results in less accurate and reliable forecasts. Nevertheless, current methods for MTSF with deterministic GNN-based models have shown promising results in  pointwise forecast accuracy but generate poor estimates of uncertainty. In literature, various techniques for uncertainty quantification approaches include Bayesian neural networks, Gaussian processes, ensemble learning techniques, bootstrapping, jackknife resampling methods based on post-hoc uncertainty estimates, etc. However, more research is needed to advance and improve GNN-based models to effectively estimate uncertainty for MTSF task. In short, there is a need and necessity for interpretable and tractable uncertainty estimates of graph-based forecasting neural network predictions. We overcome the above limitations by jointly learning explicit(predefined) graph and implicit hypergraph structure representations for modeling the complex inter-series correlations and long-term dependencies among variables in MTS data to achieve better forecast accuracy in the MTSF task. In addition, the framework reduces computational requirements to handle larger datasets and improve forecast accuracy. The use of cost-effective GPU hardware accelerates the training process and reduces memory requirements, making the framework more practical for real-world use.

\end{document}